\documentclass{article}
\usepackage[table]{xcolor}

\usepackage[final]{neurips_2019}

\usepackage[utf8]{inputenc} 
\usepackage[T1]{fontenc}    
\usepackage{hyperref}       
\usepackage{url}            
\usepackage{booktabs}       
\usepackage{amsfonts}       
\usepackage{nicefrac}       
\usepackage{microtype}      

\usepackage{amsmath}
\usepackage{amssymb}
\usepackage{makecell}

\usepackage[linewidth=1pt]{mdframed}
\usepackage{tikz-dependency}
\usepackage{amsmath, graphicx, amsfonts, amssymb}
\usepackage{algorithmicx}
\usepackage{algorithm}
\usepackage{booktabs}
\usepackage[noend]{algpseudocode}
\usepackage[percent]{overpic}
\usepackage{subcaption}
\usepackage[page]{appendix}

\usepackage{listings}
\NewEnviron{elaboration}{
\par
\begin{tikzpicture}
\node[rectangle,minimum width=\textwidth] (m) {\begin{minipage}{.98\textwidth}\BODY\end{minipage}};
\draw[dashed] (m.south west) rectangle (m.north east);
\end{tikzpicture}
}

\begin{document}

\title{Interactive Fiction Games: A Colossal Adventure}
\author{
Matthew Hausknecht\\
Microsoft Research\\
\And
Prithviraj Ammanabrolu\\
Georgia Institute of Technology\\
\And
Marc-Alexandre C\^ot\'{e}\\
Microsoft Research Montr\'{e}al\\
\And
Xingdi Yuan\\
Microsoft Research Montr\'{e}al\\
}
\maketitle
\begin{abstract}
  A hallmark of human intelligence is the ability to understand and communicate with language. Interactive Fiction games are fully text-based simulation environments where a player issues text commands to effect change in the environment and progress through the story. We argue that IF games are an excellent testbed for studying language-based autonomous agents. In particular, IF games combine challenges of combinatorial action spaces, language understanding, and commonsense reasoning. To facilitate rapid development of language-based agents, we introduce Jericho, a learning environment for man-made IF games and conduct a comprehensive study of text-agents across a rich set of games, highlighting directions in which agents can improve.
\end{abstract}

\section{Introduction}
\begin{figure}[htp]
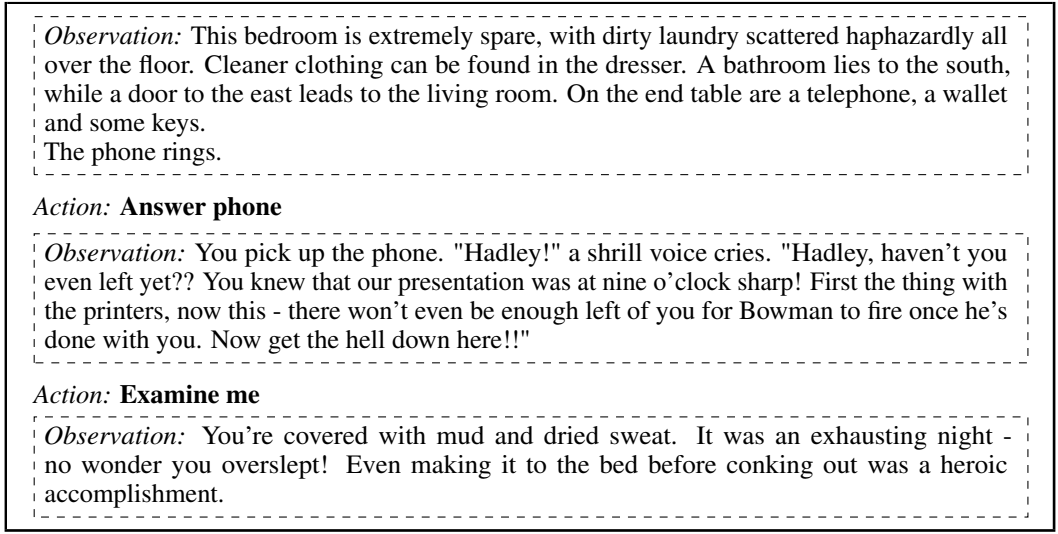

\begin{mdframed}
\begin{elaboration}
  \parbox{.99\textwidth}{
\emph{Observation:} This bedroom is extremely spare, with dirty laundry scattered haphazardly all over the floor. Cleaner clothing can be found in the dresser. A bathroom lies to the south, while a door to the east leads to the living room. On the end table are a telephone, a wallet and some keys.\\
The phone rings.
}
\end{elaboration}
\begin{flushleft}
\emph{Action:} \textbf{Answer phone}
\end{flushleft}
\begin{elaboration}
  \noindent\parbox{.99\textwidth}{
\emph{Observation:} You pick up the phone. "Hadley!" a shrill voice cries. "Hadley, haven't you even left yet?? You knew that our presentation was at nine o'clock sharp! First the thing with the printers, now this - there won't even be enough left of you for Bowman to fire once he's done with you. Now get the hell down here!!"
}
\end{elaboration}
\begin{flushleft}
\emph{Action:} \textbf{Examine me}
\end{flushleft}
\begin{elaboration}
  \noindent\parbox{.99\textwidth}{
\emph{Observation:} You're covered with mud and dried sweat. It was an exhausting night - no wonder you overslept! Even making it to the bed before conking out was a heroic accomplishment.
}
\end{elaboration}
\end{mdframed}
\caption{Excerpt from Adam Cadre's 9:05.}
\label{fig:905}
\end{figure}

Interactive fiction (IF) games are software environments in which players observe textual descriptions of the simulated world, issue text actions, and receive score as they progress through the story. As the excerpt of an IF game in Figure \ref{fig:905} shows, humans bring competencies in natural language understanding, commonsense reasoning, and deduction to bear in order to infer the context and objectives of a game. We believe that IF environments provide a good testbed for studying the development of these capabilities in artificial agents. Beyond games, real-world applications such as voice-activated personal assistants can also benefit from advances in these capabilities at the intersection of natural language understanding, natural language generation, and sequential decision making. These real world applications require the ability to reason with ungrounded natural language (unlike multimodal environments that provide visual grounding for language) and IF games provide an excellent suite of environments to tackle these challenges.

From a machine learning perspective, Interactive Fiction games exist at the intersection of natural language processing and sequential decision making. Like many NLP tasks, they require natural language understanding, but unlike most NLP tasks, IF games are sequential decision making problems in which actions change the subsequent world states of the game and choices made early in a game may have long term effects on the eventual endings. Reinforcement Learning~\citep{sutton98} studies sequential decision making problems and has shown promise in vision-based~\citep{jaderberg16} and control-based~\citep{openai18} environments, but has less commonly been applied in the context of language-based tasks.



The contributions of this paper are as follows: First, we introduce Jericho, a learning environment for human-made IF games. Second, we introduce a template-based action space that we argue is appropriate for language generation. Third, we conduct an empirical evaluation of learning agents across a large set of human-made games.

\section{Research Challenges}
From the perspective of reinforcement learning, IF games can be modeled as partially observable Markov decision processes (POMDPs) defined by $(S, T, A, O, R)$. Observations $o \in O$ correspond to the game's text responses, while latent states $s \in S$ correspond to player and item locations, inventory contents, monsters, etc. Text-based actions $a \in A$ change the game state according to an mostly-deterministic latent transition function $T(s'|s,a)$, and the agent receives rewards $r$ from an unknown reward function $R(s, a)$. To succeed in these environments, agents must generate natural language actions, reason about entities and affordances, and represent their knowledge about the game world. We present these challenges in greater detail:

\textbf{Combinatorial Action Space} Reinforcement learning has studied agents that operate in discrete or continuous action space environments. However, IF games require the agent to operate in the combinatorial action space of natural language. Combinatorial spaces pose extremely difficult exploration problems for existing agents. For example, an agent generating a four-word sentence from a modest vocabulary of size 700, is effectively exploring a space of $|700^4| = 240$ billion possible actions. Further complicating this challenge, natural language commands are interpreted by the game's parser which recognizes only a subset of possible commands. For example, out of the 240 billion possible actions there may be 100 million that are grammatical and successfully parseable, and out of these there may be only a hundred \textit{valid actions} - contextually relevant actions that generate a change in world state.


\textbf{Commonsense Reasoning} Due to the lack of graphics, IF games rely on the player's commonsense knowledge as a prior for how to interact with the game world. For example, a human player encountering a locked chest intuitively understands that the chest needs to be unlocked with some type of key, and once unlocked, the chest can be opened and will probably contain useful items. They may make a mental note to return to the chest if a key is found in the future. They may even mark the location of the chest on a map to easily find their way back.
These inferences are possible for humans who have years of embodied experience interacting with chests, cabinets, safes, and all variety of objects. Artificial agents lack the commonsense knowledge gained through years of grounded language acquisition and have no reason to prefer opening a chest to eating it. Also known as \textit{affordance extraction}~\citep{gibson77,fulda17}, the problem of choosing which actions or verbs to pair with a particular noun is central to IF game playing. However, the problem of commonsense reasoning extends much further than affordance extraction: Games require planning which items to carry in a limited inventory space, strategically deciding  whether to fight or flee from a monster, and spatial reasoning such as stacking garbage cans against the wall of a building to access a second-floor window.


\textbf{Knowledge Representation}
IF games span many distinct locations, each with unique descriptions, objects, and characters. Players move between locations by issuing navigational commands like \textit{go west}. Due to the large number of locations in many games, humans often create maps to navigate efficiently and avoid getting lost. This gives rise to the \textbf{Textual-SLAM} problem, a textual variant of Simultaneous localization and mapping (SLAM)~\citep{thrun05} problem of constructing a map while navigating a new environment. In particular, because connectivity between locations is not necessarily Euclidean, agents need to detect when a navigational action has succeeded or failed and whether the location reached was previously seen or new. Beyond location connectivity, it's also helpful to keep track of the objects present at each location, with the understanding that objects can be nested inside of other objects, such as food in a refrigerator or a sword in a chest.

\section{Related Work}

One approach to affordance extraction ~\citep{fulda17} identified a vector in word2vec~\citep{mikolov13} space that encodes affordant behavior. When applied to the noun \textit{sword}, this vector produces affordant verbs such as \textit{vanquish}, \textit{impale}, \textit{duel}, and \textit{battle}. The authors use this method to prioritize verbs for a Q-Learning agent to pair with in-game objects.

An alternative strategy has been to reduce the combinatorial action space of parser-based games into a discrete space containing the minimum set of actions required to finish the game. This approach requires a walkthrough or expert demonstration in order to define the space of minimal actions, which limits its applicability to new and unseen games. Following this approach, \cite{zahavy18} employ this strategy with their action-elimination network, a classifier that predicts which predefined actions will not effect any world change or be recognized by the parser. Masking these invalid actions, the learning agent subsequently evaluates the set of remaining valid actions and picks the one with the highest predicted Q-Value.

The TextWorld framework~\citep{cote18} supports procedural generation of parser-based IF games, allowing complexity and content of the generated games to be scaled to the difficulty needed for research. TextWorld domains have already proven suitable for reinforcement learning agents~\citep{yuan18} which were shown to be capable of learning on a set of environments and then generalizing to unseen ones at test time. 
Recently, \citet{yuan2019qait} propose QAit, a set of question answering tasks based on games generated using TextWorld.
QAit focuses on helping agents to learn procedural knowledge in an information-seeking fashion, it also introduces the practice of generating unlimited training games on the fly.
With the ability to scale the difficulty of domains, TextWorld enables creating a curriculum of learning tasks and helping agents eventually scale to human-made games.

\cite{ammanabrolu} present the Knowledge Graph DQN or KG-DQN, an approach where a knowledge graph built during exploration is used as a state representation for a deep reinforcement learning based agent.
They also use question-answering techniques---asking the question of what action is best to take next---to pretrain a deep $Q$-network.
These techniques are then shown to aid in overcoming the twin challenges of a partially observable state space and a combinatorial action space.
\cite{ammanabrolutransfer} further expand on this work, exploring methods of transferring control policies in text-games, using knowledge graphs to seed an agent with useful commonsense knowledge and to transfer knowledge between different games within a domain.
They show that training on a source game and transferring to target game within the same genre---e.g. horror
---is more effective and efficient than training from scratch on the target game.

Finally, although not a sequential decision making problem, Light~\citep{urbanek2019light} is a crowdsourced dataset of text-adventure game dialogues. The authors demonstrate that supervised training of transformer-based models have the ability to generate contextually relevant dialog, actions, and emotes.

\section{Jericho Environment}
Jericho is an open-source\footnote{Jericho is available at \url{https://github.com/Microsoft/jericho}.} Python-based IF environment, which provides an OpenAI-Gym-like interface~\citep{brockman16gym} for learning agents to connect with IF games. Jericho is intended for reinforcement learning agents, but also supports the ability to load and save game states, enabling planning algorithms Monte-Carlo Tree Search~\citep{coulom07} as well as reinforcement learning approaches that rely on the ability to restore state such as Backplay~\citep{resnick18} and GoExplore~\citep{ecoffet19}. Jericho additionally provides the option to seed the game's random number generator for replicability.\footnote{Most IF games are deterministic environments. Notable exceptions include Anchorhead and Zork1.}  

Jericho supports a set of human-made IF games that cover a variety of genres: dungeon crawl, Sci-Fi, mystery, comedy, and horror. Games were selected from classic Infocom titles such as {\em Zork} and {\em Hitchhiker's Guide to the Galaxy}, as well as newer, community-created titles like {\em Anchorhead} and {\em Afflicted}. Supported games use a point-based scoring system, which serves as the agent's reward. Beyond the set of supported games, unsupported games may be played through Jericho, without the support of score detection, move counts, or world-change detection. 


\textbf{Template-Based Action Generation}
We introduce a novel template-based action space in which the agent first chooses an action template (e.g. \emph{put \_ in \_}) and then fills in the blanks using words from the parser's vocabulary. Notationally, we employ $u \Leftarrow w_1, w_2$ to denote the filling of template $u$ with vocabulary words $w_1, w_2$. Jericho provides the capability to extract game-specific vocabulary and action templates. These templates contain up to two blanks, so a typical game with 200 templates and a 700 word vocabulary yields an action space of $\mathcal{O}(\mathcal{T}\mathcal{V}^2) \approx 98$ million, three orders of magnitude smaller than the 240-billion space of 4-word actions using vocabularly alone.

\textbf{World Object Tree}
The world object tree\footnote{More on game trees \url{https://inform-fiction.org/zmachine/standards/z1point1/index.html}.} is a semi-interpretable latent representation of game state used to codify the relationship between the objects and locations that populate the game world. Each object in the tree has a \emph{parent}, \emph{child}, and \emph{sibling}. Relationships between objects are used to encode posession: a location object contains children corresponding to the items present at that location. Similarly, the player object has the player's current location as a parent and inventory items as children. Possible applications of the object tree include ground-truth identification of player location, ground-truth detection of the objects present at the player's location, and world-change-detection.

\textbf{Identifying Valid Actions}
\label{sec:valid}
\emph{Valid actions} are actions recognized by the game's parser that cause changes in the game state. When playing new games, identifying valid actions is one of the primary difficulties encountered by humans and agents alike. Jericho has the facility to detect valid actions by executing a candidate action and looking for resulting changes to the world-object-tree. However, since some changes in game state are reflected only in global variables, it's rare but possible to experience false negatives. In order to identify all the valid actions in a given state, Jericho uses the following procedure:

\begin{algorithm}
\caption{Procedure for Identifying Valid Actions}
\label{alg:valid}
\begin{algorithmic}[1]
\State $\mathcal{E} \leftarrow$ Jericho environment
\State $\mathcal{T} \leftarrow$ Set of action templates 
\State $o \leftarrow$ Textual observation 
\State $\mathcal{P} \leftarrow \{ p_1 \dots p_n \}$ Interactive objects identified with noun-phrase extraction or world object tree.
\State $Y \leftarrow \varnothing$ List of valid actions
\State $s \leftarrow \mathcal{E}.save()$ -- Save current game state
\For {template $u \in \mathcal{T}$}
\For {all combinations $p_1, p_2 \in \mathcal{P}$}
    \State Action $a \leftarrow u \Leftarrow p_1, p_2$
    \If {$\mathcal{E}.world\_changed(\mathcal{E}.step(a))$}
        \State $Y \leftarrow Y \cup a$
        \State $\mathcal{E}.load(s)$ -- Restore saved game state
    \EndIf
\EndFor
\EndFor
\Return $Y$
\end{algorithmic}
\end{algorithm}

\textbf{Handicaps}
\label{sec:handicaps}
In summary, to ease the difficulty of IF games, Jericho optionally provides the following handicaps: 1) Fixed random seed to enforce determinism. 2) Use of Load, Save functionality. 3) Use of game-specific templates $\mathcal{U}$ and vocabulary $\mathcal{V}$. 4) Use of world object tree as an auxillary state representation or method for detecting player location and objects. 5) Use of world-change-detection to identify valid actions. For reproducibilty, we report the handicaps used by all algorithms in this paper and encourage future work to do the same.

\section{Algorithms}
In this section we present three agents: a \emph{choice-based} single-game agent (DRRN), a \emph{parser-based} single-game agent (TDQN), and a parser-based general-game agent (NAIL). Single-game agents are trained and evaluated on the same game, while general game playing agents are designed to play unseen games.


\textbf{Common Input Representation}
The input encoder $\phi_o$ converts observations into vectors using the following process: Text observations are tokenized by a SentencePiece model~\citep{kudo18} using an 8000-large vocabulary trained on strings extracted from sessions of humans playing a variety of different IF games\footnote{\url{http://www.allthingsjacq.com/index.html}}. Tokenized observations are processed by separate GRU encoders for the narrative (i.e., the game's response to the last action), description of current location, contents of inventory, and previous text command. The outputs of these encoders are concatenated into a vector $\nu_o$. DRRN and Template-DQN build on this common input representation.

\textbf{DRRN}
The Deep Reinforcement Relevance Network (DRRN)~\cite{he16} is an algorithm for \emph{choice-based} games, which present a set of valid actions $A_{valid}(s)$ at every game state. We implement DRRN using a GRU $\phi_{act}(a)$ to encode each valid action into a vector $\nu_a$, which is concatenated with the encoded observation vector $\nu_o$. Using this combined vector, DRRN then computes a Q-Value $Q(o, a)$ estimating the total discounted reward expected if action $a$ is taken and $\pi_{DRRN}$ is followed thereafter. This procedure is repeated for each valid action $a_i \in A_{valid}(s)$. Action selection is performed by sampling from a softmax distribution over $Q(o, a_i)$. The network is updated by sampling a minibatch of transitions $(o, a, r, o', A_{valid}(s')) \sim \mathcal{D}$ from a prioritized replay memory~\cite{schaul2016} and minimizing the temporal difference error $\delta = r + \gamma * \max_a' Q(o', a') - Q(o, a)$. Rather than performing a separate forward pass for each valid action, we batch valid-actions and perform a single forward pass computing Q-Values for all valid actions.

DRRN uses Jericho's world-change-detection handicap to identify valid actions following Algorithm \ref{alg:valid}. Additionally, it uses Jericho's Load, Save handicap to acquire additional observations without changing the game state: an \emph{inventory} command is issued for the inventory observation $o_{inv}$ and a \emph{look} command is issued for the location description $o_{desc}$.

\begin{figure}[t!]
\begin{subfigure}{.5\textwidth}
    \centering
    \begin{overpic}[width=.7\textwidth]{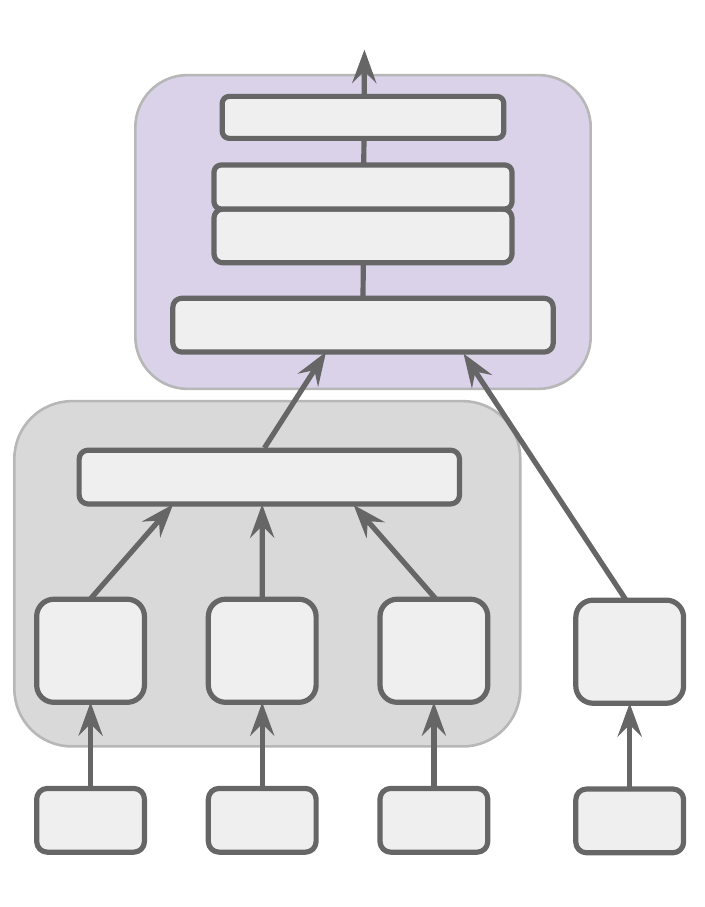} 
    \put (30,97) {$\displaystyle Q(o,a)$}
    \put (35,85) {Linear}
    \put (35,78) {\small ReLU}
    \put (35,72) {Linear}
    \put (34,62) {Concat}
    \put (24,53) {\large$\displaystyle \nu_o$}
    \put (62,46) {\large$\displaystyle \nu_a$}
    \put (15,93) {\Large $\displaystyle\phi_a$}
    \put (23,46) {Concat}
    \put (4,27) {\tiny $\displaystyle GRU_n$}
    \put (23,27) {\tiny $\displaystyle GRU_i$}
    \put (42,27) {\tiny $\displaystyle GRU_d$}
    \put (64,27) {\tiny $\displaystyle GRU_a$}
    \put (1,57) {\Large $\displaystyle\phi_o$}
    \put (6,9) {$o_{nar}$}
    \put (25,9) {$o_{inv}$}
    \put (43,9) {$o_{desc}$}
    \put (68,8) {$a$}
    \end{overpic}
    \caption{DRRN}
    \label{fig:drrn}
\end{subfigure}
\begin{subfigure}{.5\textwidth}
    \centering
    \begin{overpic}[width=.7\textwidth]{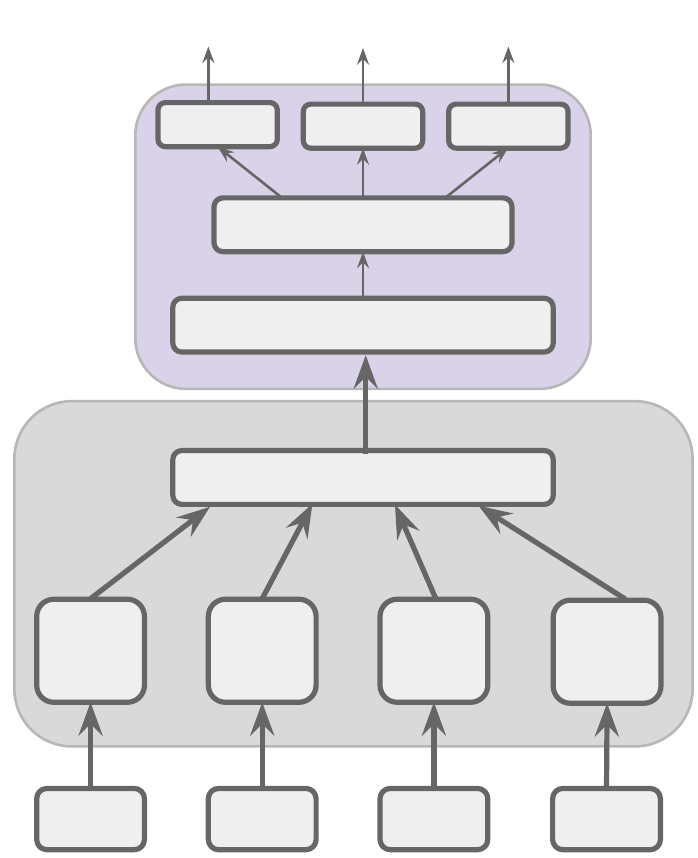} 
    \put (15,96) {\small $\displaystyle Q(o,u)$}
    \put (34,96) {\small $\displaystyle Q(o,p_1)$}
    \put (55,96) {\small $\displaystyle Q(o,p_2)$}
    \put (19,84) {\small Linear}
    \put (36,84) {\small Linear}
    \put (53,84) {\small Linear}
    \put (37,72) {ReLU}
    \put (37,60) {Linear}
    \put (36,50) {\large$\displaystyle \nu_o$}
    \put (36,43) {Concat}
    \put (4,23) {\tiny $\displaystyle GRU_n$}
    \put (25,23) {\tiny $\displaystyle GRU_i$}
    \put (44,23) {\tiny $\displaystyle GRU_d$}
    \put (65,23) {\tiny $\displaystyle GRU_a$}
    \put (6,4) {$o_{nar}$}
    \put (26,4) {$o_{inv}$}
    \put (45,4) {$o_{desc}$}
    \put (66,4) {$a_{t-1}$}
    \end{overpic}
    \caption{TDQN}
    \label{fig:tdqn}
\end{subfigure}%
\caption{\textbf{Models}: The observation encoder $\phi_o$ uses separate GRUs to process the narrative text $o_{nar}$, the players inventory $o_{inv}$, and the location description $o_{desc}$ into a vector $\nu_o$. DRRN uses a separate $GRU_a$ for processing action text into a vector $\nu_a$ which is used to estimate a joint Q-Value $Q(o,a)$ over the observation $o$ and an action $a$. In contrast, Template-DQN estimates Q-Values $Q(o, u)$ for all templates $u \in \mathcal{T}$ and $Q(o, p)$ for all vocabulary $p \in \mathcal{V}$.}
\end{figure}
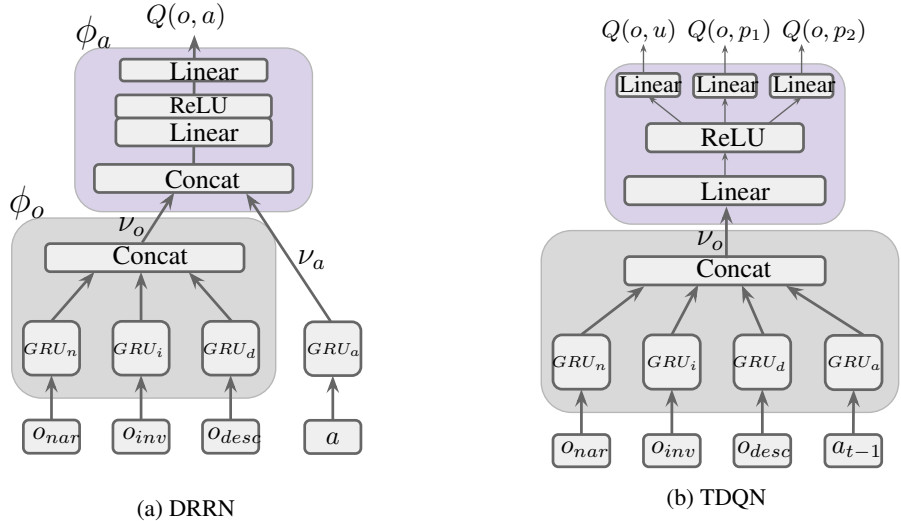

\textbf{Template-DQN}
\cite{narasimhan15} introduced LSTM-DQN, an agent for \emph{parser-based} games that handles the combinatorial action space by generating \emph{verb-object} actions from a pre-defined set verbs and objects. Specifically, LSTM-DQN uses two output layers to estimate Q-Values over possible verbs and objects. Actions are selected by pairing the maximally valued verb with the maximally valued noun.

We introduce Template-DQN (TDQN) which extends LSTM-DQN by incorporating template-based action generation. This is accomplished using three output heads: one for estimating Q-Values over templates $Q(o, u); u \in \mathcal{T}$ and two for estimating Q-Values $Q(o, p_1), Q(o, p_2); p \in \mathcal{V}$ over vocabulary to fill in the blanks of the template. Even in template-action spaces, exploration remains an issue; The largest action space considered in the original LSTM-DQN paper was 222 actions (6 verbs and 22 objects). In contrast, {\em Zork1} has 237 templates with a 697 word vocabulary, yielding an action space of 115 million. Computationally, this space is too large to naively explore as the vast majority of actions will be un-grammatical or contextually irrelevant. To help guide the agent towards valid actions, we introduce a supervised binary-cross entropy loss with targets for each of the templates and vocabulary words appearing in the list of valid actions. The idea behind this loss is to nudge the agent towards the valid templates and words. This loss is evenly mixed with the standard temporal difference error during each update. Due to the supervised loss, TDQN uses the same set of handicaps as DRRN.


\textbf{NAIL}
NAIL~\citep{hausknecht19nail} is the state-of-the-art agent for general interactive fiction game playing~\citep{atkinson18}. Rather than being trained and evaluated on a single game, NAIL is designed to play unseen IF games where it attempts to accumulate as much score as possible in a single episode of interaction. Operating with no handicaps, NAIL employs a set of manually-created heuristics to build a map of objects and locations, reason about which actions were valid or invalid, and uses a web-based language model to decide how to interact with objects. NAIL serves as a point of reference for future work in general IF game playing.

Implementations of DRRN and TDQN are available at \url{https://github.com/microsoft/tdqn}. NAIL is available at \url{https://github.com/microsoft/nail_agent}.

\section{Experiments}
\label{sec:experiments}
We evaluate the agents across a set of thirty-two Jericho-supported games with the aims of 1) showing the feasibility of reinforcement learning on a variety of different IF games, 2) creating a reproducible benchmark for future work, 3) investigating the difference between choice-based and template-based agents, and 4) comparing performance of the general IF game playing agent (NAIL), single-game agents (DRRN and TDQN), and a random agent (RAND) which uniformly sample commands from a set of canonical actions: \{\emph{north, south, east, west, up, down, look, inventory, take all, drop, yes}\}.

\begin{figure}
\centering
\begin{minipage}{.24\linewidth}
  \centering
  \includegraphics[width=\linewidth]{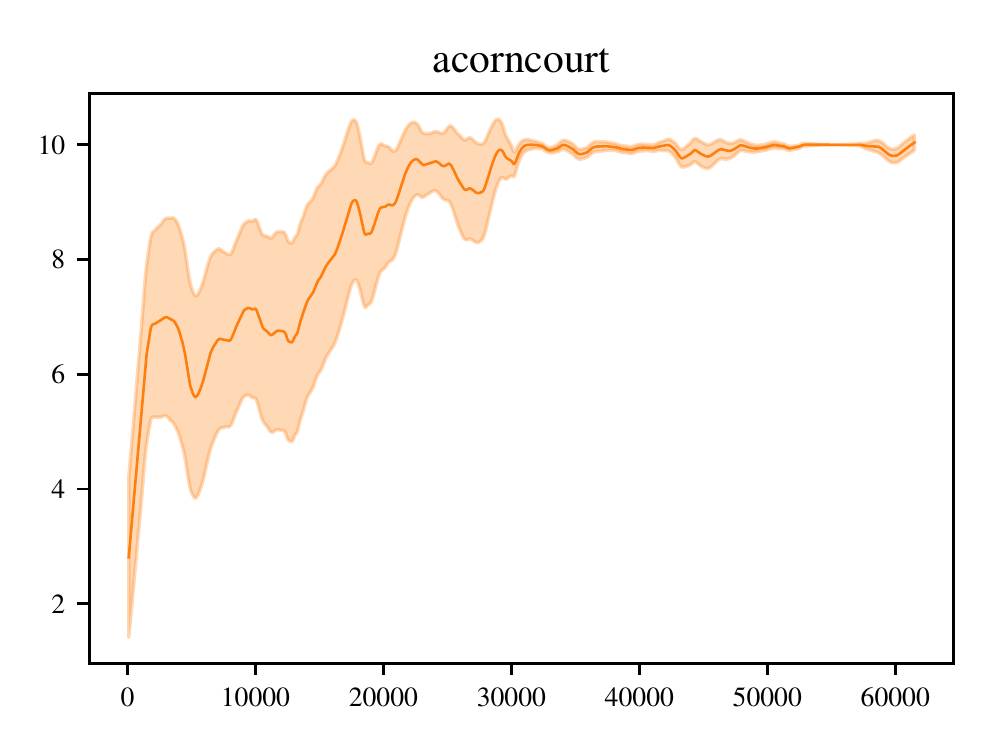}
\end{minipage}
\begin{minipage}{.24\linewidth}
  \centering
  \includegraphics[width=\linewidth]{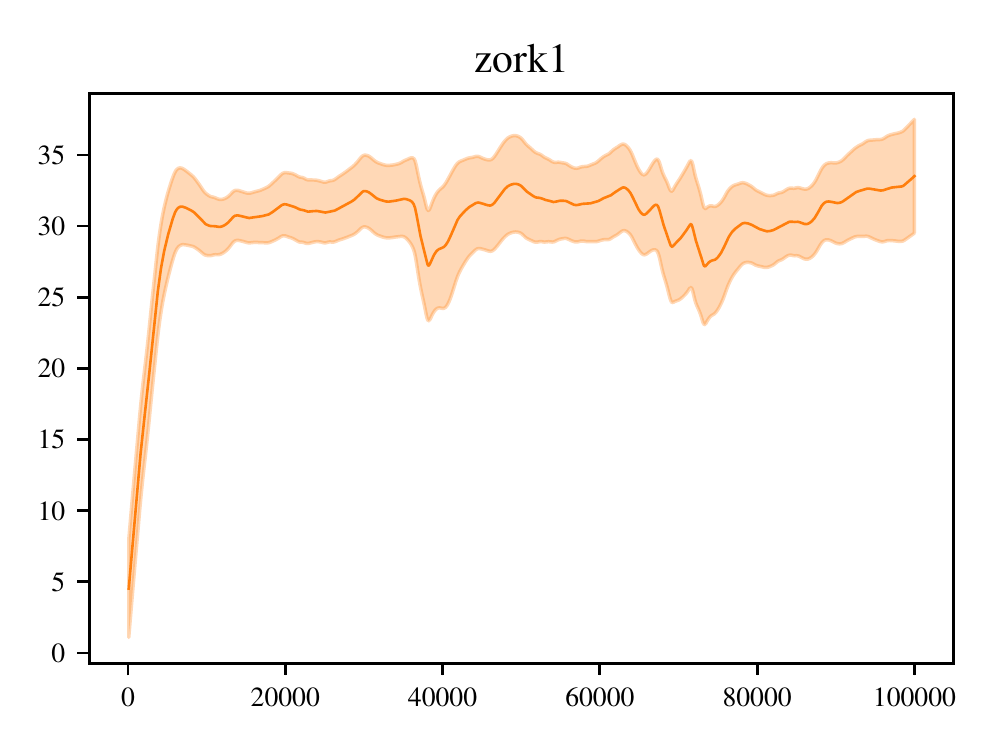}
\end{minipage}
\begin{minipage}{.24\linewidth}
  \centering
  \includegraphics[width=\linewidth]{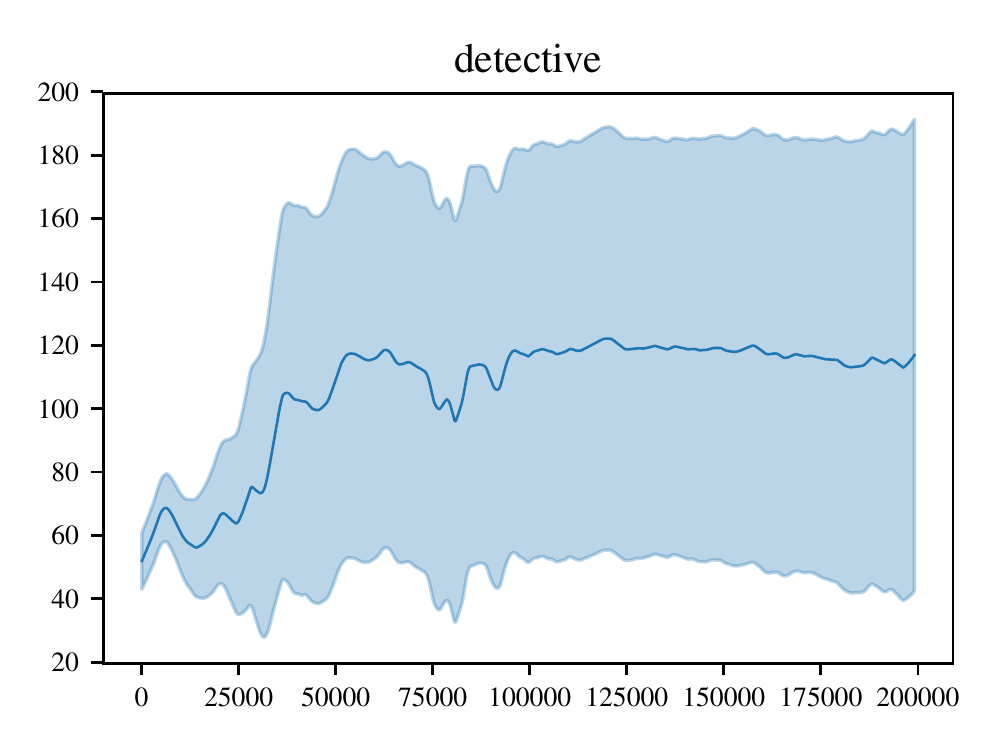}
\end{minipage}
\begin{minipage}{.24\linewidth}
  \centering
  \includegraphics[width=\linewidth]{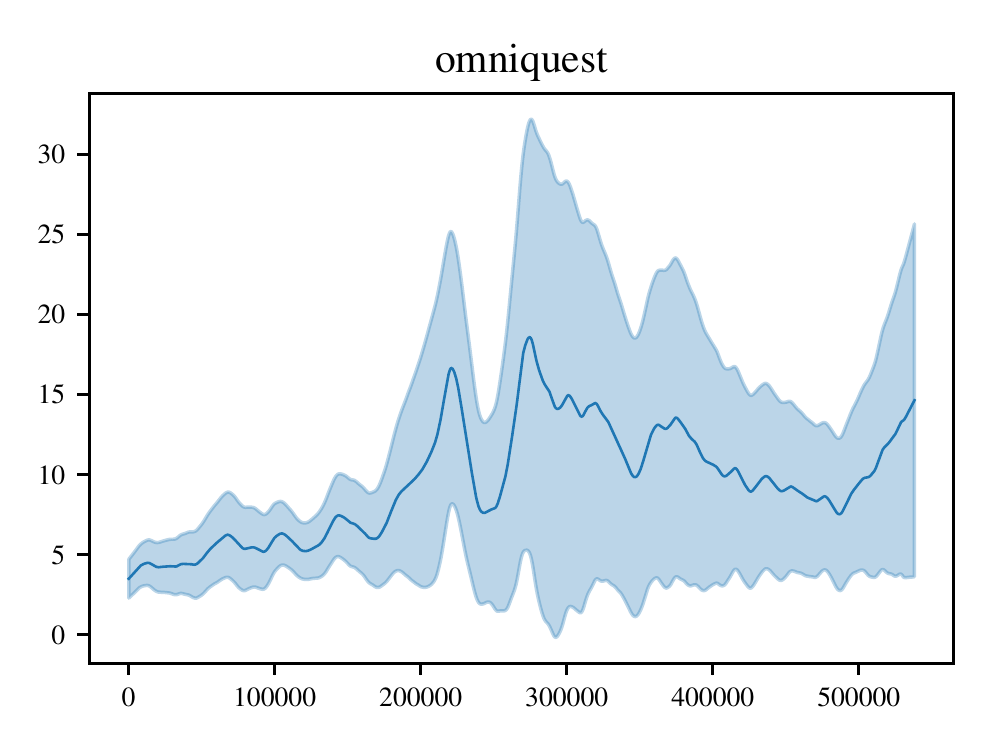}
\end{minipage}
\caption{Episode score as a function of training steps for DRRN (top) and TDQN (bottom). Shaded regions denote standard deviation across five independent runs for each game. Additional learning curves in supplementary material.}
\label{fig:learning_curves}
\end{figure}

Five separate DRRN and TDQN agents were trained for each game. No environment seeds were set so environments remained stochastic throughout. We compute the score for each agent by averaging return over the last hundred episodes of learning. Hyperparameters for TDQN and DRRN were optimized on Zork1 then held fixed across games. Results in Table \ref{tab:results}, supported by learning curves in Figure \ref{fig:learning_curves} show that reinforcement learning is a viable across many of the games. 

In order to quantify overall progress towards story completion, we normalize agent score by maximum possible game score and average across all games. The resulting progress scores are as follows: \textbf{RANDOM $1.8\%$, NAIL $4.9\%$, TDQN $6.1\%$, and DRRN $10.7\%$ completion}. Comparing the different agents, the random agent shows that more than simple navigation and take actions are needed to succeed at the vast majority of games. Comparing DRRN to TDQN highlights the utility of choice-based game playing agents who need only estimate Q-Values over pre-identified valid-actions. In contrast, TDQN needs to estimate Q-Values over the full space of templates and vocabulary words. As a result, we observed that TDQN was more prone to over-estimating Q-Values due to the Q-Learning update computing a max over a much larger number of possible actions.

Comparing NAIL to single-game agents: NAIL performs surprisingly well considering it uses no handicaps, no training period, and plays the game for only a single episode. It should be noted that NAIL was developed on many of the games used in this evaluation, but contains no game-specific information. The fact that the reinforcement learning agents outperform NAIL serves to highlight the difficulty of engineering a general-purpose IF agent as well as the promise of learning policies from data rather than hand-coding them. 

All algorithms have a ways to go before they are solving games of even average difficulty. None of the agents were able to get any score on five of the games. Games like {\em Anchorhead} are highly complex and others pose difficult exploration problems like {\em 9:05} which features only a single terminal reward indicating success or failure at the end of the episode. Additional experiment details and hyperparameters are located in the supplementary material.

\begin{table}[htp]
    \centering
    \small
    \begin{tabular}{r|rr|llll|l}
        Game & \textbf{$|\mathcal{T}|$} & \textbf{$|\mathcal{V}|$} & RAND & NAIL & TDQN & DRRN & MaxScore\\
        \midrule
        905 & 82 & 296 & 0 & 0 & 0 & 0 & 1 \\
        acorncourt & 151 & 343 & 0 & 0 & 1.6 & \textbf{10} & 30 \\
        advent$^\dagger$ & 189 & 786 & 36 & 36 & 36 & 36 & 350 \\
        adventureland & 156 & 398 & 0 & 0 & 0 & \textbf{20.6} & 100 \\
        afflicted & 146 & 762 & 0 & 0 & 1.3 & \textbf{2.6} & 75 \\
        anchor & 260 & 2257 & 0 & 0 & 0 & 0 & 100 \\
        awaken & 159 & 505 & 0 & 0 & 0 & 0 & 50 \\
        balances & 156 & 452 & 0 & \textbf{10} & 4.8 & \textbf{10} & 51 \\
        deephome & 173 & 760 & 1 & \textbf{13.3} & 1 & 1 & 300 \\
        detective & 197 & 344 & 113.7 & 136.9 & 169 & \textbf{197.8} & 360 \\
        dragon & 177 & 1049 & 0 & \textbf{0.6} & -5.3 & -3.5 & 25 \\
        enchanter & 290 & 722 & 0 & 0 & 8.6 & \textbf{20.0} & 400 \\
        gold & 200 & 728 & 0 & 3 & \textbf{4.1} & 0 & 100 \\
        inhumane & 141 & 409 & 0 & 0.6 & \textbf{0.7} & 0 & 90 \\
        jewel & 161 & 657 & 0 & \textbf{1.6} & 0 & 1.6 & 90 \\
        karn & 178 & 615 & 0 & 1.2 & 0.7 & \textbf{2.1} & 170 \\
        library & 173 & 510 & 0 & 0.9 & 6.3 & \textbf{17} & 30 \\
        ludicorp & 187 & 503 & 13.2 & 8.4 & 6 & \textbf{13.8} & 150 \\
        moonlit & 166 & 669 & 0 & 0 & 0 & 0 & 1 \\
        omniquest & 207 & 460 & 0 & 5.6 & \textbf{16.8} & 5 & 50 \\
        pentari & 155 & 472 & 0 & 0 & 17.4 & \textbf{27.2} & 70 \\
        reverb & 183 & 526 & 0 & 0 & 0.3 & \textbf{8.2} & 50 \\
        snacktime & 201 & 468 & 0 & 0 & \textbf{9.7} & 0 & 50 \\
        sorcerer & 288 & 1013 & 5 & 5 & 5 & \textbf{20.8} & 400 \\
        spellbrkr & 333 & 844 & 25 & \textbf{40} & 18.7 & 37.8 & 600 \\
        spirit & 169 & 1112 & \textbf{2.4} & 1 & 0.6 & 0.8 & 250 \\
        temple & 175 & 622 & 0 & 7.3 & \textbf{7.9} & 7.4 & 35 \\
        tryst205 & 197 & 871 & 0 & 2 & 0 & \textbf{9.6} & 350 \\
        yomomma & 141 & 619 & 0 & 0 & 0 & \textbf{0.4} & 35 \\
        zenon & 149 & 401 & 0 & 0 & 0 & 0 & 20 \\
        zork1 & 237 & 697 & 0 & 10.3 & 9.9 & \textbf{32.6} & 350 \\
        zork3 & 214 & 564 & 0.2 & \textbf{1.8} & 0 & 0.5 & 7 \\
        ztuu & 186 & 607 & 0 & 0 & 4.9 & \textbf{21.6} & 100 \\
        
	\end{tabular}
    \caption{\textbf{Raw scores} across Jericho supported games. $|\mathcal{T}|$ denotes the number of templates and $|\mathcal{V}|$ is the size of the parser's vocabulary. $^\dagger$Advent starts with a score of 36.}
    \label{tab:results}
\end{table}

\begin{table}[htp]
    \rowcolors{2}{gray!10}{white}
    \centering
    \scriptsize
    \begin{tabular}{r|rrrrccccc}
        Game & Template Action & Solution & Avg. Steps & Stochastic & Dialog & Darkness & Nonstandard & Inventory \\
        & Space $\times 10^6$ & Length & Per Reward & & & & Actions & Limit \\
        \midrule
        detective & 19 & 51 & 2 &  &  &  &  &  \\ 
        library & 37 & 52 & 5 &  & \checkmark &  & \checkmark &  \\ 
        pentari & 32 & 34 & 5 & \checkmark &  &  & \checkmark &  \\ 
        ztuu & 64 & 84 & 5 & \checkmark &  & \checkmark &  &  \\ 
        loose & 311 & 51 & 5 &  & \checkmark &  & \checkmark &  \\ 
        reverb & 47 & 74 & 6 &  & \checkmark &  &  &  \\ 
        afflicted & 79 & 98 & 7 &  &  & \checkmark &  &  \\ 
        acorncourt & 17 & 17 & 8 &  &  &  &  &  \\ 
        awaken & 37 & 57 & 8 &  &  &  &  &  \\ 
        snacktime & 42 & 34 & 8 &  &  &  &  &  \\ 
        dragon & 182 & 101 & 9 & \checkmark &  & \checkmark & & \checkmark \\ 
        infidel & 102 & 226 & 10 &  &  & \checkmark &  & \checkmark \\ 
        omniquest & 37 & 78 & 13 &  &  & \checkmark &  &  \\
        adventureland & 23 & 170 & 13 & \checkmark &  & \checkmark &  & \checkmark \\ 
        inhumane & 22 & 123 & 14 &  &  &  &  &  \\ 
        temple & 63 & 182 & 20 &  &  &  &  &  \\ 
        905 & 7 & 22 & 22 &  &  &  &  &  \\ 
        moonlit & 70 & 59 & 59 &  &  &  &  &  \\ 
        \midrule
        murdac & 10 & 304 & 4 & \checkmark &  & \checkmark & \checkmark & \checkmark \\ 
        ludicorp & 45 & 364 & 4 &  &  &  & \checkmark & \checkmark \\
        enter & 32 & 102 & 5 &  & \checkmark &  &  &  \\ 
        wishbringer & 312 & 184 & 5 & \checkmark & \checkmark &  &  & \checkmark \\ 
        deephome & 93 & 327 & 6 & \checkmark & \checkmark &  & \checkmark & \checkmark \\ 
        yomomma & 50 & 98 & 6 &  & \checkmark &  &  &  \\ 
        advent & 107 & 277 & 7 & \checkmark &  & \checkmark & \checkmark & \checkmark \\ 
        jewel & 65 & 223 & 8 &  & \checkmark &  &  &  \\ 
        plundered & 242 & 188 & 8 &  &  &  &  &  \\ 
        seastalker & 243 & 204 & 8 &  & \checkmark &  &  &  \\ 
        zork1 & 114 & 400 & 9 & \checkmark &  & \checkmark &  & \checkmark \\ 
        gold & 97 & 345 & 10 & \checkmark & \checkmark &  &  &  \\ 
        zork2 & 112 & 296 & 11 & \checkmark & \checkmark & \checkmark &  & \checkmark \\ 
        balances & 30 & 122 & 12 &  &  & \checkmark & \checkmark &  \\ 
        theatre & 167 & 296 & 14 &  & \checkmark &  &  &  \\ 
        lurking & 207 & 294 & 15 & \checkmark & \checkmark & \checkmark &  & \checkmark \\ 
        karn & 63 & 362 & 17 & \checkmark & \checkmark & \checkmark &  &  \\ 
        zenon & 22 & 83 & 17 &  & \checkmark &  &  &  \\ 
        lostpig & 1118 & 146 & 18 &  & \checkmark & \checkmark &  &  \\ 
        night & 34 & 90 & 18 &  & \checkmark & \checkmark & \checkmark &  \\ 
        hollywood & 216 & 397 & 28 & \checkmark &  & \checkmark &  &  \\ 
        zork3 & 67 & 273 & 39 & \checkmark & \checkmark & \checkmark &  & \checkmark \\ 
        partyfoul & 81 & 56 & 56 & \checkmark &  &  &  &  \\ 
        huntdark & 45 & 67 & 67 &  &  &  &  &  \\ 
        weapon & 42 & 72 & 72 &  & \checkmark &  &  &  \\ 
        \midrule
        curses & 398 & 816 & 9 & \checkmark & \checkmark & \checkmark &  & \checkmark \\ 
        sherlock & 739 & 339 & 10 & \checkmark & \checkmark &  &  & \checkmark \\ 
        sorcerer & 296 & 254 & 11 & \checkmark &  & \checkmark & \checkmark & \checkmark \\ 
        trinity & 2274 & 610 & 12 &  & \checkmark & \checkmark &  & \checkmark \\ 
        tryst205 & 141 & 518 & 12 &  & \checkmark &  & \checkmark & \checkmark \\ 
        spellbrkr & 236 & 412 & 13 & \checkmark &  & \checkmark & \checkmark & \checkmark \\ 
        anchor & 1268 & 531 & 15 & \checkmark & \checkmark & \checkmark &  &  \\ 
        enchanter & 151 & 265 & 15 & \checkmark &  & \checkmark & \checkmark & \checkmark \\ 
        planetfall & 96 & 399 & 18 & \checkmark &  &  &  & \checkmark \\ 
        hhgg & 277 & 361 & 20 & \checkmark & \checkmark & \checkmark &  & \checkmark \\ 
        ballyhoo & 310 & 416 & 21 &  & \checkmark &  &  & \checkmark \\ 
        spirit & 195 & 1264 & 21 &  & \checkmark & \checkmark & \checkmark & \checkmark \\ 
        cutthroat & 185 & 336 & 34 & \checkmark &  & \checkmark & \checkmark & \checkmark \\ 
	\end{tabular}
	\caption{\small\textbf{Categories of Difficulty}: Possible games (top): Reinforcement Learning agents have a credible chance to solve in the short term. Difficult games (middle): current agents can make progress on, but will require significant advances to solve. Extreme games (bottom): highly challenging for humans and inconceivable for current RL agents to solve. \textit{Action space} is the size of the template-action space $=|\mathcal{T}\mathcal{V}^2|$ in millions of actions. \textit{Solution Length} is the number of steps needed for the walkthrough to complete the game. \textit{Average Steps per Reward} is the average number of steps the agent needs to take between getting rewards. \textit{Stochastic} indicates if the game's solution changes with the random seed. \textit{Dialog} indicates that it's necessary to speak with or issue commands to another character such as "john, turn off light." \textit{Darkness} is a common trope in IF games that limits explorable areas until a light source such as a lantern or torch is found. \textit{Nonstandard Actions} are actions that wouldn't be found in an English dictionary - such as casting a Frotz spell. \textit{Inventory Limit} restricts the number of items carried by the player and requires decisions about which items to discard and keep.}
	\label{tab:games}
\end{table}

\section{Notable Games}
Jericho supports a variety of games, covering a diverse set of structures and genres.
These games provide different challenges from the perspective of reinforcement learning based agents.
Table \ref{tab:games} categorizes the games into three difficulty tiers:


\textbf{Possible Games}
are games that learning agents have a credible chance to solve in the near future.
These games fall on the lower end on the difficulty spectrum and serve as good initial testbeds for developing new agents because they feature frequent rewards, are solvable by basic navigation and interaction actions, and are largely devoid of the complexities like dialog and inventory limits.
{\em Detective} is one of the easiest games and existing agents can finish the game albeit without collecting all possible points.
{\em Acorncourt} also serves as a sanity check, albeit a more difficult one---with the DRRN outperforming all other agents.
This game, although short, requires a high proportion of complex actions, which makes action generation---as in the case of TDQN---more difficult.
{\em Omniquest} is a dungeon-crawler style game where TDQN outperforms the rest of the agents due to the relatively smaller number of valid templates as compared to valid actions---i.e. many valid actions come from the same template---TDQN has a smaller effective search space than DRRN.

\textbf{Difficult Games}
feature sparser rewards, more complex puzzles and interactions, and require more steps to solve. 
Current learning agents are able to collect some score on these games, but we expect that new insights and algorithms will be necessary to solve these games.
Games like {\em Zork1} typify this category: existing agents can effectively explore the opening areas, but quickly plateau when going underground due to challenges like darkness, mazes, and fighting.
For example, darkness requires the player to have a light source such as the lantern equipped and activated in order to move.
This challenge is further complicated by limited inventory space, which prevents the player from picking up all the items they encounter and instead forces them to reason about which items are essential and which can be discarded.

\textbf{Extreme Games}
are those featuring extensive puzzles often involving high-level, creative problem solving that challenges even skilled human players.
These games are the most complex in terms of the length of required solutions, and difficulty of puzzles, and non-standard nature of actions.
A prime example is {\em Anchorhead}, a Lovecraftian horror complex enough that it has been the subject of prior work on cognition in script learning and drama management~\citep{Giannatos2011}.
Its puzzles features physical reasoning such as arranging trash cans in an alleyway in order to access a second floor window, as well as long-range dependencies between when hints are given and when they're needed.
Our agents are yet to accumulate any score on this game.
Another category of extreme games is the Enchanter Trilogy (\textit{Enchanter}, \textit{Sorcerer}, and \textit{Spellbreaker}) which was developed by Infocom as a followup to the Zork triology. 
In addition to larger vocabularies, these games require the player to cast spells like \textit{Frotz} and \textit{Rezrov}, the effects of which must be learned by playing the game. 
We expect that such non-standard actions will require agents adaptively learn the names and effects of these spells and reason how to use them on the fly.

\section{Future Work}
\textbf{Unsupervised learning}: DRRN and TDQN agents were trained and evaluated on individual games. While sufficient for a proof-of-concept, this evaluation falls short of demonstrating truly general IF game playing. To this end, it's valuable to evaluate agents on a separate set of games which they have not been trained on. In the Jericho framework, we propose to use the set of Jericho supported games as a \emph{test set} and the larger set of unsupported games available at \url{https://github.com/BYU-PCCL/z-machine-games} as the \emph{training set}. In this paradigm, it's necessary have a strong unsupervised learning component to guide the agent's exploration and learning since unsupported games do not provide rewards, and in fact many IF games do not have scores. We hypothesize that \emph{surrogate reward functions}, like novelty-based rewards~\citep{bellemare16count,pathak17}, will be useful for discovering locations, successful interactions, and objects.

\textbf{Better Template-based Agents}: There are several directions for creating better template-based agents by improving on the limitations of TDQN: When generating actions, TDQN assumes independence between templates and vocabulary words. To understand the problem with this assumption consider the templates \emph{go \_} and \emph{take \_} and the vocabulary words \emph{north, apple}. Independently, \emph{take} and \emph{north} may have the highest Q-Values but together yield the invalid action \emph{take north}. Conditional generation of words based on the chosen template may improve the quality of TDQN's actions.


Second, recent work on transformer-based neural architectures has yielded impressive gains in many NLP tasks~\citep{devlin18}, including text-adventure game dialogues~\citep{urbanek2019light}. We expect these advances may be applicable to human-made IF games, but will need to be adapted from a supervised training regime into reinforcement learning.

\section{Conclusion}
Interactive Fiction games are rich narrative adventures that challenge even skilled human players. In contrast to other video game environments, IF games stress natural language understanding and commonsense reasoning, and feature combinatorial action spaces. To aid in the study of these environment, we introduce Jericho, an experimental platform with the key of feature of extracting game-specific action templates and vocabulary. Using these features, we proposed a novel template-based action space which serves to reduce the complexity of full scale language generation. Using this space, we introduced the Template-DQN agent TDQN, which generates actions first by selecting a template then filling in the blanks with words from the vocabulary.

We evaluated TDQN against DRRN, a choice-based agent, and NAIL, a general IF agent. The fact that DRRN outperformed TDQN illustrates the difficulty of language generation, even with templates. However, TDQN showed encouraging progress, outperforming both NAIL and the random agent. However, in many ways these agents represent very different training paradigms, sets of assumptions, and levels of handicap. Rather than comparing agents against each other we aim to provide these scores as benchmark results for future work in all three categories of IF game playing. We believe Jericho can help the community propel research on language understanding agents and expect these environments can serve the community as benchmarks for years to come.

\section*{Acknowledgements} Many thanks to Adith Swaminathan for keen insights, to Greg Yang for the idea of IF games as a domain of interest, to Ricky Loynd and Alekh Agarwal for unwavering support.

\bibliography{main.bib}
\bibliographystyle{abbrvnat}

\newpage
\begin{appendices}

\section{Experiment Details}
Episodes are terminated after 100 valid steps or game over/victory. Since agents that frequently decode invalid actions would never make it more than few steps past the start of the game, we only count valid-actions against the hundred step limit. DRRN and TDQN are trained individually on each game. DRRN was trained for 100,000 updates, using data collected from 16 environments in parallel. This corresponds to 1.6 million environment interactions. TDQN was trained for 1 million environment steps using a single environment. Hyperparameters for DRRN and TDQN were tuned on the game of Zork1 and held constant across all other games. Both the random agent and NAIL are given 1000 total environment steps. Final reported scores are an average over 5 runs of each algorithm.

\begin{figure*}[h]
\begin{minipage}{.19\textwidth}
  \centering
  \includegraphics[width=\linewidth]{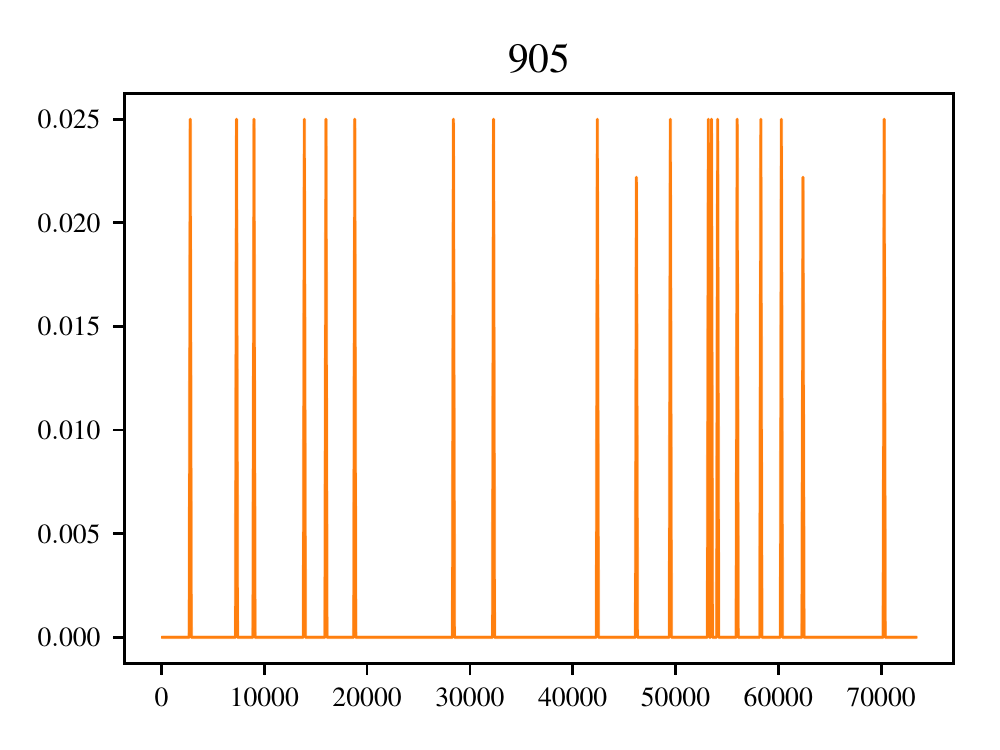}
\end{minipage}
\begin{minipage}{.19\textwidth}
  \centering
  \includegraphics[width=\linewidth]{figures/plots/drrn_acorncourt.pdf}
\end{minipage}
\begin{minipage}{.19\textwidth}
  \centering
  \includegraphics[width=\linewidth]{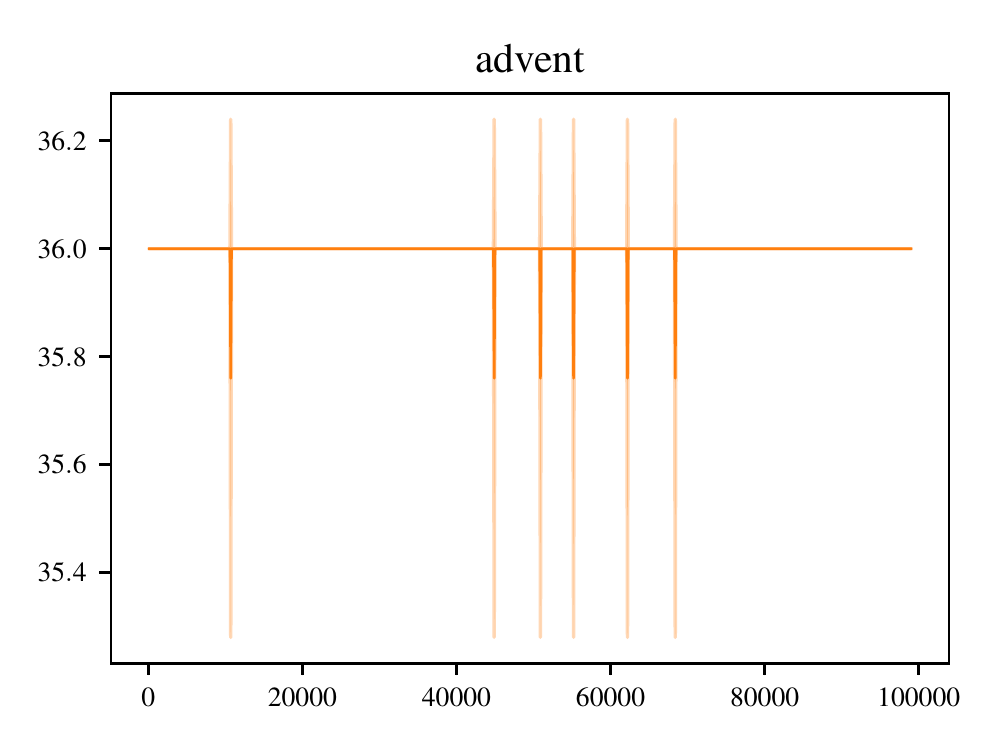}
\end{minipage}
\begin{minipage}{.19\textwidth}
  \centering
  \includegraphics[width=\linewidth]{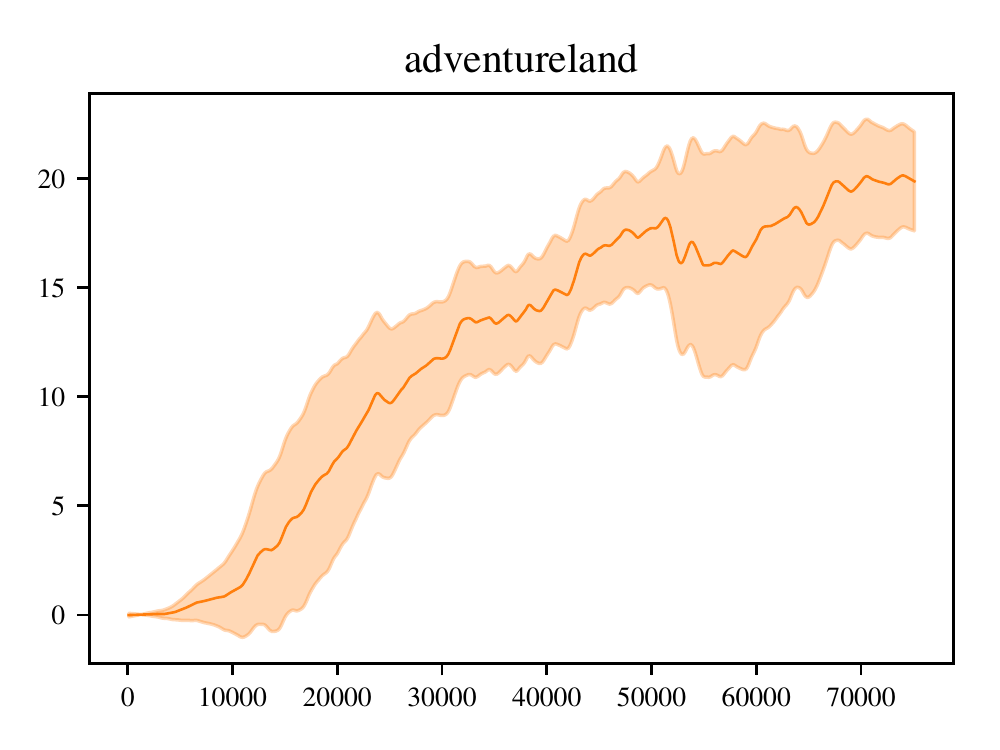}
\end{minipage}
\begin{minipage}{.19\textwidth}
  \centering
  \includegraphics[width=\linewidth]{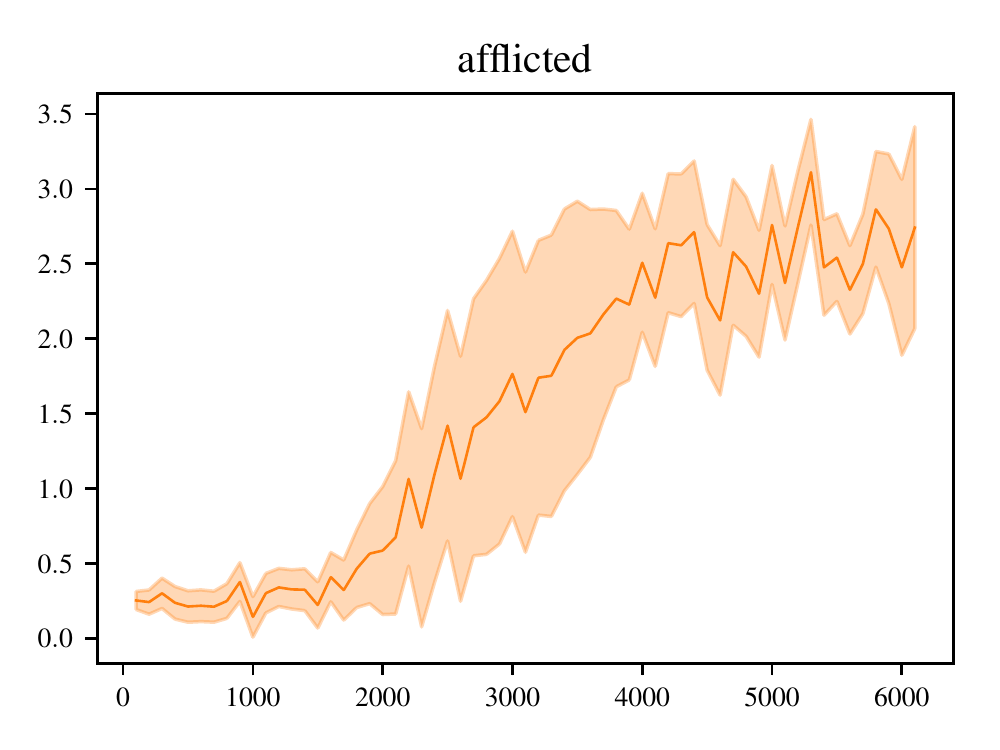}
\end{minipage}
\begin{minipage}{.19\textwidth}
  \centering
  \includegraphics[width=\linewidth]{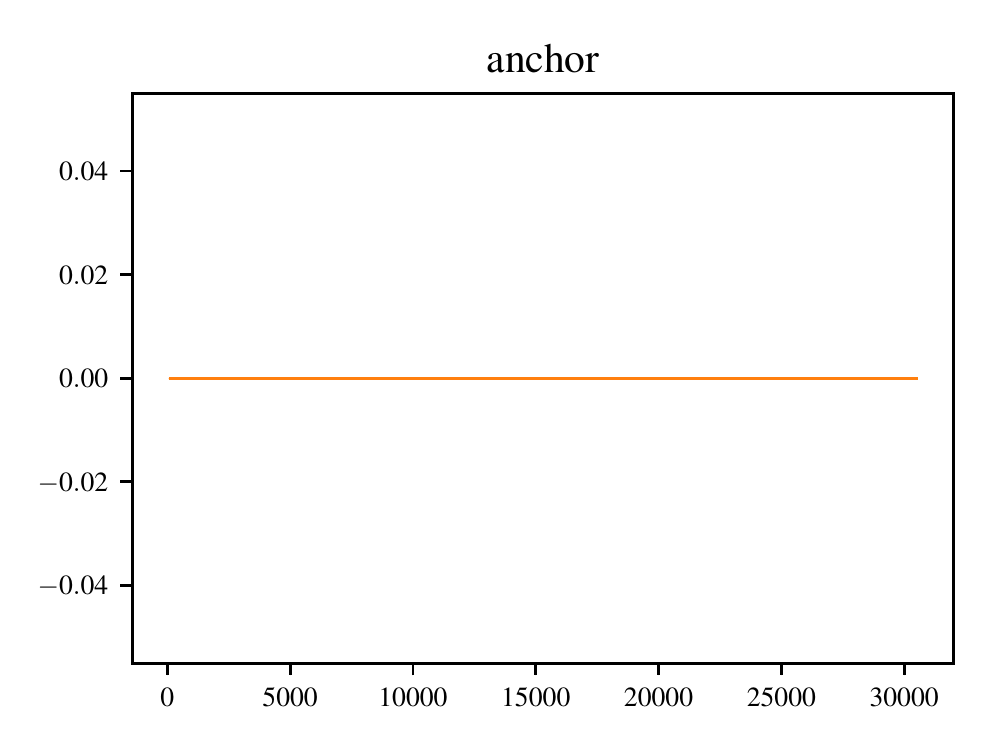}
\end{minipage}
\begin{minipage}{.19\textwidth}
  \centering
  \includegraphics[width=\linewidth]{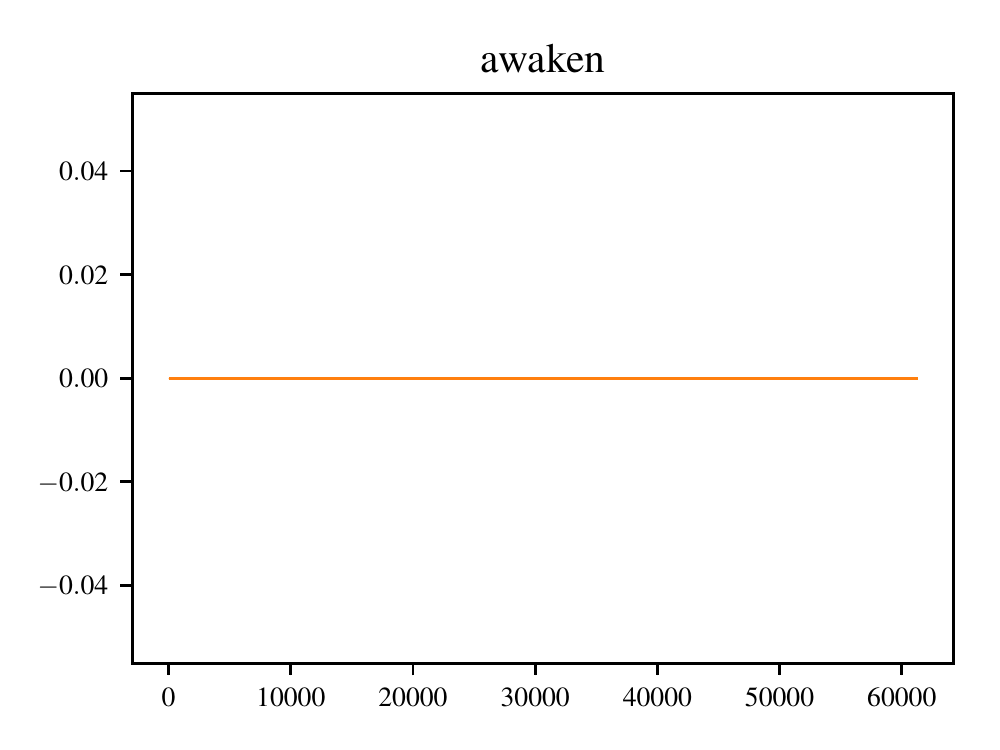}
\end{minipage}
\begin{minipage}{.19\textwidth}
  \centering
  \includegraphics[width=\linewidth]{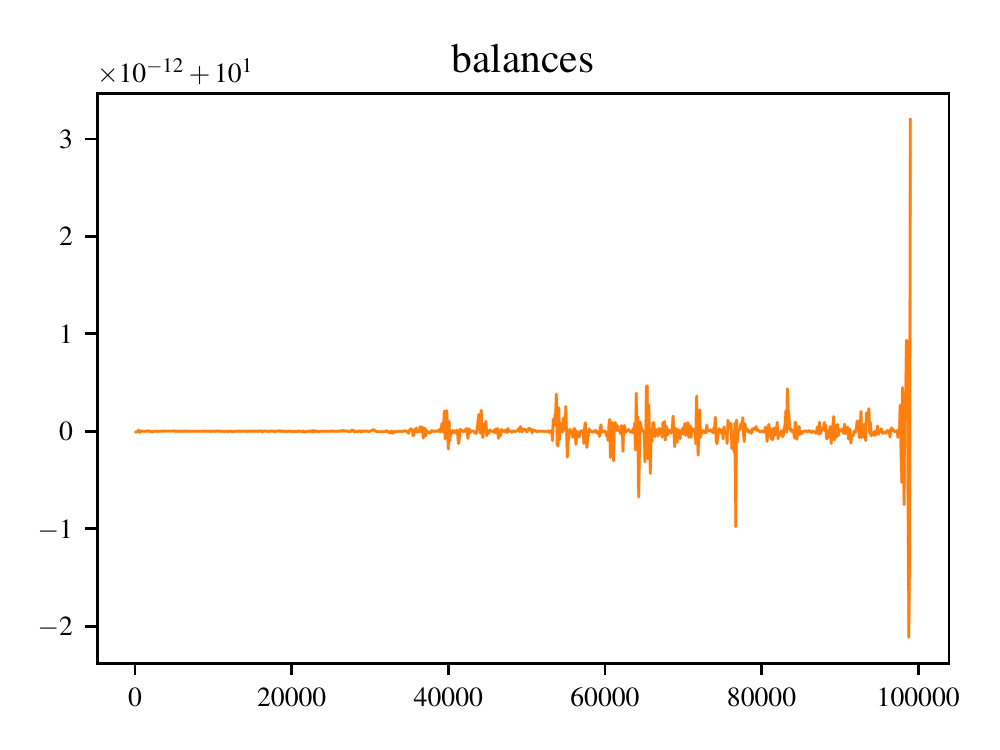}
\end{minipage}
\begin{minipage}{.19\textwidth}
  \centering
  \includegraphics[width=\linewidth]{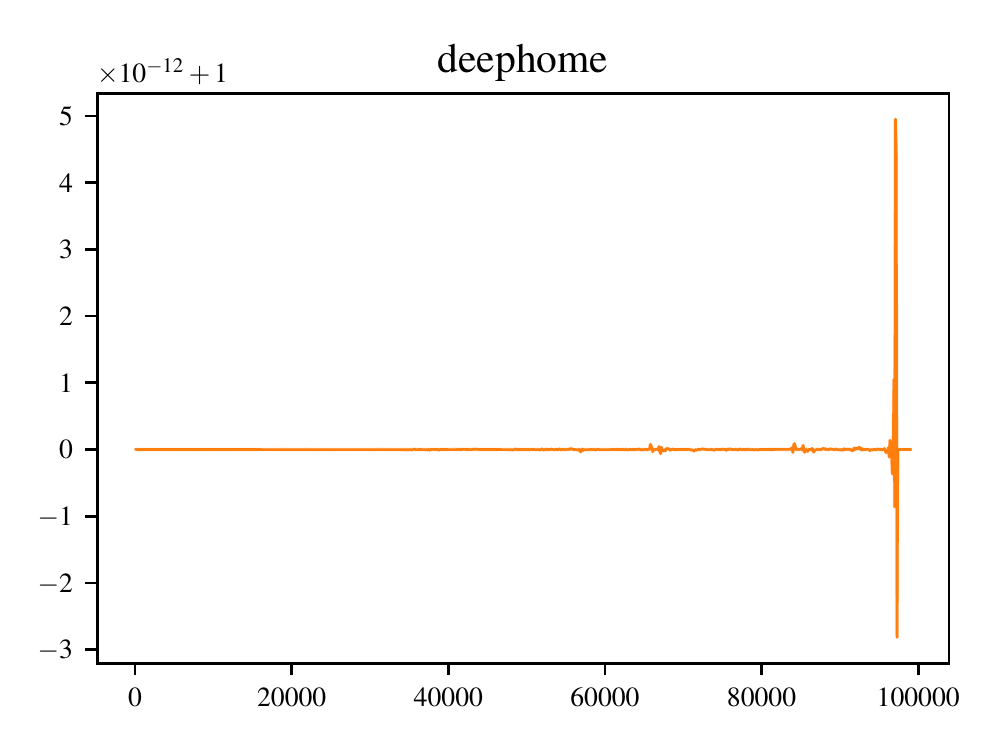}
\end{minipage}
\begin{minipage}{.19\textwidth}
  \centering
  \includegraphics[width=\linewidth]{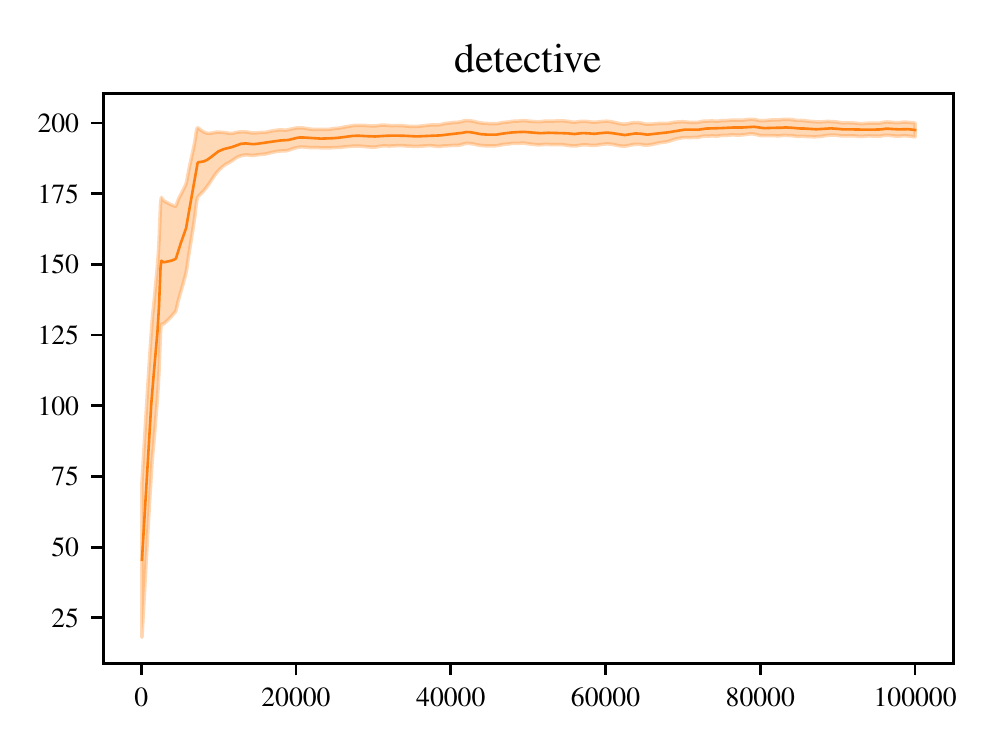}
\end{minipage}
\begin{minipage}{.19\textwidth}
  \centering
  \includegraphics[width=\linewidth]{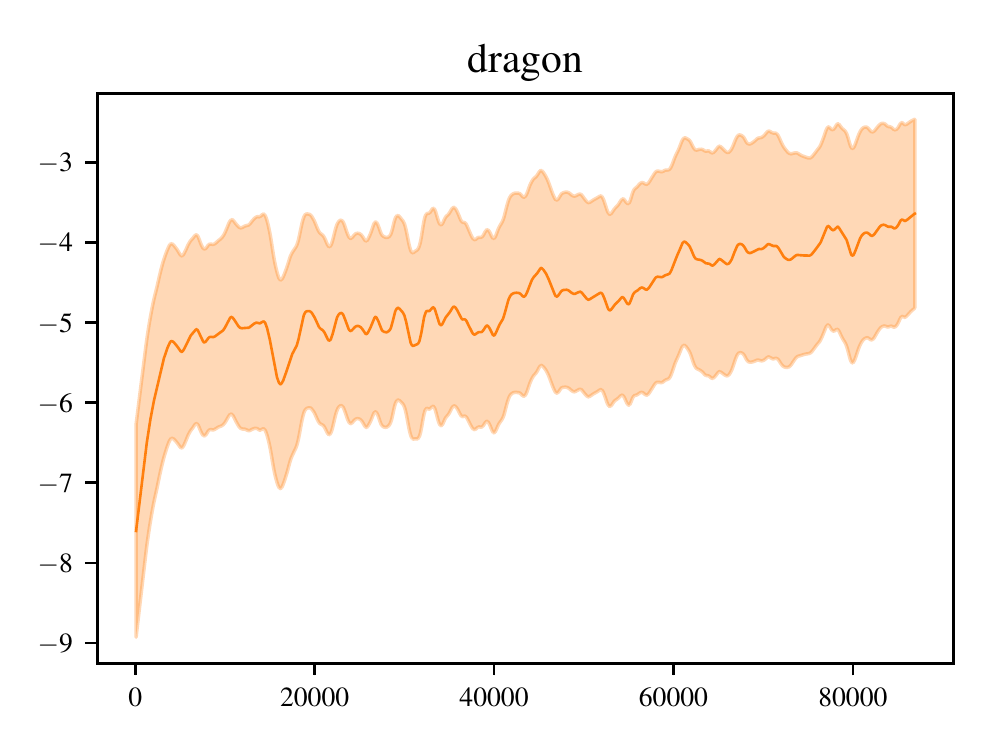}
\end{minipage}
\begin{minipage}{.19\textwidth}
  \centering
  \includegraphics[width=\linewidth]{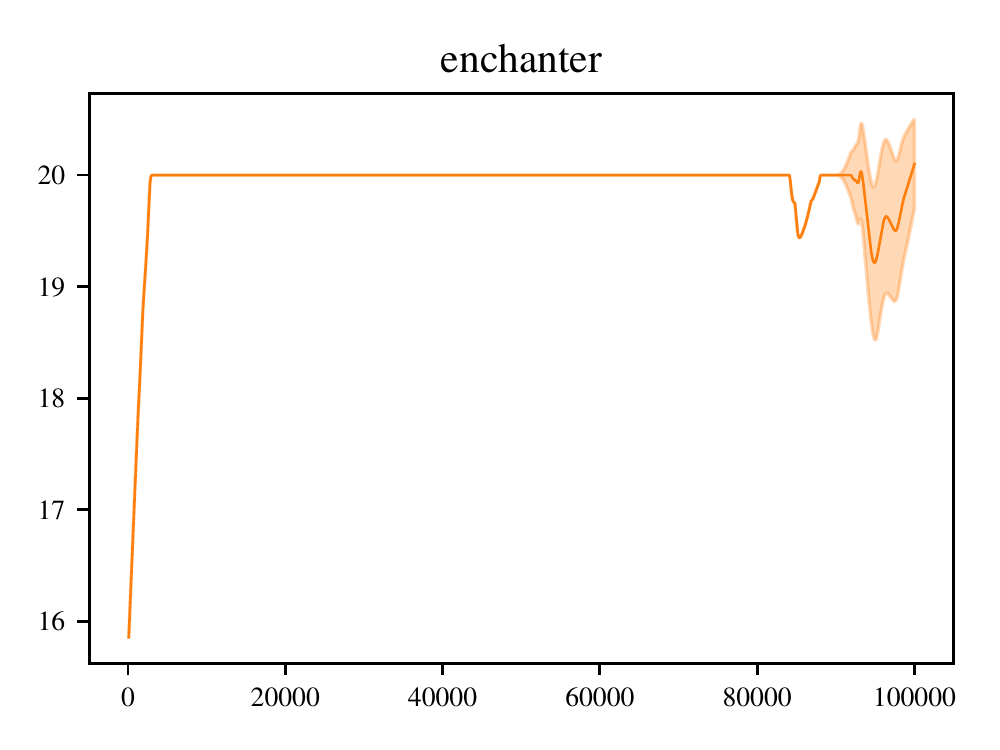}
\end{minipage}
\begin{minipage}{.19\textwidth}
  \centering
  \includegraphics[width=\linewidth]{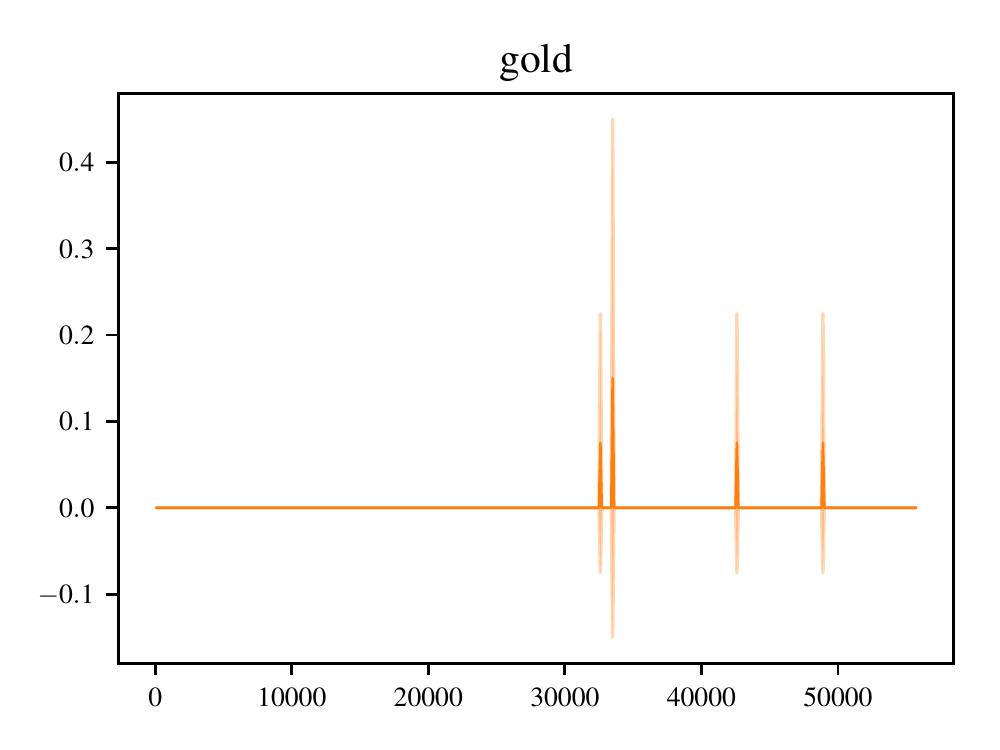}
\end{minipage}
\begin{minipage}{.19\textwidth}
  \centering
  \includegraphics[width=\linewidth]{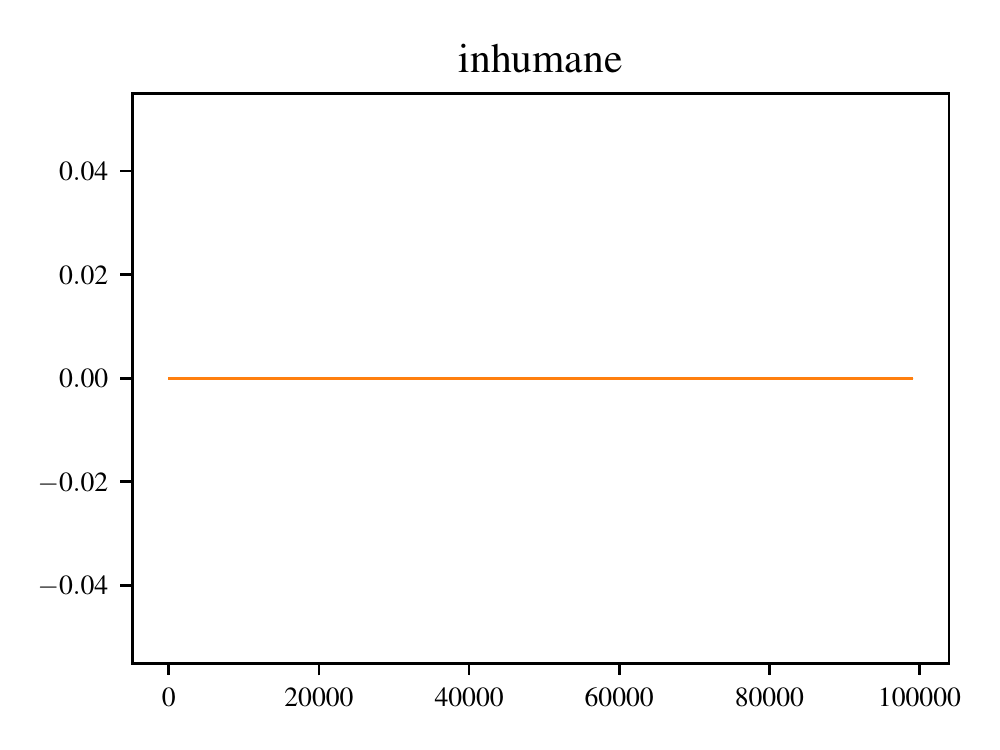}
\end{minipage}
\begin{minipage}{.19\textwidth}
  \centering
  \includegraphics[width=\linewidth]{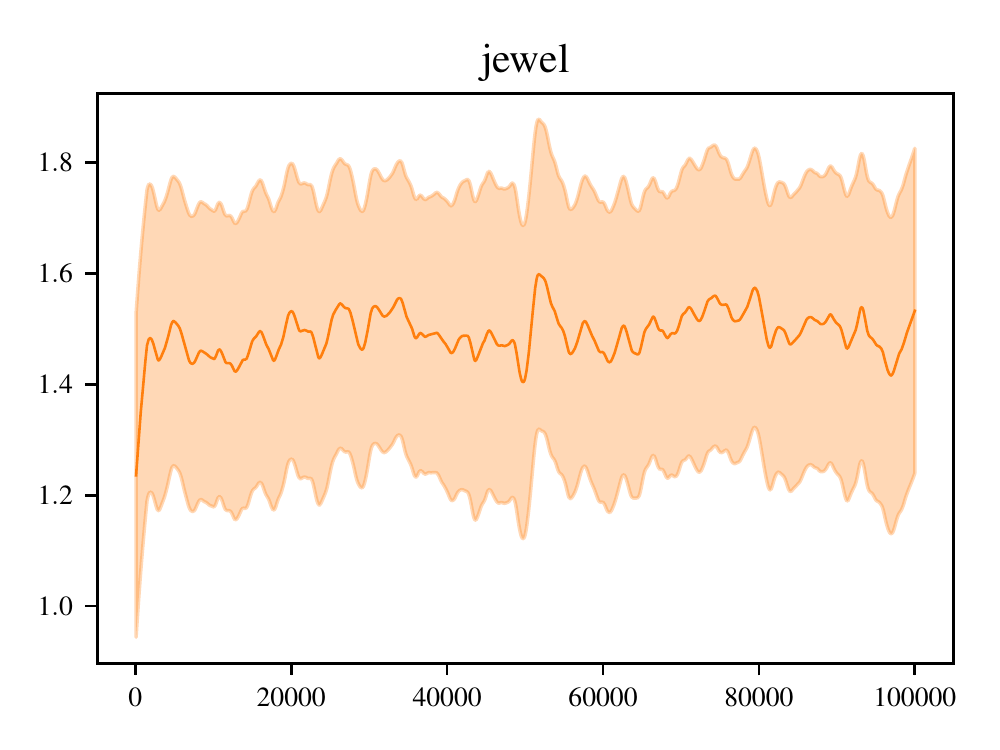}
\end{minipage}
\begin{minipage}{.19\textwidth}
  \centering
  \includegraphics[width=\linewidth]{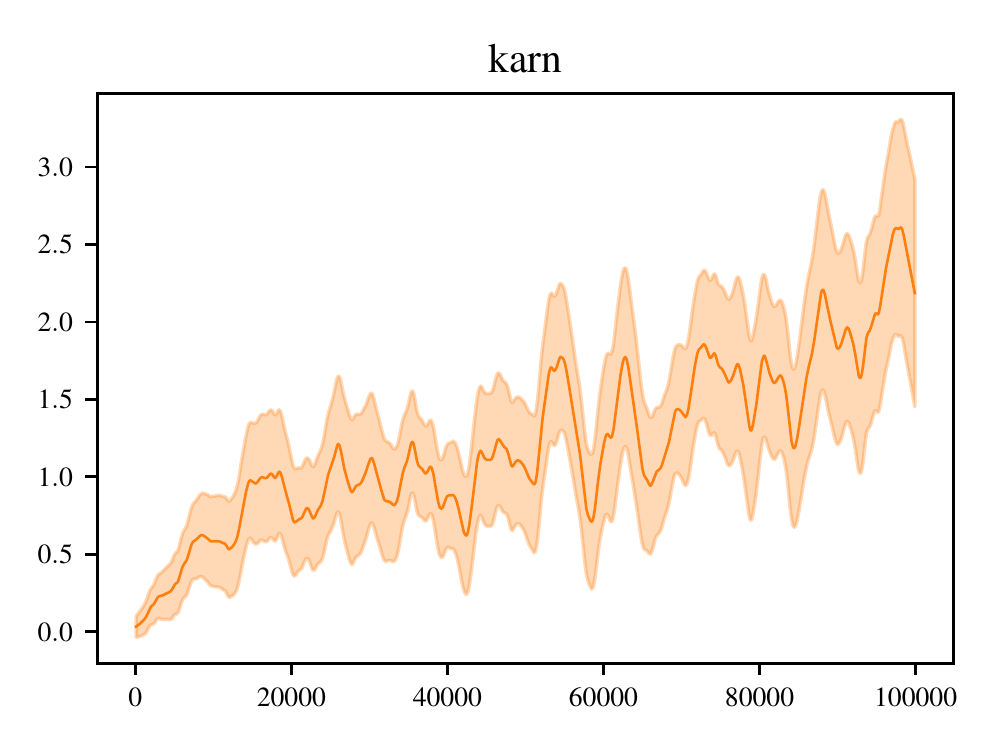}
\end{minipage}
\begin{minipage}{.19\textwidth}
  \centering
  \includegraphics[width=\linewidth]{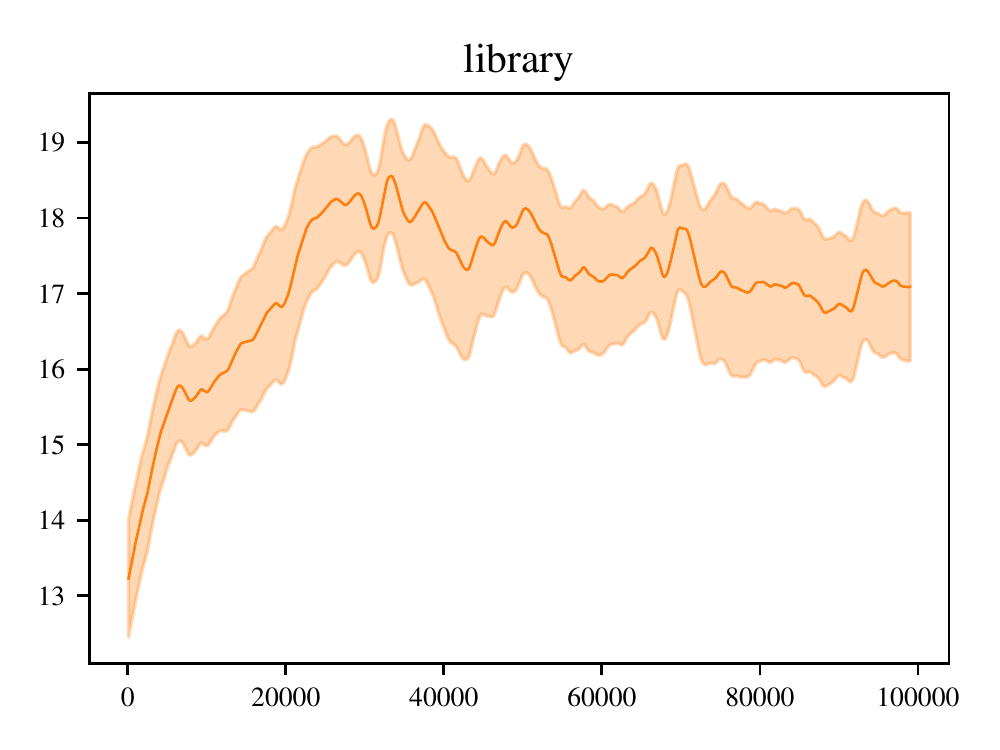}
\end{minipage}
\begin{minipage}{.19\textwidth}
  \centering
  \includegraphics[width=\linewidth]{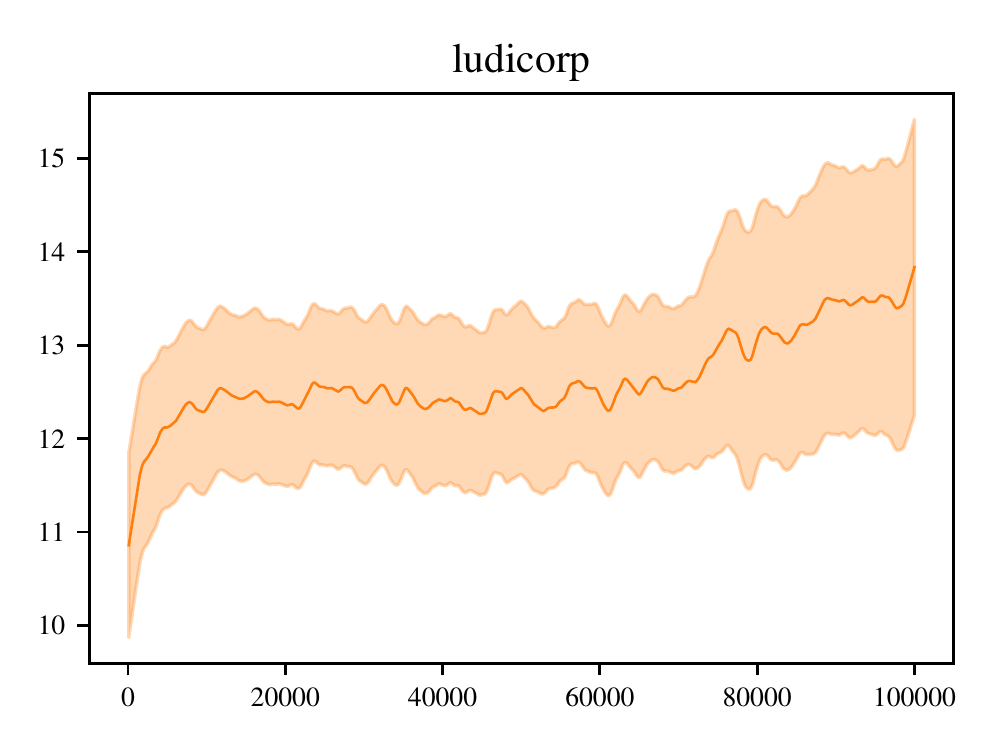}
\end{minipage}
\begin{minipage}{.19\textwidth}
  \centering
  \includegraphics[width=\linewidth]{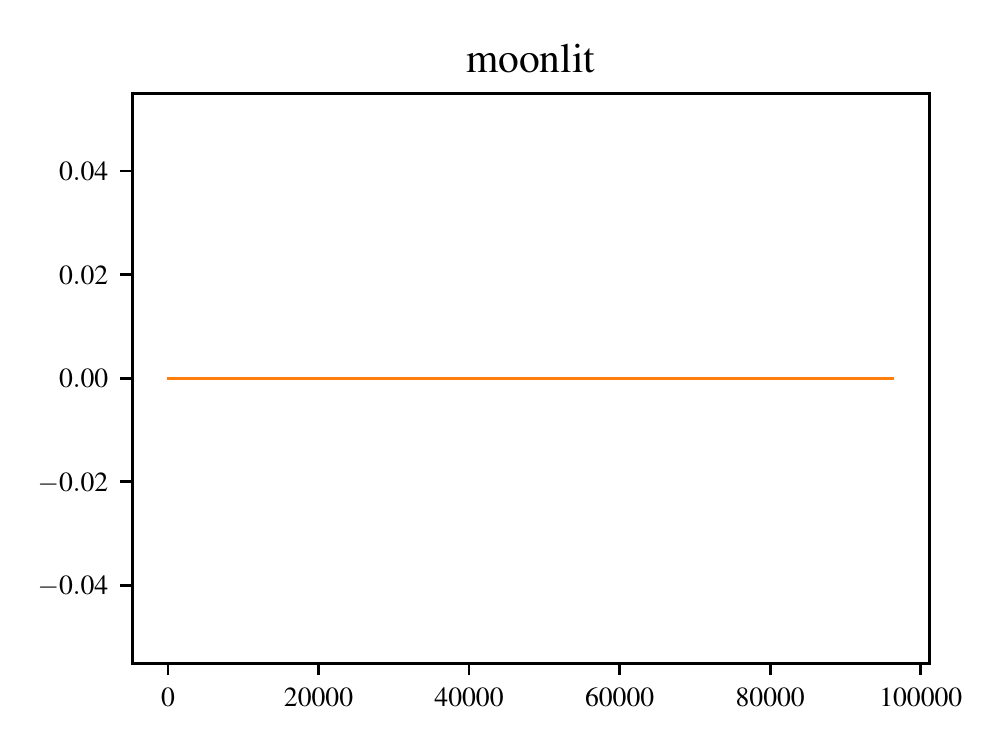}
\end{minipage}
\begin{minipage}{.19\textwidth}
  \centering
  \includegraphics[width=\linewidth]{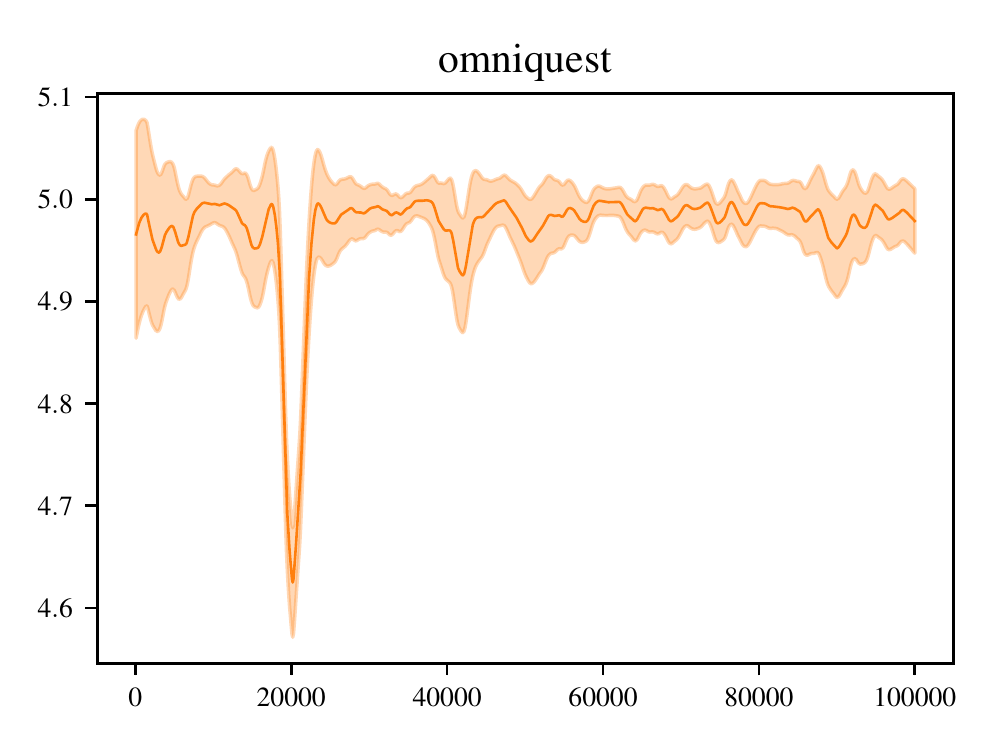}
\end{minipage}
\begin{minipage}{.19\textwidth}
  \centering
  \includegraphics[width=\linewidth]{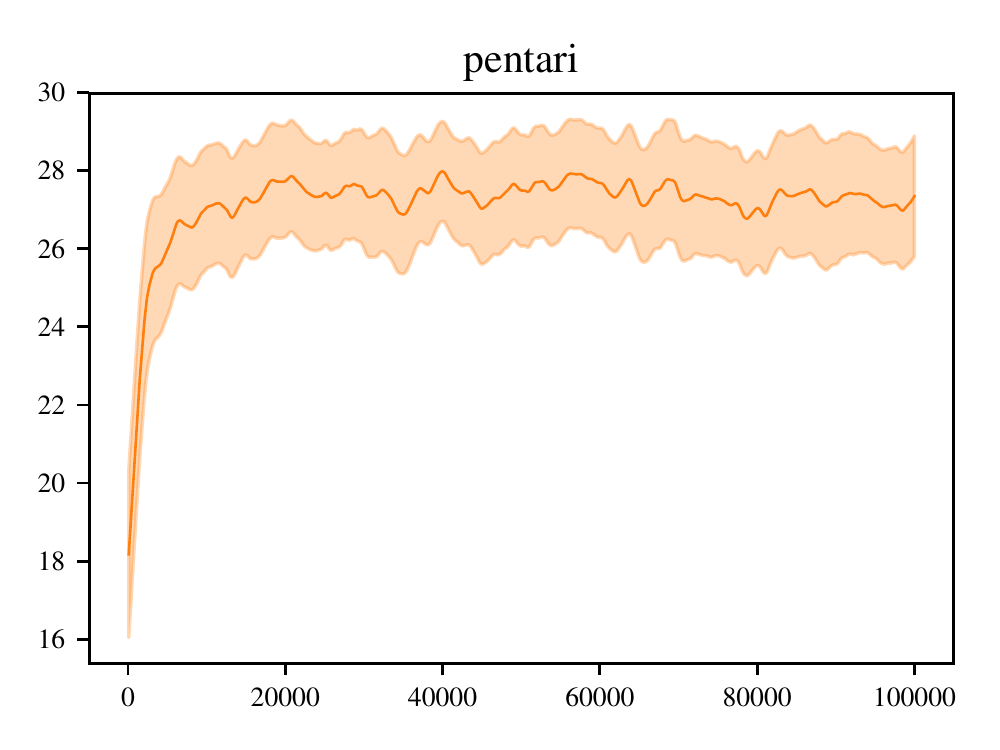}
\end{minipage}
\begin{minipage}{.19\textwidth}
  \centering
  \includegraphics[width=\linewidth]{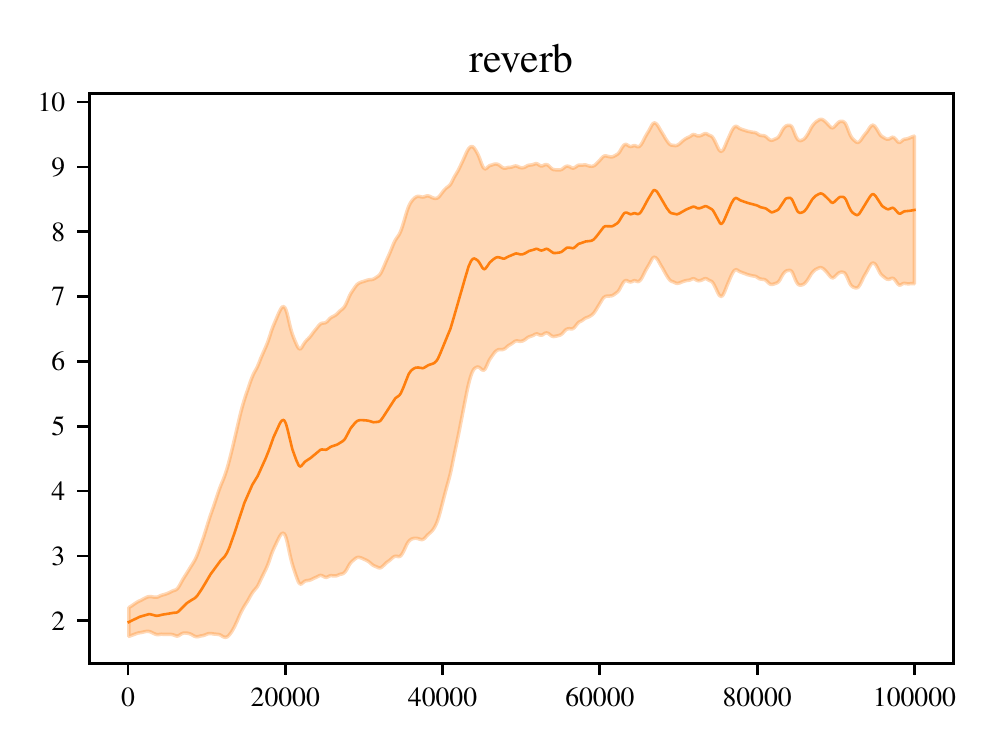}
\end{minipage}
\begin{minipage}{.19\textwidth}
  \centering
  \includegraphics[width=\linewidth]{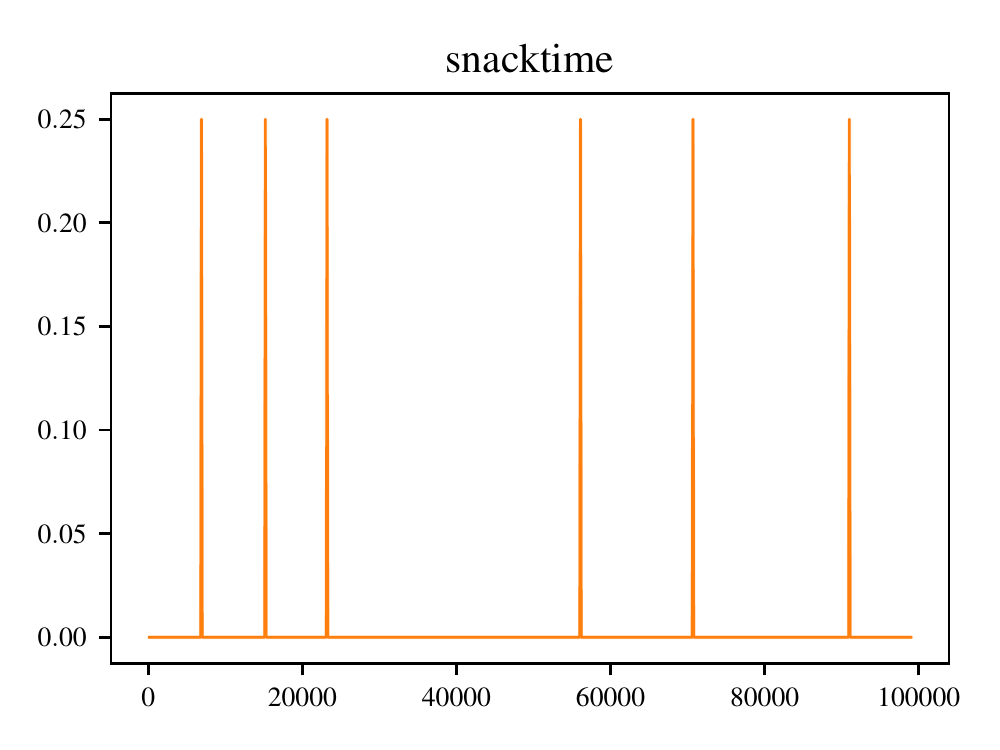}
\end{minipage}
\begin{minipage}{.19\textwidth}
  \centering
  \includegraphics[width=\linewidth]{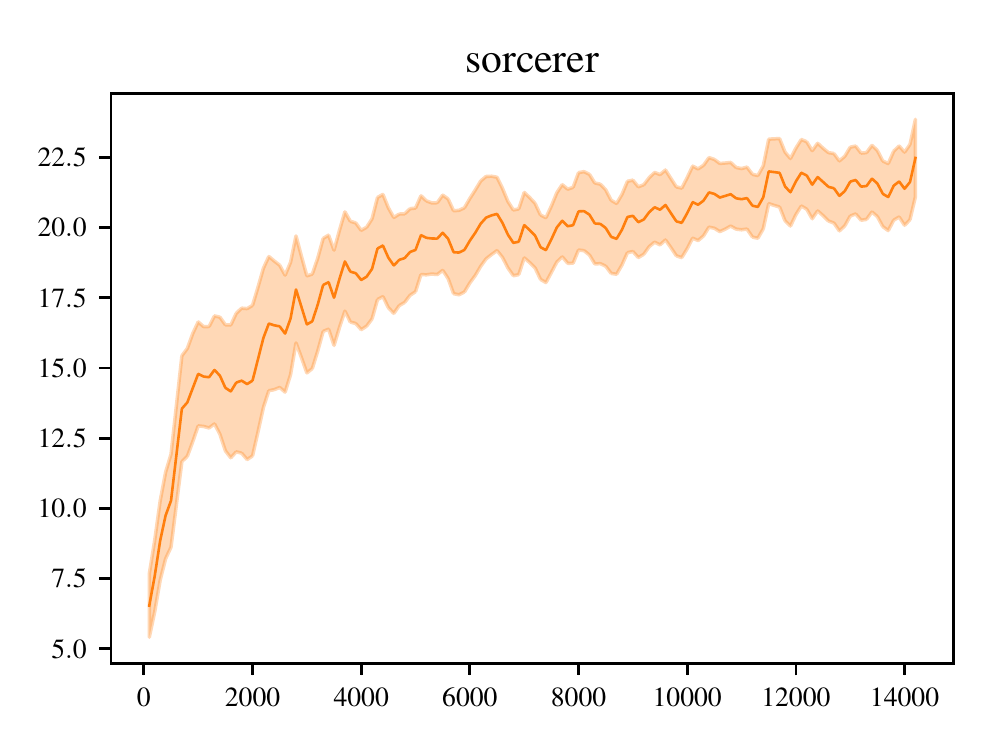}
\end{minipage}
\begin{minipage}{.19\textwidth}
  \centering
  \includegraphics[width=\linewidth]{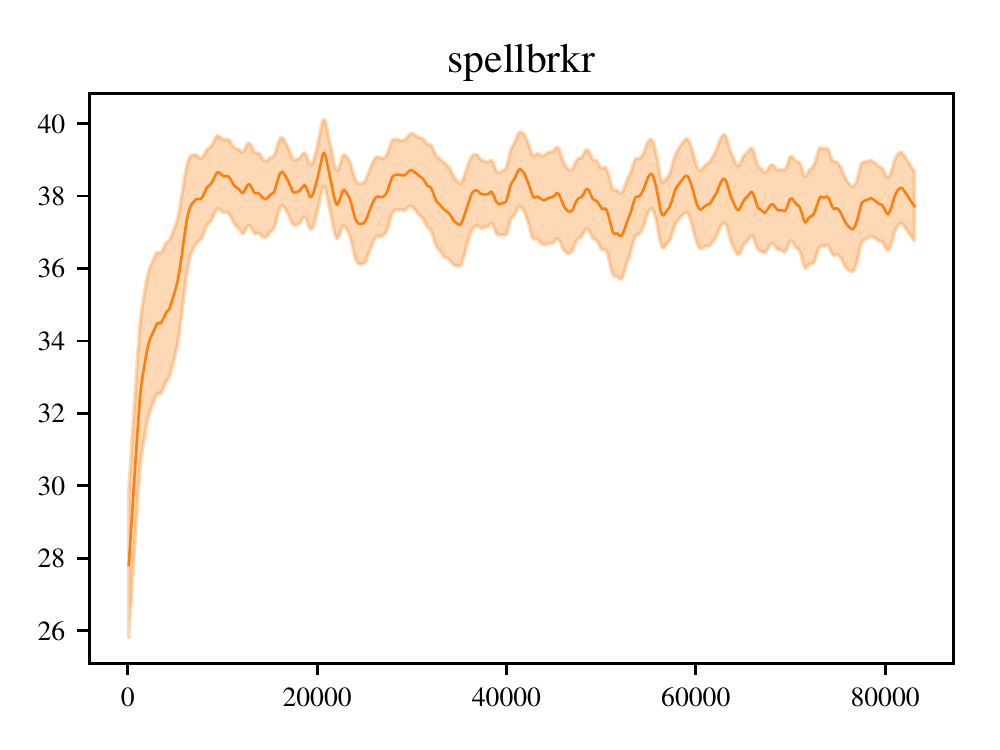}
\end{minipage}
\begin{minipage}{.19\textwidth}
  \centering
  \includegraphics[width=\linewidth]{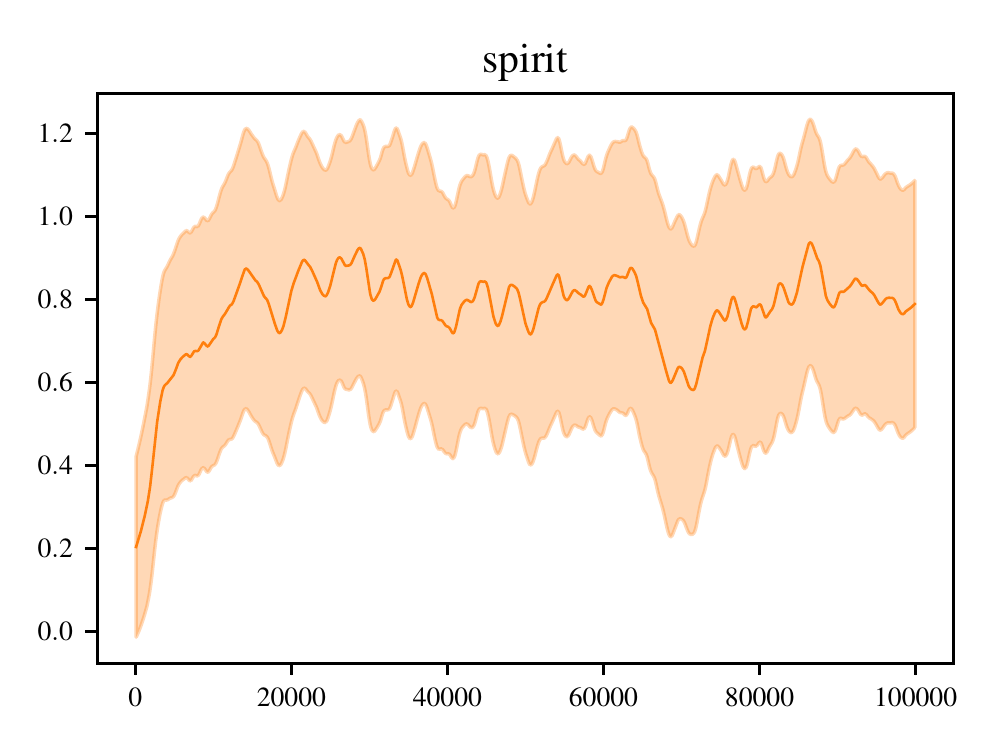}
\end{minipage}
\begin{minipage}{.19\textwidth}
  \centering
  \includegraphics[width=\linewidth]{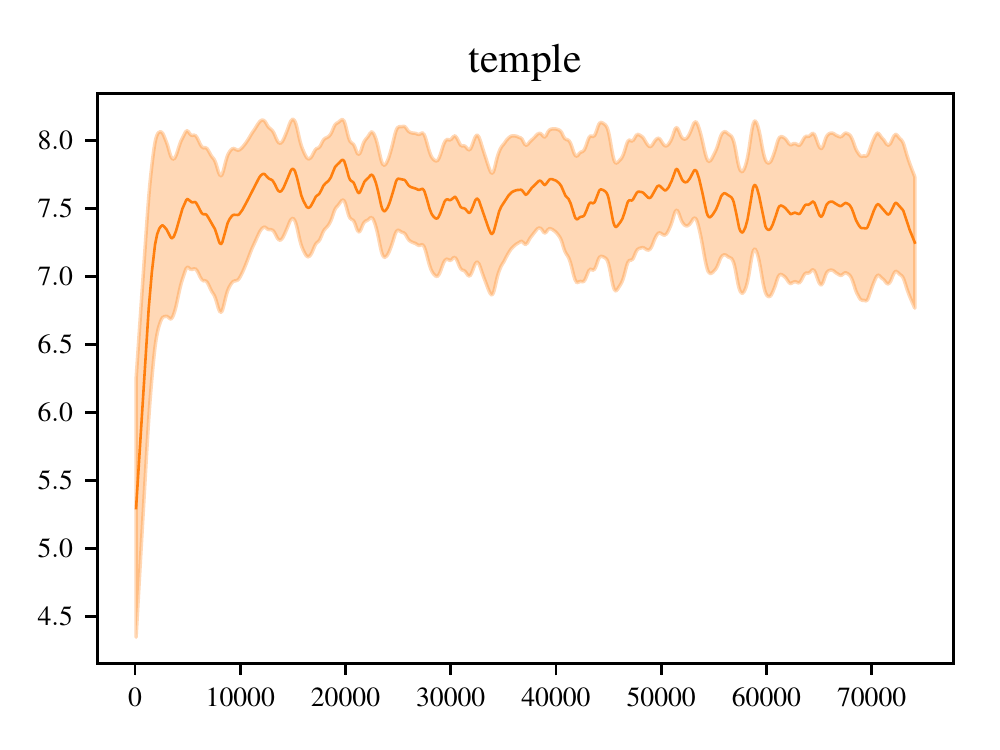}
\end{minipage}
\begin{minipage}{.19\textwidth}
  \centering
  \includegraphics[width=\linewidth]{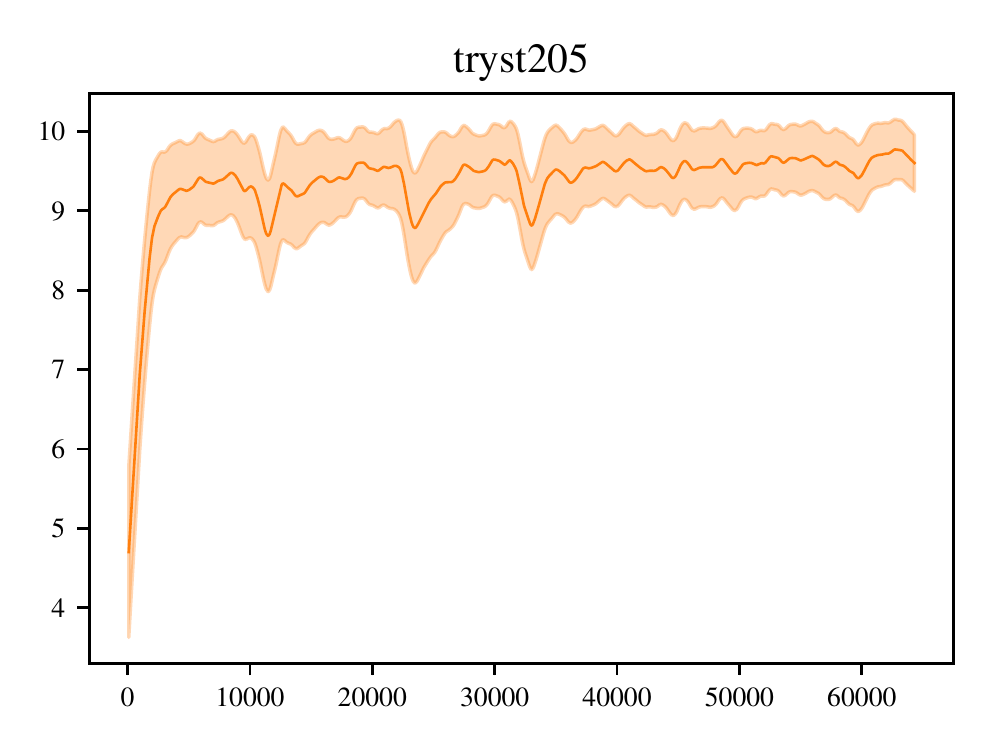}
\end{minipage}
\begin{minipage}{.19\textwidth}
  \centering
  \includegraphics[width=\linewidth]{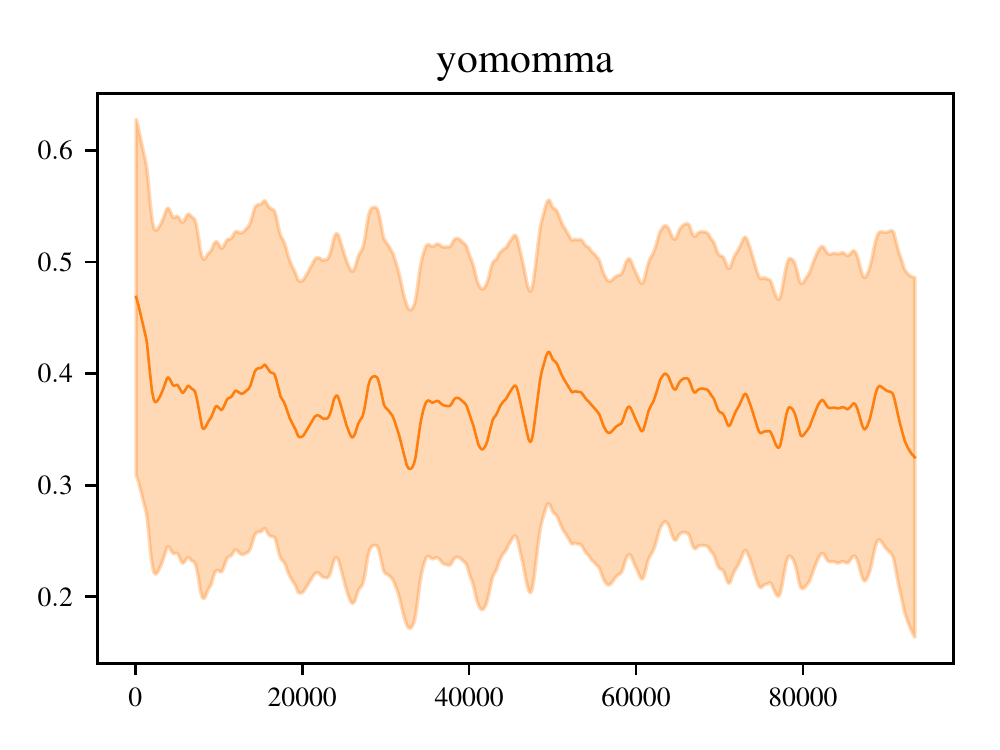}
\end{minipage}
\begin{minipage}{.19\textwidth}
  \centering
  \includegraphics[width=\linewidth]{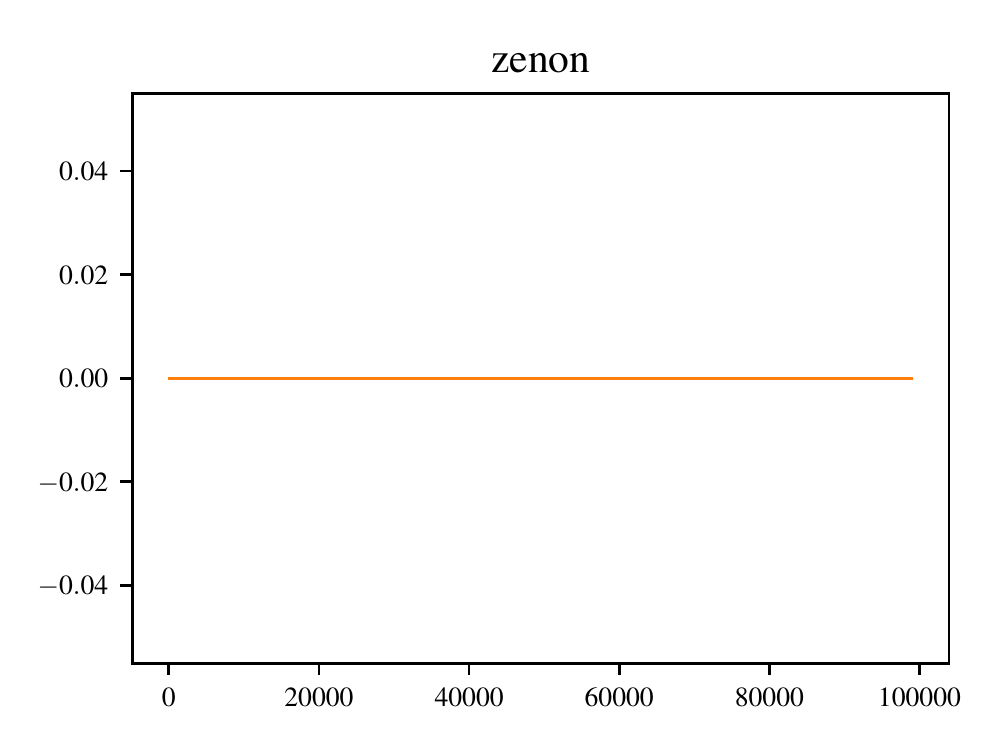}
\end{minipage}
\begin{minipage}{.19\textwidth}
  \centering
  \includegraphics[width=\linewidth]{figures/plots/drrn_zork1.pdf}
\end{minipage}
\begin{minipage}{.19\textwidth}
  \centering
  \includegraphics[width=\linewidth]{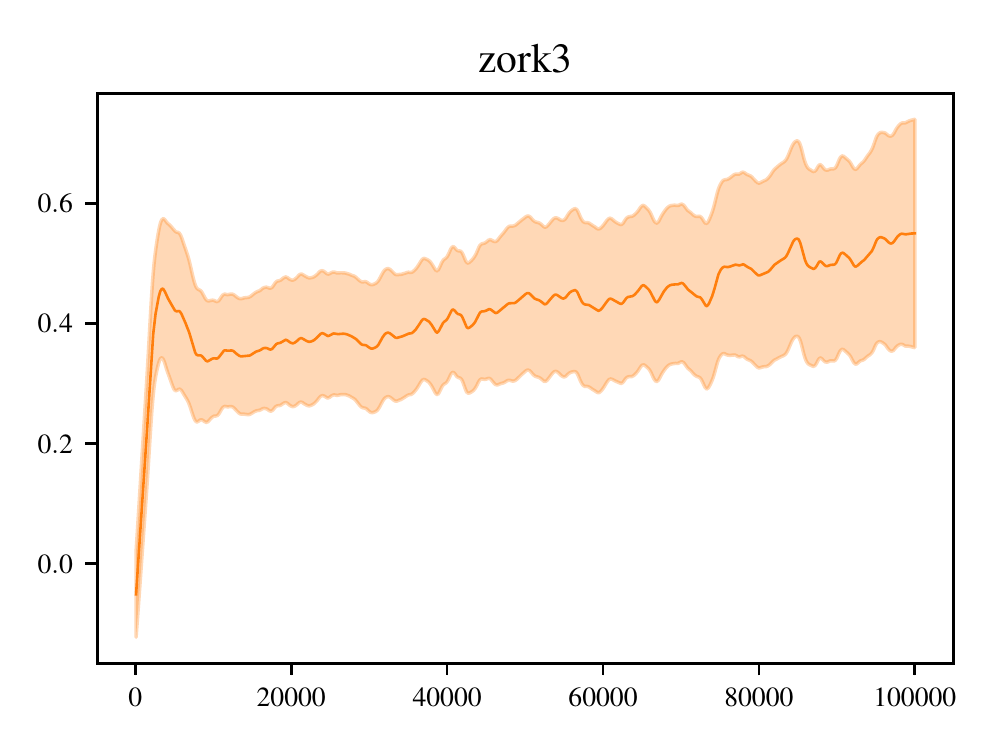}
\end{minipage}
\begin{minipage}{.19\textwidth}
  \centering
  \includegraphics[width=\linewidth]{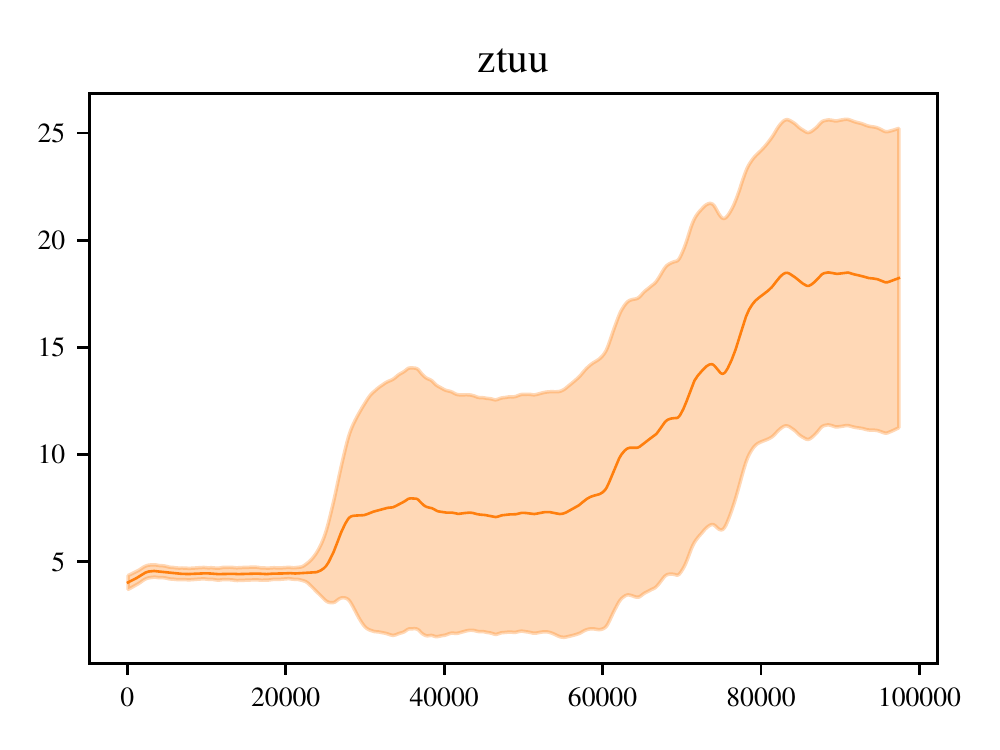}
\end{minipage}
\caption{Learning curves for DRRN.}
\end{figure*}

\newpage

\begin{figure*}[h]
\begin{minipage}{.19\textwidth}
  \centering
  \includegraphics[width=\linewidth]{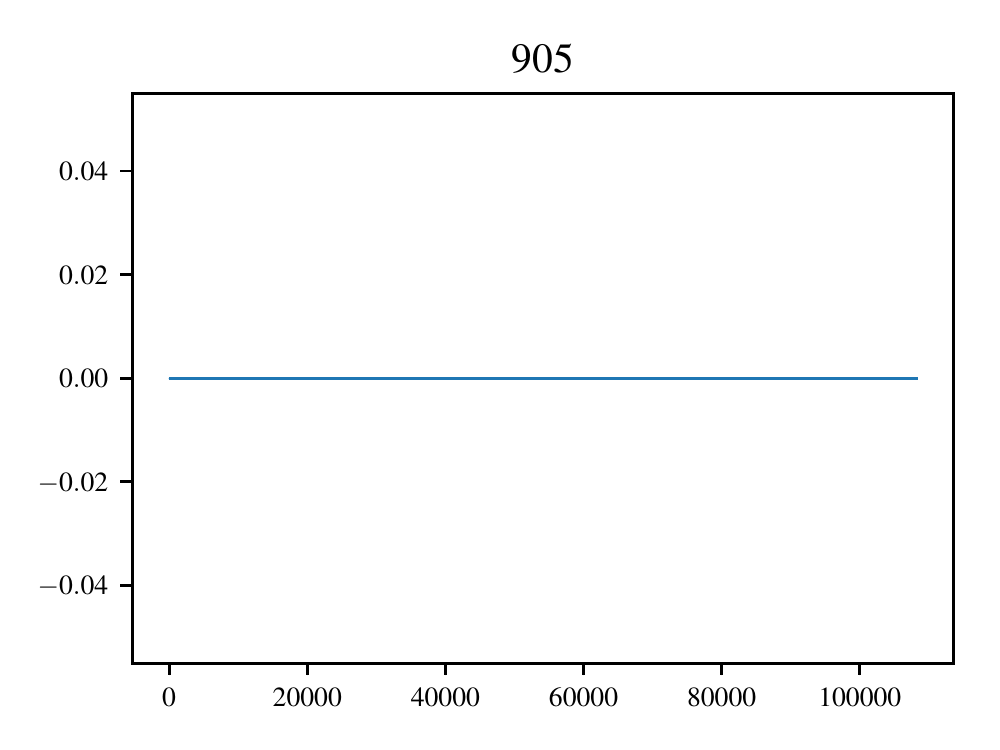}
\end{minipage}
\begin{minipage}{.19\textwidth}
  \centering
  \includegraphics[width=\linewidth]{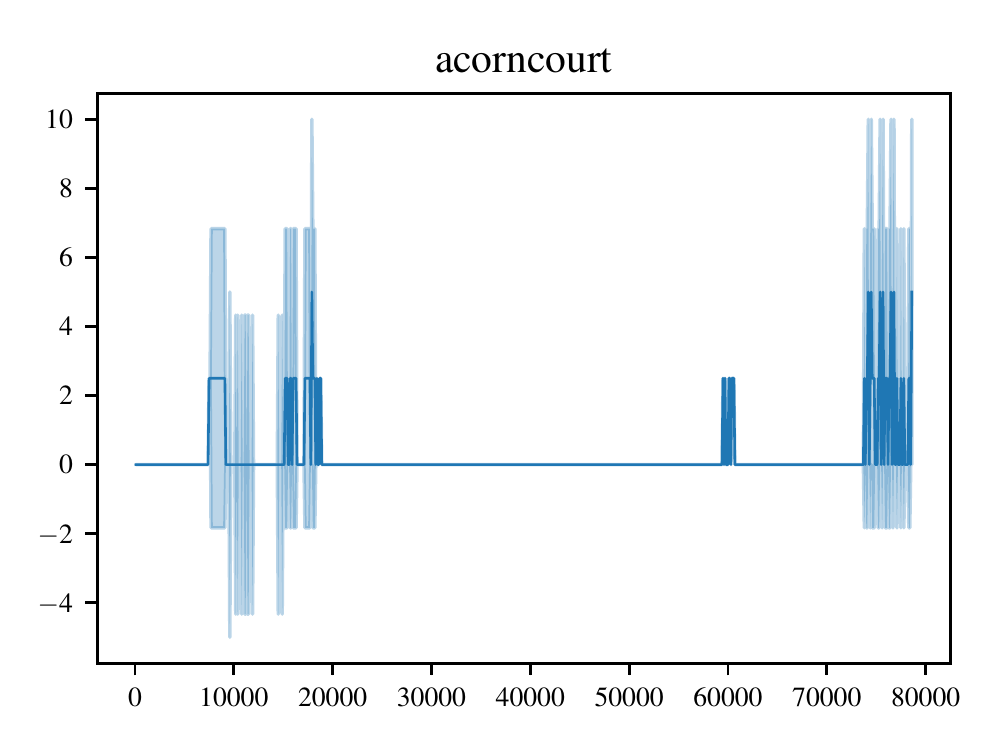}
\end{minipage}
\begin{minipage}{.19\textwidth}
  \centering
  \includegraphics[width=\linewidth]{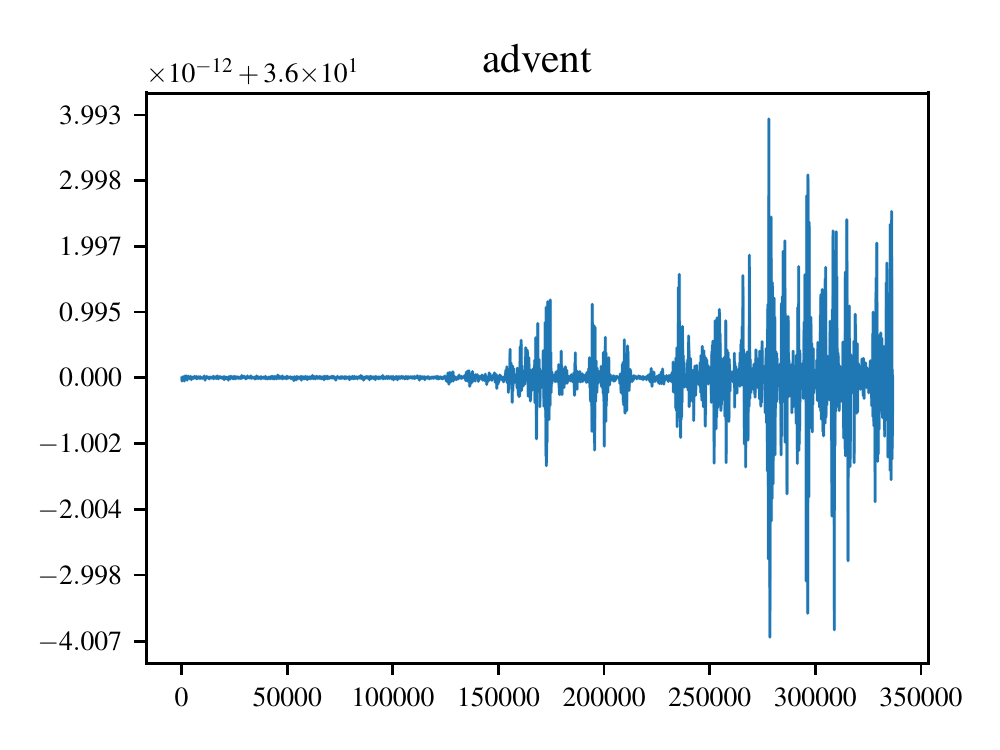}
\end{minipage}
\begin{minipage}{.19\textwidth}
  \centering
  \includegraphics[width=\linewidth]{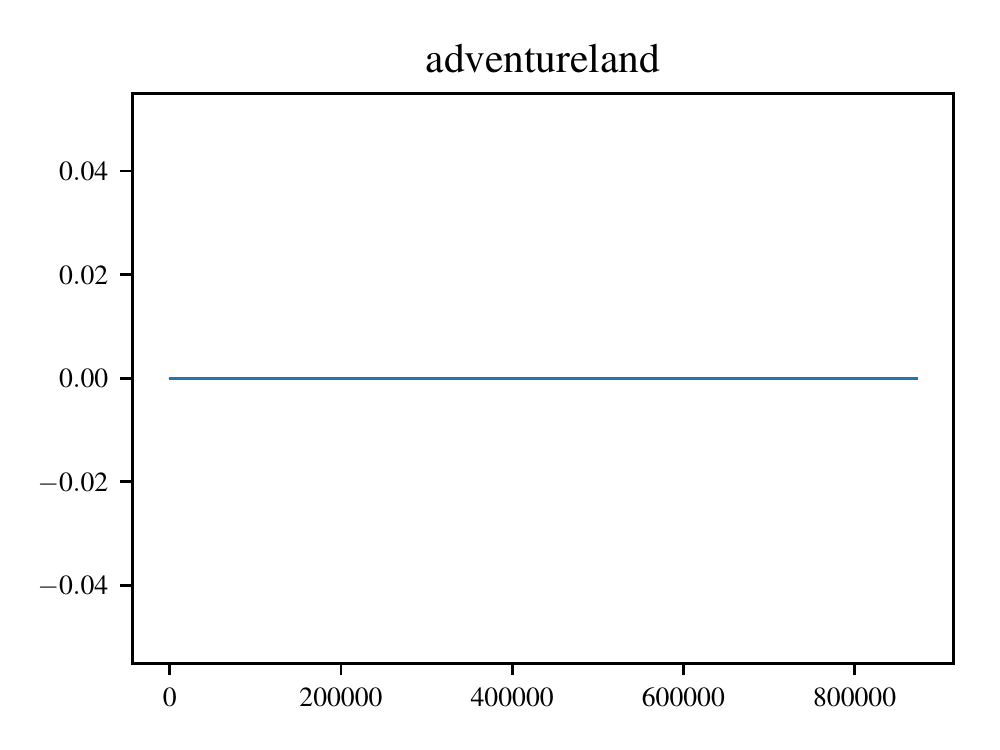}
\end{minipage}
\begin{minipage}{.19\textwidth}
  \centering
  \includegraphics[width=\linewidth]{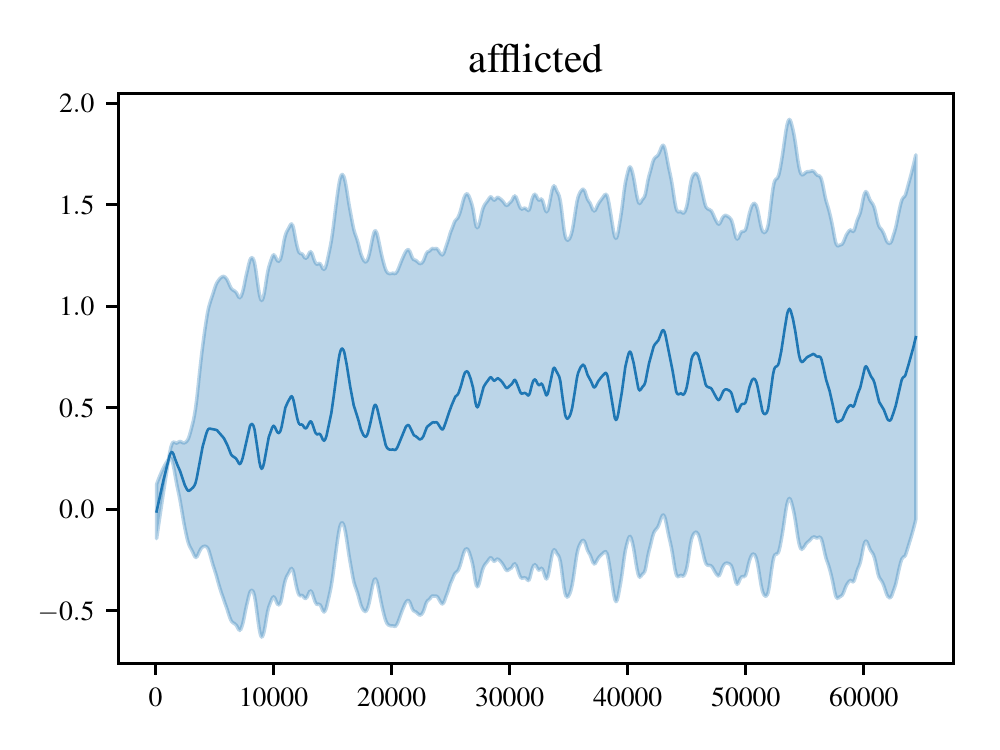}
\end{minipage}
\begin{minipage}{.19\textwidth}
  \centering
  \includegraphics[width=\linewidth]{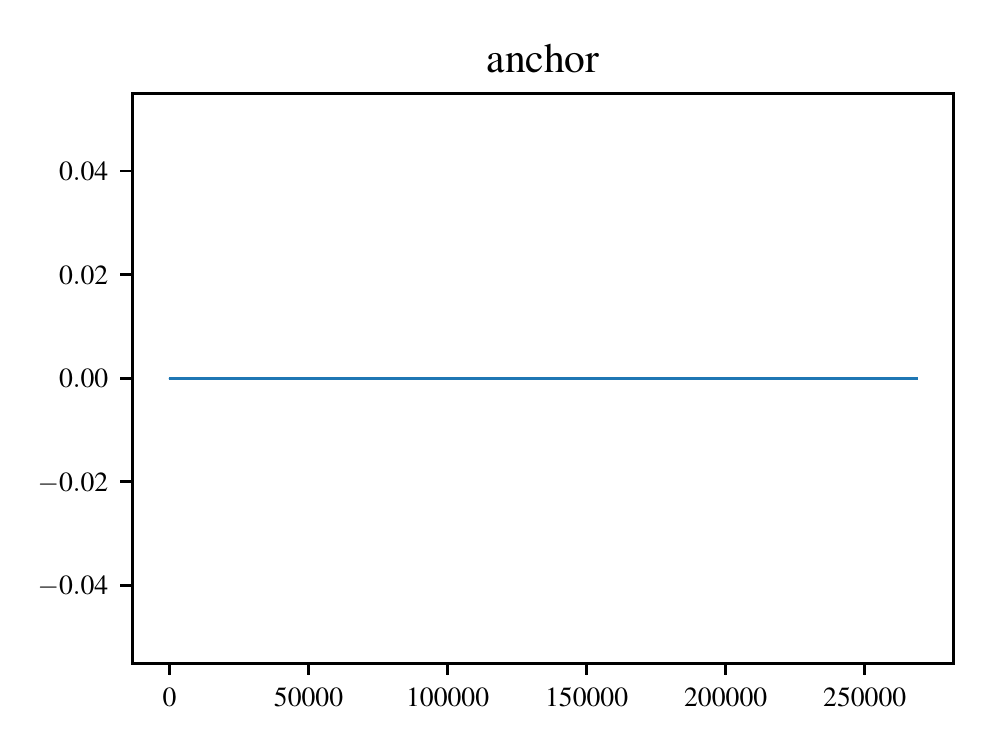}
\end{minipage}
\begin{minipage}{.19\textwidth}
  \centering
  \includegraphics[width=\linewidth]{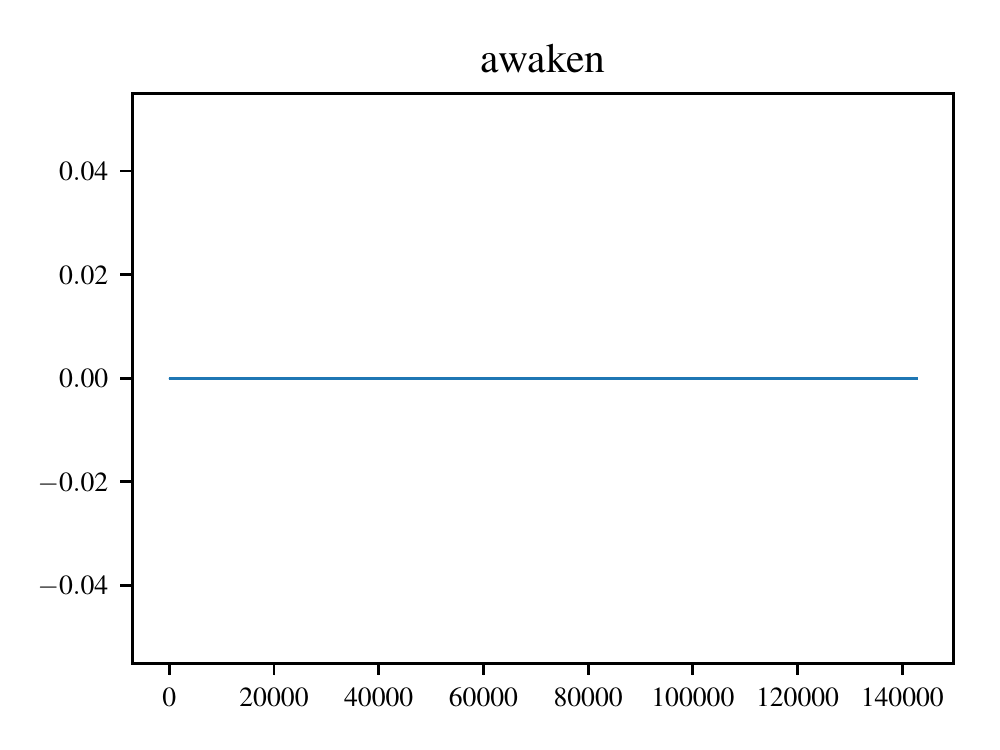}
\end{minipage}
\begin{minipage}{.19\textwidth}
  \centering
  \includegraphics[width=\linewidth]{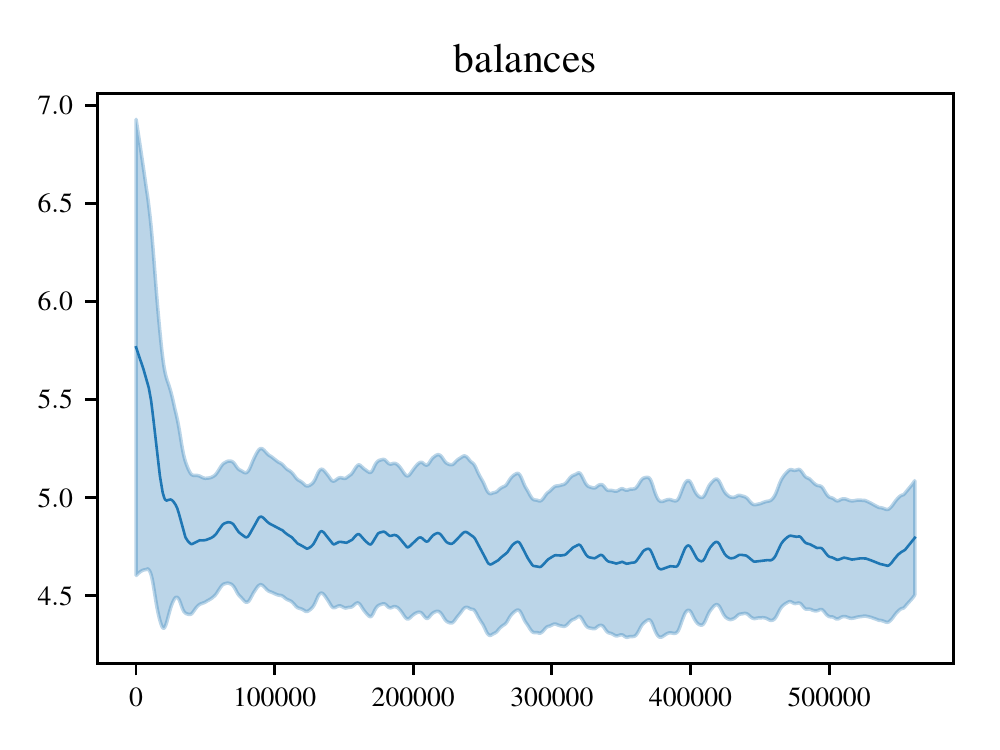}
\end{minipage}
\begin{minipage}{.19\textwidth}
  \centering
  \includegraphics[width=\linewidth]{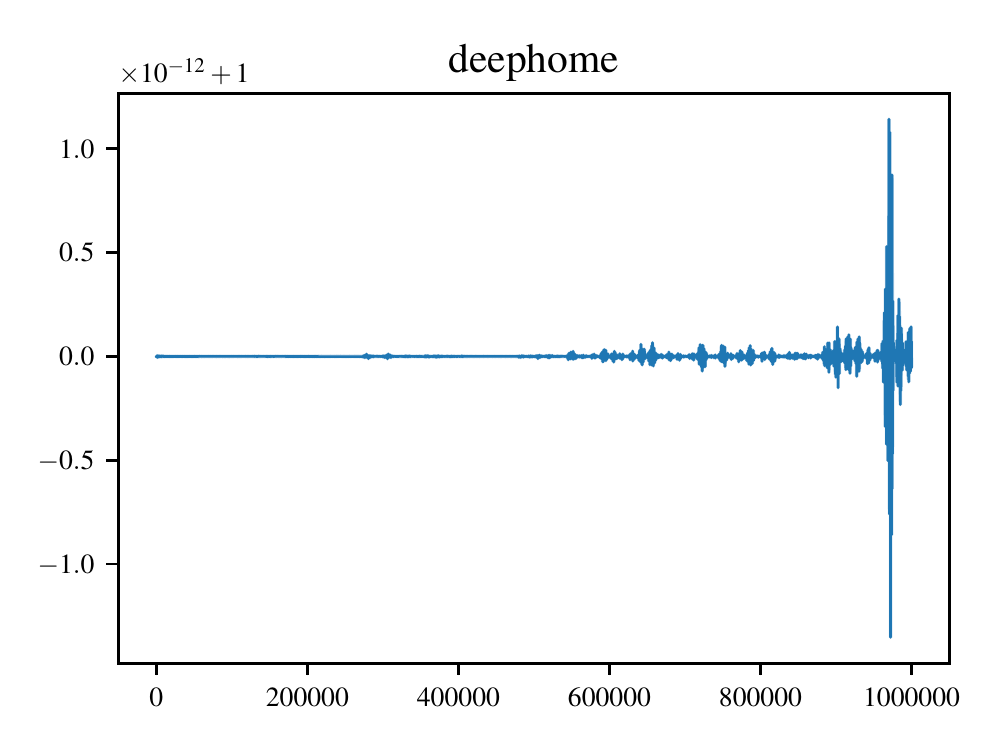}
\end{minipage}
\begin{minipage}{.19\textwidth}
  \centering
  \includegraphics[width=\linewidth]{figures/plots/tdqn_detective.pdf}
\end{minipage}
\begin{minipage}{.19\textwidth}
  \centering
  \includegraphics[width=\linewidth]{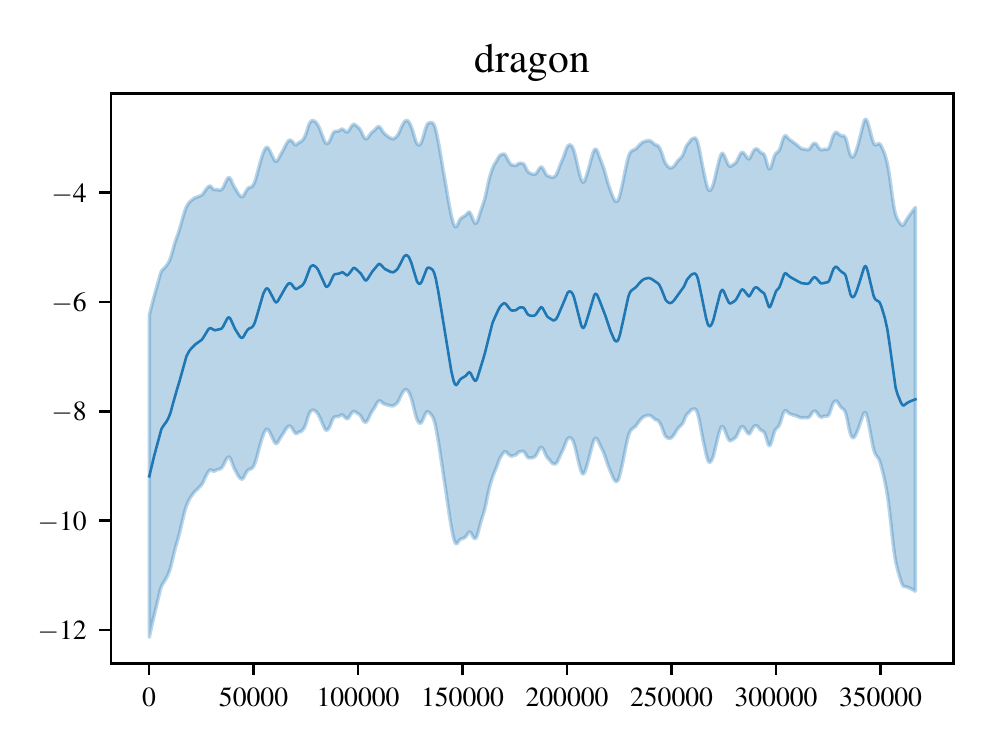}
\end{minipage}
\begin{minipage}{.19\textwidth}
  \centering
  \includegraphics[width=\linewidth]{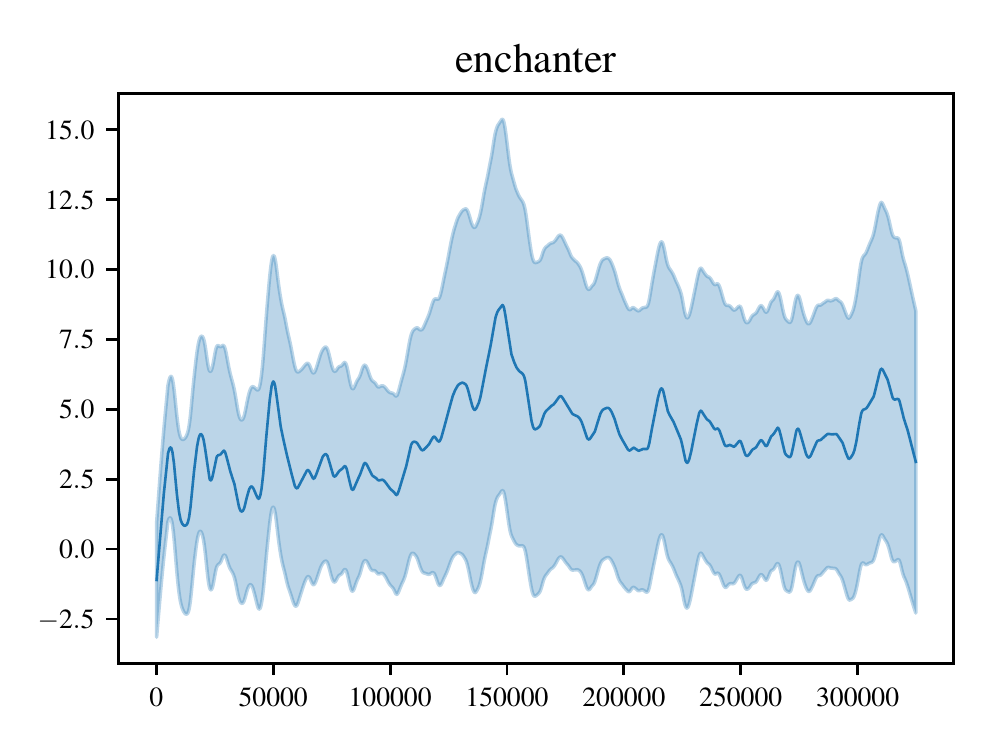}
\end{minipage}
\begin{minipage}{.19\textwidth}
  \centering
  \includegraphics[width=\linewidth]{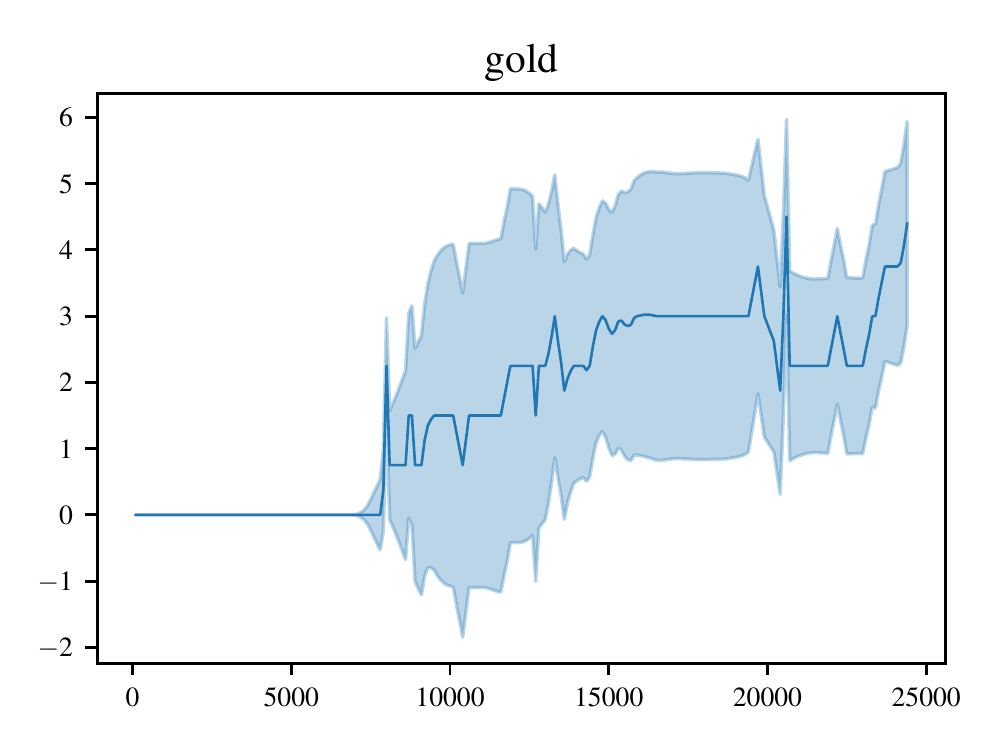}
\end{minipage}
\begin{minipage}{.19\textwidth}
  \centering
  \includegraphics[width=\linewidth]{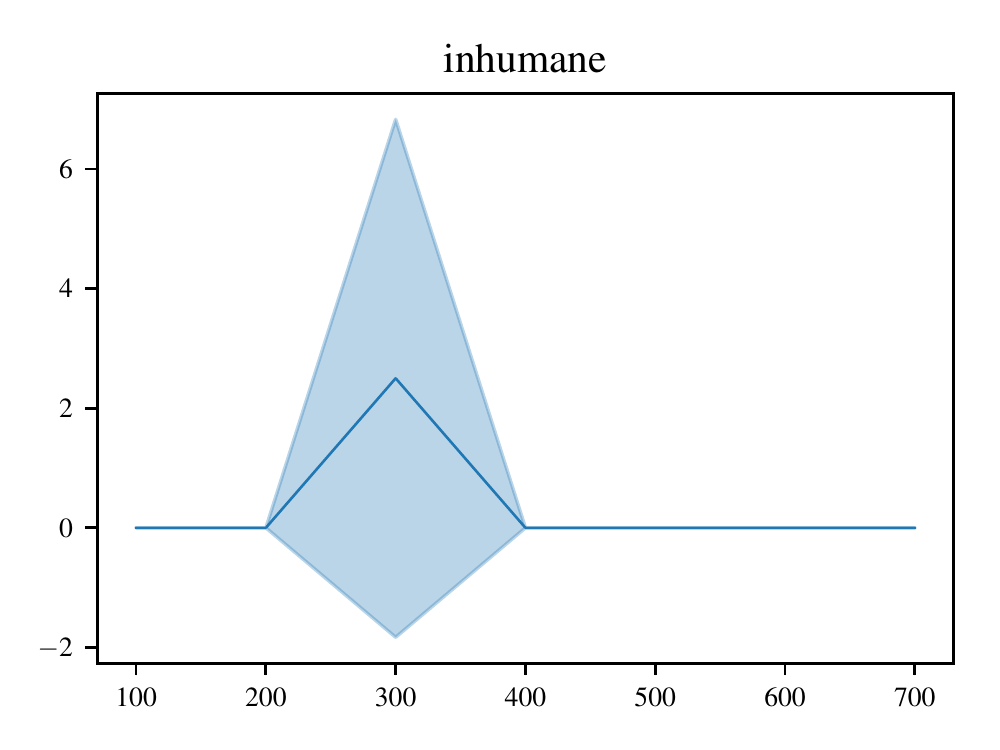}
\end{minipage}
\begin{minipage}{.19\textwidth}
  \centering
  \includegraphics[width=\linewidth]{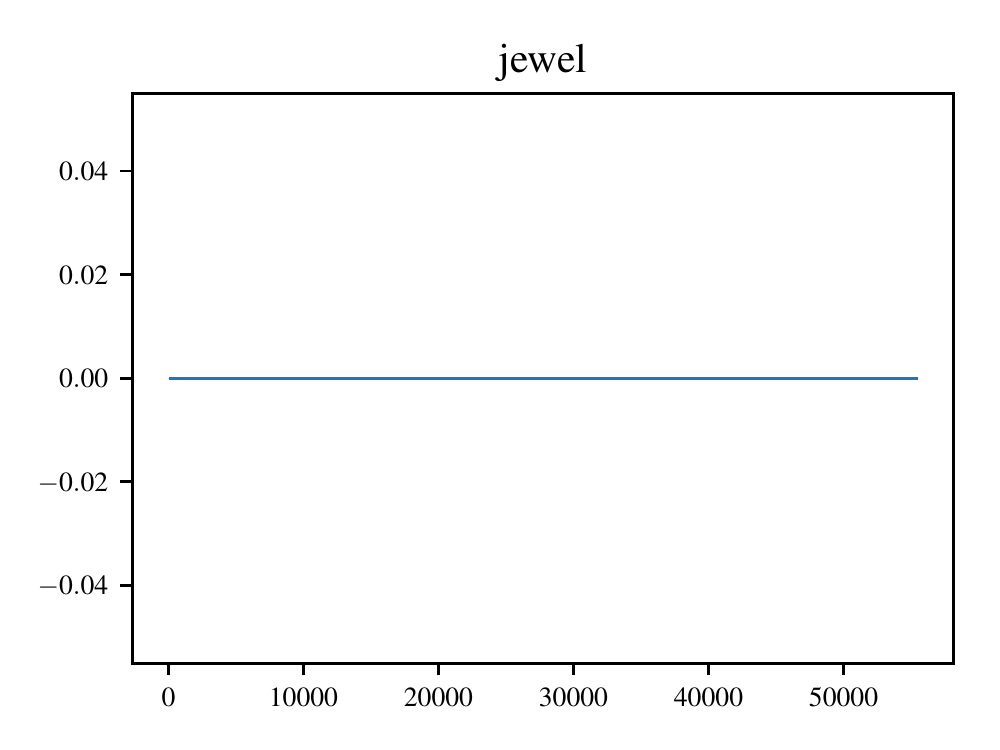}
\end{minipage}
\begin{minipage}{.19\textwidth}
  \centering
  \includegraphics[width=\linewidth]{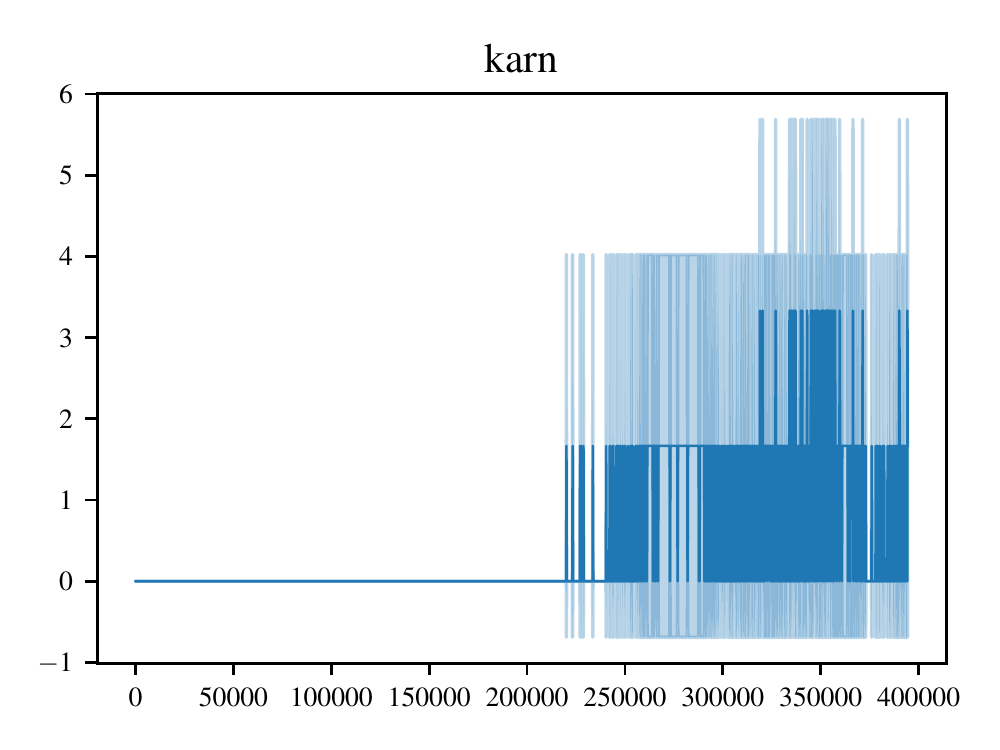}
\end{minipage}
\begin{minipage}{.19\textwidth}
  \centering
  \includegraphics[width=\linewidth]{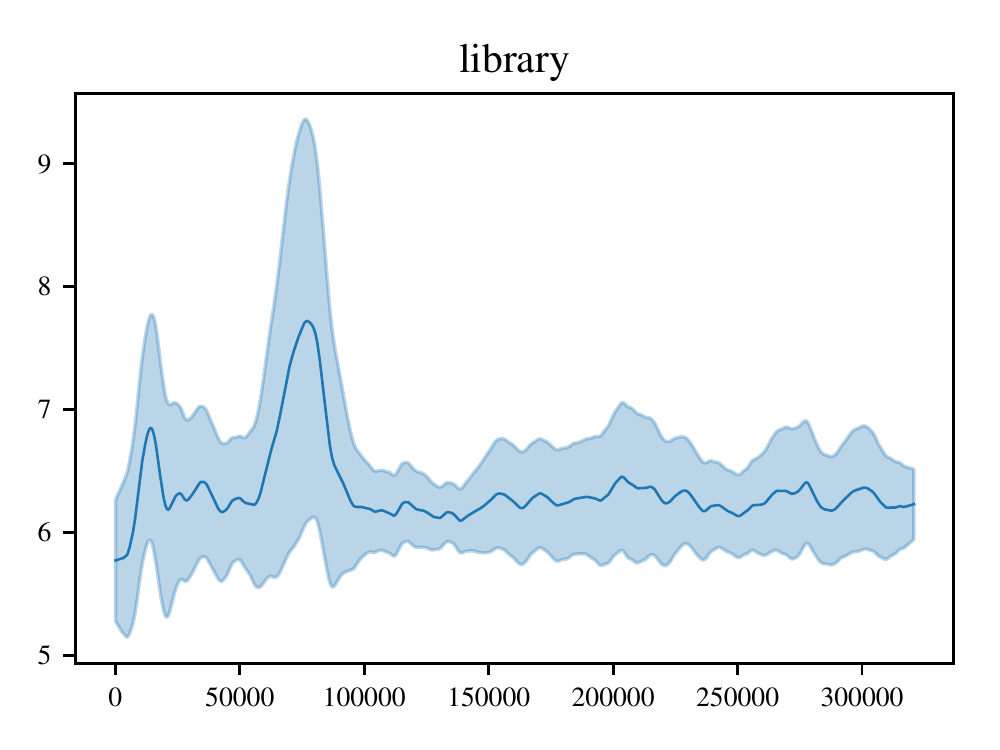}
\end{minipage}
\begin{minipage}{.19\textwidth}
  \centering
  \includegraphics[width=\linewidth]{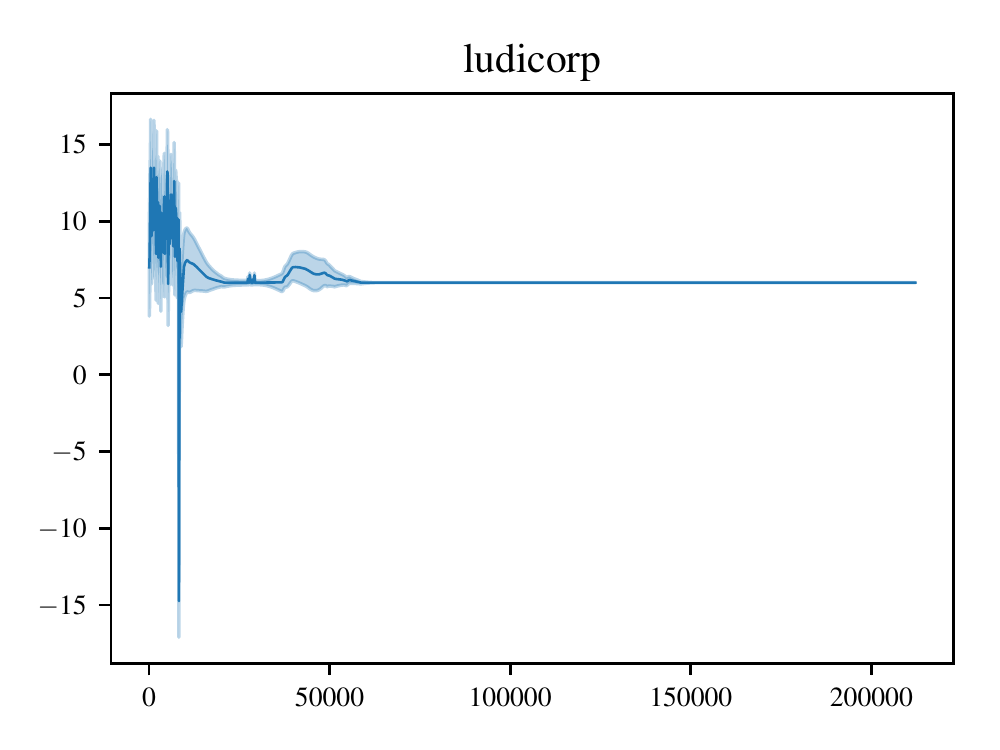}
\end{minipage}
\begin{minipage}{.19\textwidth}
  \centering
  \includegraphics[width=\linewidth]{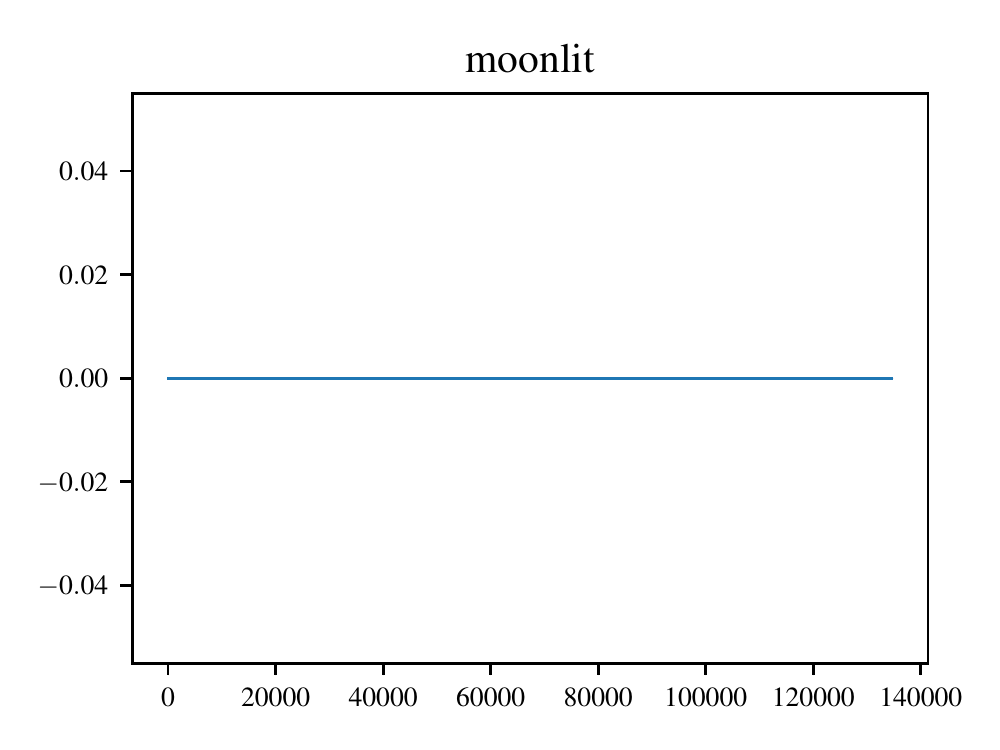}
\end{minipage}
\begin{minipage}{.19\textwidth}
  \centering
  \includegraphics[width=\linewidth]{figures/plots/tdqn_omniquest.pdf}
\end{minipage}
\begin{minipage}{.19\textwidth}
  \centering
  \includegraphics[width=\linewidth]{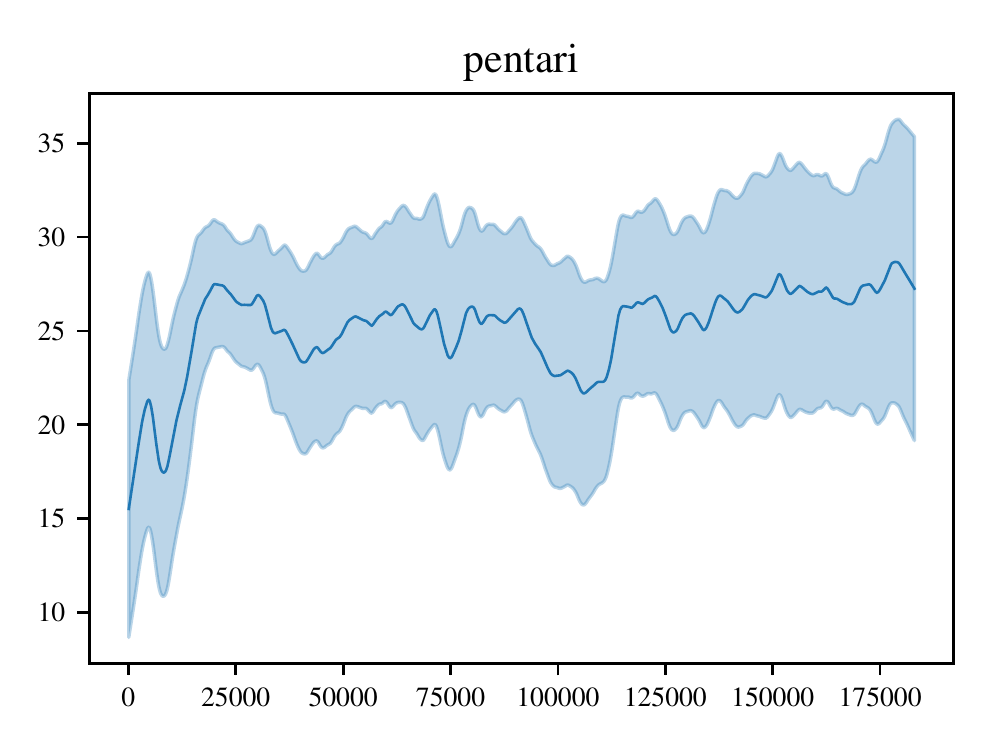}
\end{minipage}
\begin{minipage}{.19\textwidth}
  \centering
  \includegraphics[width=\linewidth]{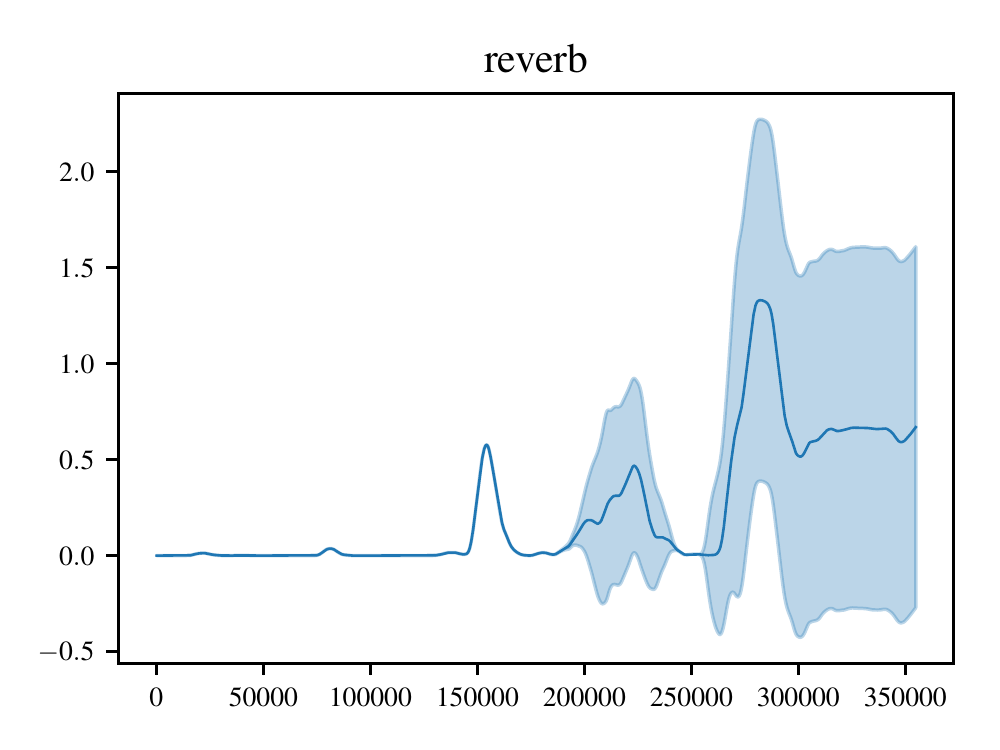}
\end{minipage}
\begin{minipage}{.19\textwidth}
  \centering
  \includegraphics[width=\linewidth]{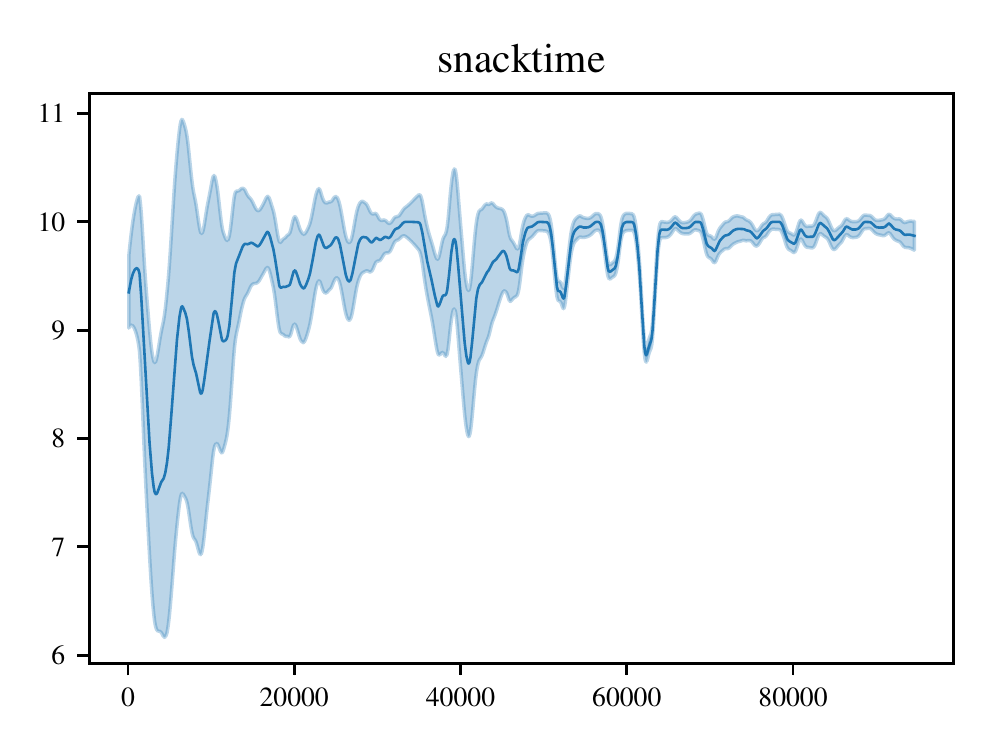}
\end{minipage}
\begin{minipage}{.19\textwidth}
  \centering
  \includegraphics[width=\linewidth]{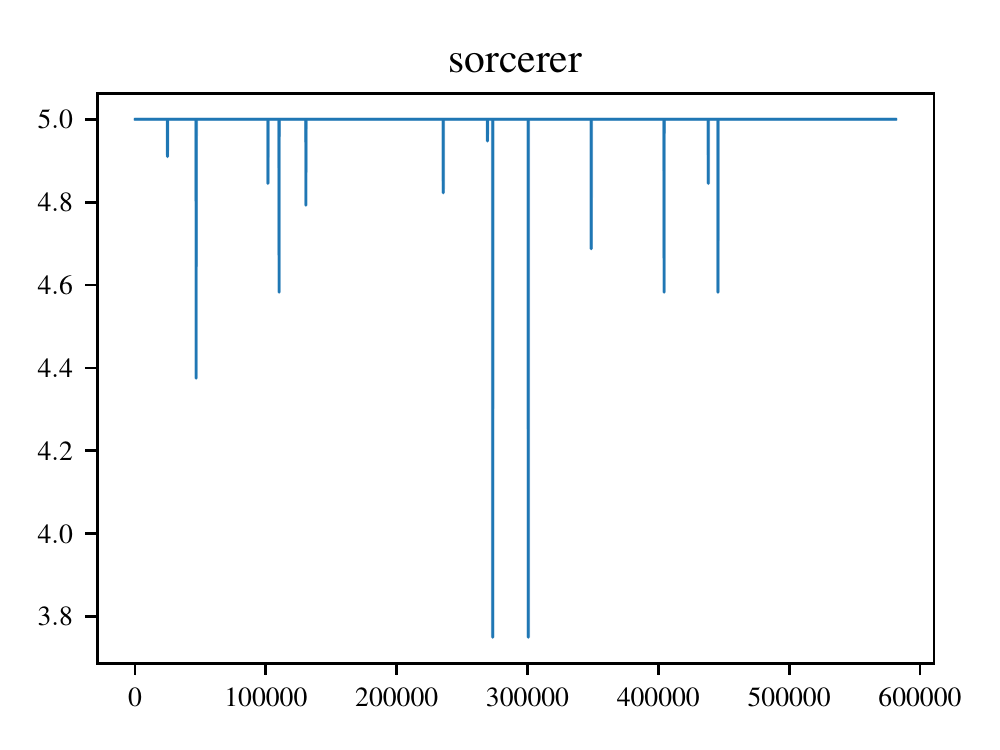}
\end{minipage}
\begin{minipage}{.19\textwidth}
  \centering
  \includegraphics[width=\linewidth]{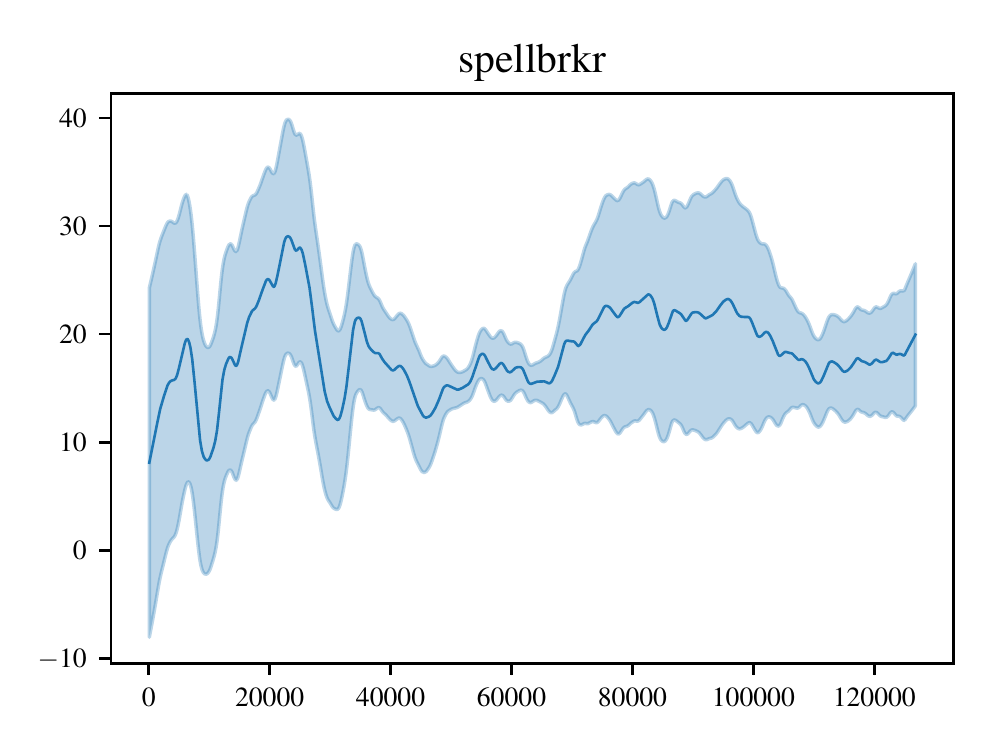}
\end{minipage}
\begin{minipage}{.19\textwidth}
  \centering
  \includegraphics[width=\linewidth]{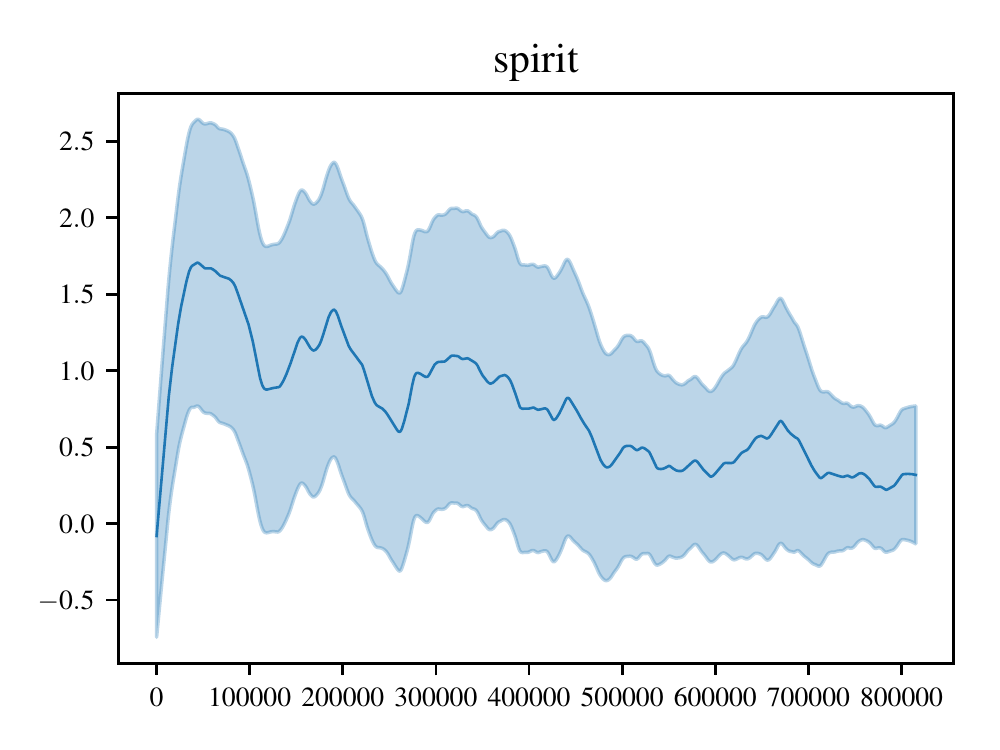}
\end{minipage}
\begin{minipage}{.19\textwidth}
  \centering
  \includegraphics[width=\linewidth]{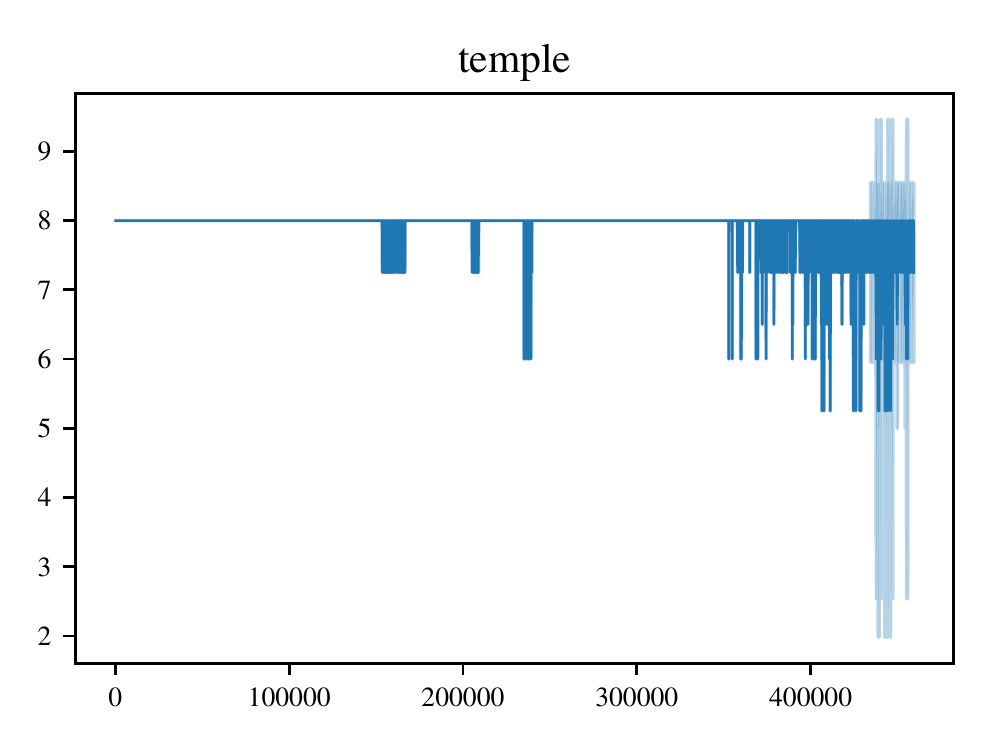}
\end{minipage}
\begin{minipage}{.19\textwidth}
  \centering
  \includegraphics[width=\linewidth]{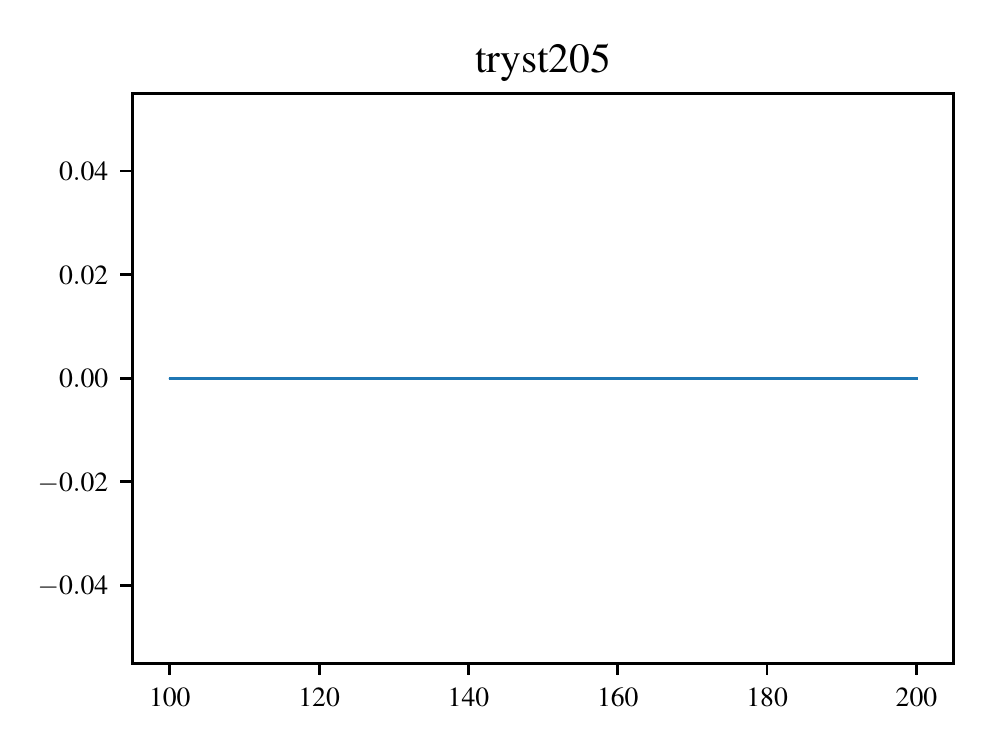}
\end{minipage}
\begin{minipage}{.19\textwidth}
  \centering
  \includegraphics[width=\linewidth]{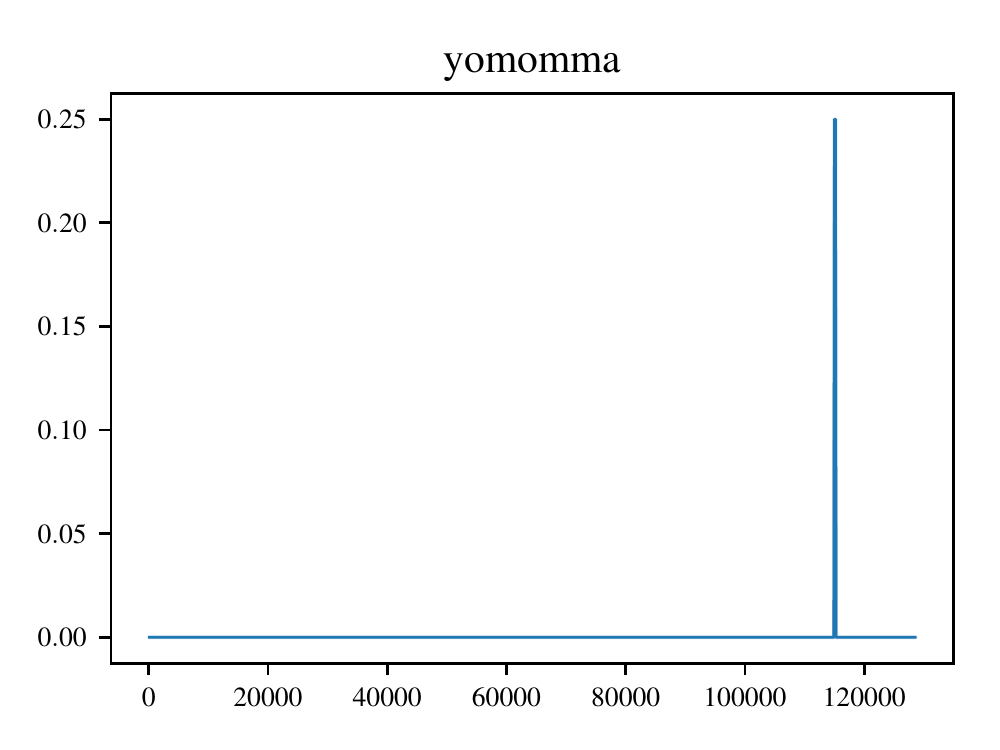}
\end{minipage}
\begin{minipage}{.19\textwidth}
  \centering
  \includegraphics[width=\linewidth]{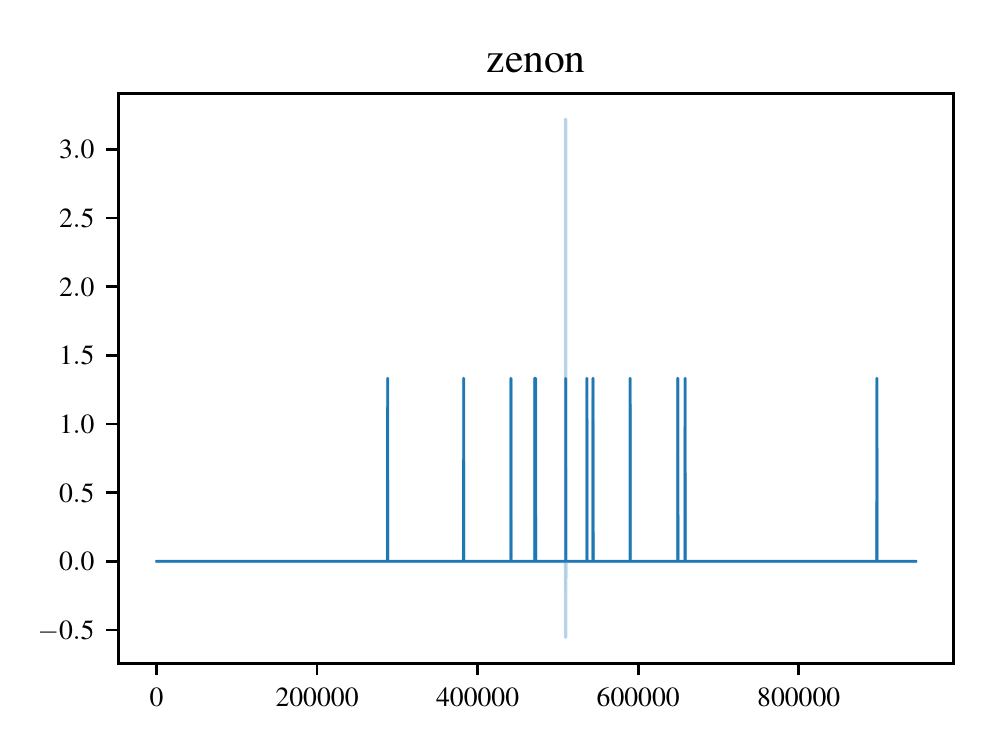}
\end{minipage}
\begin{minipage}{.19\textwidth}
  \centering
  \includegraphics[width=\linewidth]{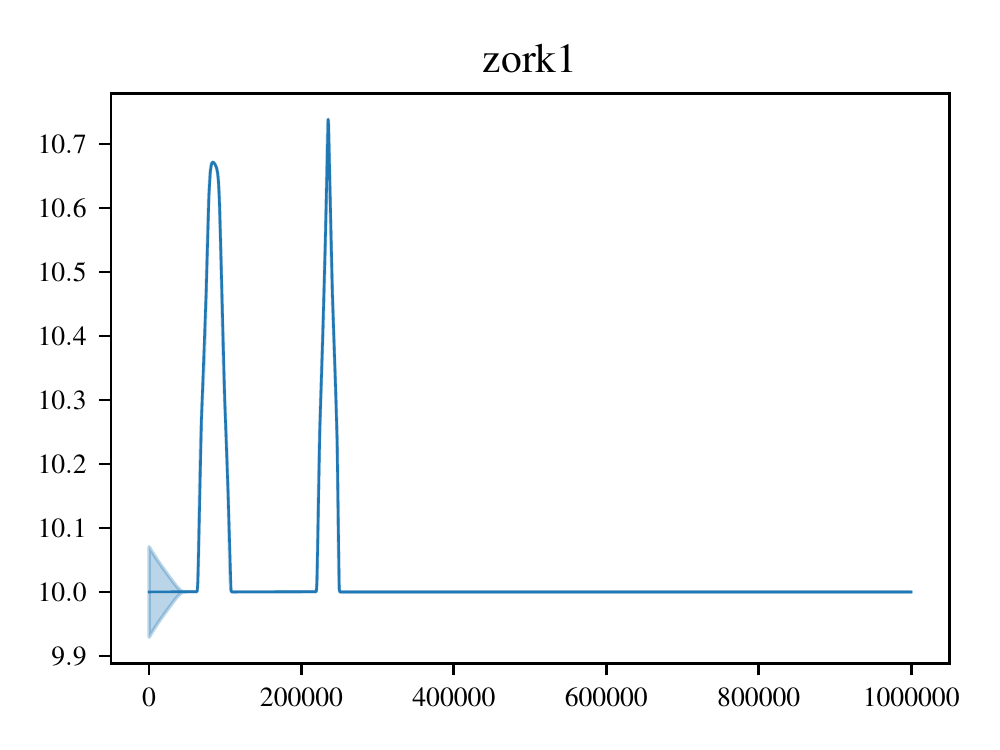}
\end{minipage}
\begin{minipage}{.19\textwidth}
  \centering
  \includegraphics[width=\linewidth]{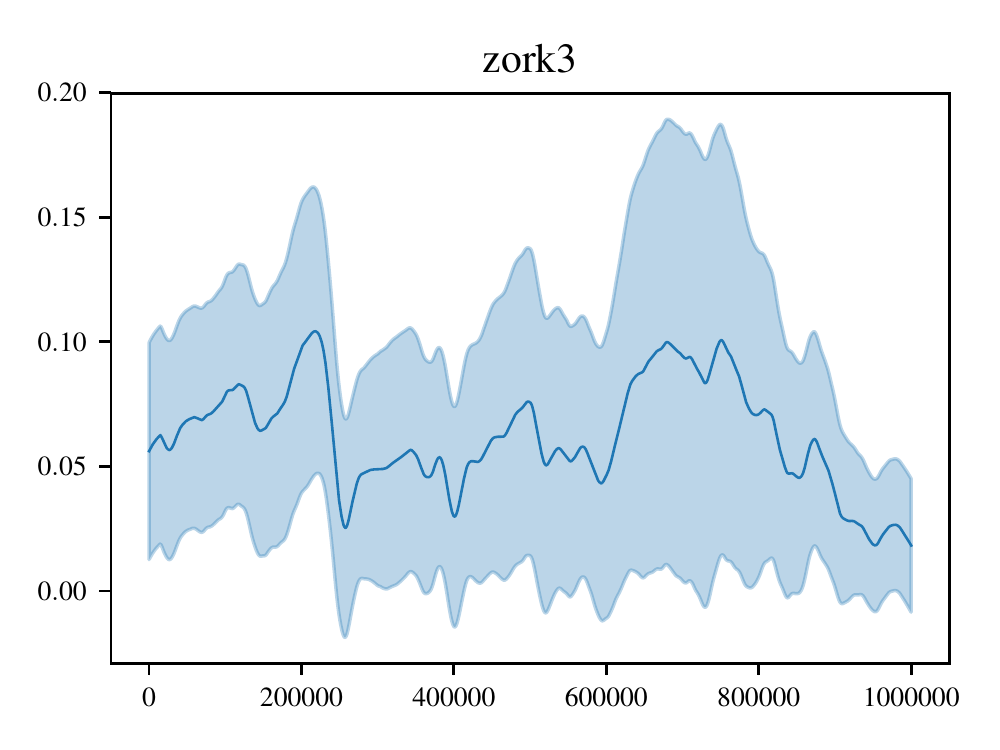}
\end{minipage}
\begin{minipage}{.19\textwidth}
  \centering
  \includegraphics[width=\linewidth]{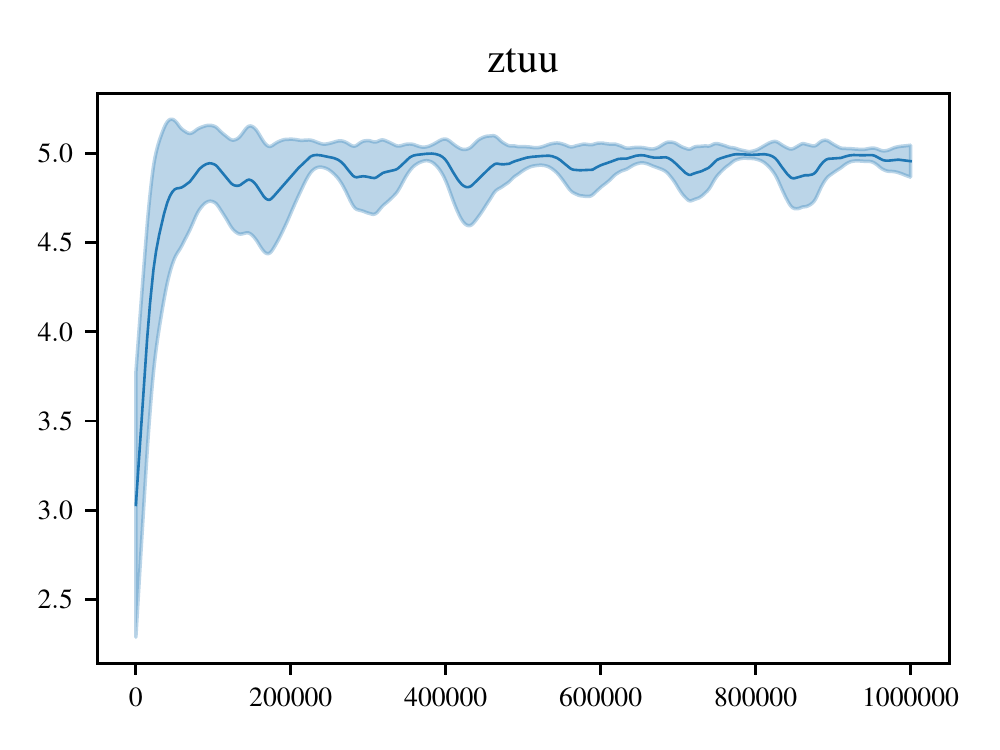}
\end{minipage}
\caption{Learning curves for TDQN.}
\end{figure*}

\newpage

\section{Select Transcripts}
\textbf{DRRN playing Zork1}
\begin{lstlisting}
Obs0: Copyright (c) 1981, 1982, 1983 Infocom, Inc. All rights reserved. ZORK is a registered trademark of Infocom, Inc. Revision 88 / Serial number 840726  West of House You are standing in an open field west of a white house, with a boarded front door. There is a small mailbox here. Inv: You are empty handed. Desc: West of House You are standing in an open field west of a white house, with a boarded front door. There is a small mailbox here.
Action0: open mailbox, Q-Value 15.78
Q-Values: 1)15.78 open mailbox 2)13.64 west 3)12.94 north 4)12.68 south 
Reward0: 0, Score 0, Done False
Obs1: Opening the small mailbox reveals a leaflet. Inv: You are empty handed. Desc: West of House You are standing in an open field west of a white house, with a boarded front door. There is a small mailbox here. The small mailbox contains:   A leaflet
Action1: take leaflet, Q-Value 17.54
Q-Values: 1)17.54 take leaflet 2)13.60 close mailbox 3)13.58 west 4)12.91 north 5)12.64 south 
Reward1: 0, Score 0, Done False
Obs2: Taken. Inv: You are carrying:   A leaflet Desc: West of House You are standing in an open field west of a white house, with a boarded front door. There is a small mailbox here.
Action2: south, Q-Value 19.56
Q-Values: 1)19.56 south 2)19.33 north 3)17.49 close mailbox 4)16.89 west 5)15.85 put down leaflet 6)15.78 put leaflet in mailbox 
Reward2: 0, Score 0, Done False
Obs3: South of House You are facing the south side of a white house. There is no door here, and all the windows are boarded. Inv: You are carrying:   A leaflet Desc: South of House You are facing the south side of a white house. There is no door here, and all the windows are boarded.
Action3: east, Q-Value 22.05
Q-Values: 1)22.05 east 2)17.51 south 3)17.27 west 4)14.76 put down leaflet 
Reward3: 0, Score 0, Done False
Obs4: Behind House You are behind the white house. A path leads into the forest to the east. In one corner of the house there is a small window which is slightly ajar. Inv: You are carrying:   A leaflet Desc: Behind House You are behind the white house. A path leads into the forest to the east. In one corner of the house there is a small window which is slightly ajar.
Action4: open window, Q-Value 25.55
Q-Values: 1)25.55 open window 2)19.50 east 3)19.37 north 4)19.18 south 5)17.09 put down leaflet 
Reward4: 0, Score 0, Done False
Obs5: With great effort, you open the window far enough to allow entry. Inv: You are carrying:   A leaflet Desc: Behind House You are behind the white house. A path leads into the forest to the east. In one corner of the house there is a small window which is open.
Action5: west, Q-Value 28.65
Q-Values: 1)28.65 west 2)21.82 close window 3)20.03 north 4)19.40 east 5)19.38 south 6)18.18 put down small 
Reward5: 10, Score 10, Done False
Obs6: Kitchen You are in the kitchen of the white house. A table seems to have been used recently for the preparation of food. A passage leads to the west and a dark staircase can be seen leading upward. A dark chimney leads down and to the east is a small window which is open. On the table is an elongated brown sack, smelling of hot peppers. A bottle is sitting on the table. The glass bottle contains:   A quantity of water Inv: You are carrying:   A leaflet Desc: Kitchen You are in the kitchen of the white house. A table seems to have been used recently for the preparation of food. A passage leads to the west and a dark staircase can be seen leading upward. A dark chimney leads down and to the east is a small window which is open. On the table is an elongated brown sack, smelling of hot peppers. A bottle is sitting on the table. The glass bottle contains:   A quantity of water
Action6: west, Q-Value 21.32
Q-Values: 1)21.32 west 2)19.79 close window 3)19.78 open glass 4)19.44 open sack 5)18.84 take all 6)18.06 take glass 7)17.91 put small on kitchen 8)16.39 take sack 9)15.65 east 10)14.53 put down small 11)5.03 up 12)2.14 jump 
Reward6: 0, Score 10, Done False
Obs7: Living Room You are in the living room. There is a doorway to the east, a wooden door with strange gothic lettering to the west, which appears to be nailed shut, a trophy case, and a large oriental rug in the center of the room. Above the trophy case hangs an elvish sword of great antiquity. A battery powered brass lantern is on the trophy case. Inv: You are carrying:   A leaflet Desc: Living Room You are in the living room. There is a doorway to the east, a wooden door with strange gothic lettering to the west, which appears to be nailed shut, a trophy case, and a large oriental rug in the center of the room. Above the trophy case hangs an elvish sword of great antiquity. A battery powered brass lantern is on the trophy case.
Action7: push rug, Q-Value 23.47
Q-Values: 1)23.47 push rug 2)23.46 take all 3)21.44 open trophy 4)21.21 east 5)21.11 turn on lantern 6)19.86 take antiquity 7)16.36 put down leaflet 8)14.58 throw leaflet at lantern 9)14.35 take lantern 
Reward7: 0, Score 10, Done False
Obs8: With a great effort, the rug is moved to one side of the room, revealing the dusty cover of a closed trap door. Inv: You are carrying:   A leaflet Desc: Living Room You are in the living room. There is a doorway to the east, a wooden door with strange gothic lettering to the west, which appears to be nailed shut, a trophy case, and a closed trap door at your feet. Above the trophy case hangs an elvish sword of great antiquity. A battery powered brass lantern is on the trophy case.
Action8: open cover, Q-Value 26.00
Q-Values: 1)26.00 open cover 2)25.68 take all 3)23.49 open case 4)23.26 turn on brass 5)22.09 east 6)21.69 take elvish 7)18.22 put down leaflet 8)16.74 throw leaflet at brass 9)14.47 take brass 
Reward8: 0, Score 10, Done False
Obs9: The door reluctantly opens to reveal a rickety staircase descending into darkness. Inv: You are carrying:   A leaflet Desc: Living Room You are in the living room. There is a doorway to the east, a wooden door with strange gothic lettering to the west, which appears to be nailed shut, a trophy case, and a rug lying beside an open trap door. Above the trophy case hangs an elvish sword of great antiquity. A battery powered brass lantern is on the trophy case.
Action9: take all, Q-Value 28.88
Q-Values: 1)28.88 take all 2)24.94 down 3)24.93 close trap 4)24.92 open case 5)24.58 turn on brass 6)23.12 take elvish 7)22.72 east 8)19.16 put down leaflet 9)15.22 throw leaflet at brass 10)14.72 take brass 
Reward9: 0, Score 10, Done False
Obs10: trophy case: The trophy case is securely fastened to the wall. sword: Taken. brass lantern: Taken. carpet: The rug is extremely heavy and cannot be carried. Inv: You are carrying:   A brass lantern   A sword   A leaflet Desc: Living Room You are in the living room. There is a doorway to the east, a wooden door with strange gothic lettering to the west, which appears to be nailed shut, a trophy case, and a rug lying beside an open trap door.
Action10: down, Q-Value 31.94
Q-Values: 1)31.94 down 2)25.86 open case 3)22.53 turn on brass 4)21.33 close trap 5)19.75 east 6)14.13 put down leaflet 7)11.69 put down sword 8)10.02 put down brass 9)8.54 throw brass at case 
Reward10: 25, Score 35, Done False
Obs11: You have moved into a dark place. The trap door crashes shut, and you hear someone barring it.  It is pitch black. You are likely to be eaten by a grue. Your sword is glowing with a faint blue glow. Inv: You are carrying:   A brass lantern   A sword   A leaflet Desc: It is pitch black. You are likely to be eaten by a grue.
Action11: turn on lantern, Q-Value 7.35
Q-Values: 1)7.35 turn on lantern 2)6.63 close door 3)6.17 put down sword 4)5.70 put down leaflet 5)-0.51 put down lantern 6)-4.24 exit door 7)-4.31 south 8)-4.33 down 9)-4.70 southeast 10)-5.16 leave 11)-5.80 northwest 12)-6.02 throw lantern at grue 13)-6.69 north 14)-6.92 exit dark 15)-7.23 northeast 16)-8.68 enter 17)-9.74 southwest 18)-10.72 exit lantern 19)-14.25 exit leaflet 
Reward11: 0, Score 35, Done False
Obs12: The brass lantern is now on.  Cellar You are in a dark and damp cellar with a narrow passageway leading north, and a crawlway to the south. On the west is the bottom of a steep metal ramp which is unclimbable. Inv: You are carrying:   A brass lantern (providing light)   A sword   A leaflet Desc: Cellar You are in a dark and damp cellar with a narrow passageway leading north, and a crawlway to the south. On the west is the bottom of a steep metal ramp which is unclimbable.
Action12: south, Q-Value 8.31
Q-Values: 1)8.31 south 2)7.14 put down sword 3)6.72 put down leaflet 4)6.62 north 5)6.61 put out lantern 6)6.45 put down lantern 7)-8.26 throw lantern at passageway 
Reward12: 0, Score 35, Done False
Obs13: East of Chasm You are on the east edge of a chasm, the bottom of which cannot be seen. A narrow passage goes north, and the path you are on continues to the east. Your sword is no longer glowing. Inv: You are carrying:   A brass lantern (providing light)   A sword   A leaflet Desc: East of Chasm You are on the east edge of a chasm, the bottom of which cannot be seen. A narrow passage goes north, and the path you are on continues to the east.
Action13: east, Q-Value 9.46
Q-Values: 1)9.46 east 2)8.36 put down sword 3)7.51 north 4)7.43 put down leaflet 5)7.20 put down lantern 6)6.85 put out lantern 7)0.93 throw leaflet at lantern 8)-10.08 jump 
Reward13: 0, Score 35, Done False
Obs14: Gallery This is an art gallery. Most of the paintings have been stolen by vandals with exceptional taste. The vandals left through either the north or west exits. Fortunately, there is still one chance for you to be a vandal, for on the far wall is a painting of unparalleled beauty. Inv: You are carrying:   A brass lantern (providing light)   A sword   A leaflet Desc: Gallery This is an art gallery. Most of the paintings have been stolen by vandals with exceptional taste. The vandals left through either the north or west exits. Fortunately, there is still one chance for you to be a vandal, for on the far wall is a painting of unparalleled beauty.
Action14: take paintings, Q-Value 10.64
Q-Values: 1)10.64 take paintings 2)10.03 put down sword 3)9.57 put down lantern 4)9.41 put out lantern 5)9.23 put down leaflet 6)8.43 north 7)7.56 west 8)7.16 throw lantern at paintings 
Reward14: 4, Score 39, Done False
Obs15: Taken. Inv: You are carrying:   A painting   A brass lantern (providing light)   A sword   A leaflet Desc: Gallery This is an art gallery. Most of the paintings have been stolen by vandals with exceptional taste. The vandals left through either the north or west exits.
Action15: put down paintings, Q-Value 7.71
Q-Values: 1)7.71 put down paintings 2)7.42 north 3)6.99 west 4)6.89 put down sword 5)6.71 put out brass 6)6.68 put down leaflet 7)6.24 put down brass 8)3.45 throw paintings at brass 
Reward15: 0, Score 39, Done False
Obs16: Dropped. Inv: You are carrying:   A brass lantern (providing light)   A sword   A leaflet Desc: Gallery This is an art gallery. Most of the paintings have been stolen by vandals with exceptional taste. The vandals left through either the north or west exits. A painting by a neglected genius is here.
Action16: north, Q-Value 7.95
Q-Values: 1)7.95 north 2)7.87 west 3)7.26 put down sword 4)6.74 take paintings 5)6.50 put down leaflet 6)6.48 put out lantern 7)6.38 put down lantern 8)1.68 throw leaflet at lantern 
Reward16: 0, Score 39, Done False
Obs17: Studio This appears to have been an artist's studio. The walls and floors are splattered with paints of 69 different colors. Strangely enough, nothing of value is hanging here. At the south end of the room is an open door (also covered with paint). A dark and narrow chimney leads up from a fireplace; although you might be able to get up it, it seems unlikely you could get back down. Loosely attached to a wall is a small piece of paper. Inv: You are carrying:   A brass lantern (providing light)   A sword   A leaflet Desc: Studio This appears to have been an artist's studio. The walls and floors are splattered with paints of 69 different colors. Strangely enough, nothing of value is hanging here. At the south end of the room is an open door (also covered with paint). A dark and narrow chimney leads up from a fireplace; although you might be able to get up it, it seems unlikely you could get back down. Loosely attached to a wall is a small piece of paper.
Action17: put down sword, Q-Value 8.38
Q-Values: 1)8.38 put down sword 2)7.84 south 3)7.65 take piece 4)7.56 put down leaflet 5)7.35 put down lantern 6)7.20 put out lantern 7)0.45 throw lantern at walls 
Reward17: 0, Score 39, Done False
Obs18: Dropped. Inv: You are carrying:   A brass lantern (providing light)   A leaflet Desc: Studio This appears to have been an artist's studio. The walls and floors are splattered with paints of 69 different colors. Strangely enough, nothing of value is hanging here. At the south end of the room is an open door (also covered with paint). A dark and narrow chimney leads up from a fireplace; although you might be able to get up it, it seems unlikely you could get back down. Loosely attached to a wall is a small piece of paper. There is a sword here.
Action18: up, Q-Value 9.93
Q-Values: 1)9.93 up 2)8.22 south 3)8.19 take sword 4)8.13 put down leaflet 5)7.95 take piece 6)7.75 put down lantern 7)7.29 take all 8)7.26 put out lantern 9)2.27 throw lantern at walls 
Reward18: 0, Score 39, Done False
\end{lstlisting}

\textbf{TDQN Playing Gold}
\begin{lstlisting}
Action: remove a Q_t: 15.37 Q_o1: 22.92 Q_o2: 10.71
Obs: (the wristwatch) I take off the wristwatch.
Reward 1: 0
Action: south Q_t: 16.10 Q_o1: 23.98 Q_o2: 10.40
Obs: On a path in the forest I'm on a path winding through the Enchanted Forest. The path winds its way from the north towards the west. In every other direction there are dark, sinister looking trees that wave their branches at me in an anthropomorphic kind of way. Ooh, I'm soooo scared!
Reward 2: 0
Action: enter west Q_t: 15.49 Q_o1: 25.32 Q_o2: 12.11
Obs: As I head off along the path a pathetic, mangey looking wolf jumps out from behind a bush and slobbers at me in what he probably hopes is a menacing way.    'I am the Big Bad Wolf!' says he. Then, after an awkward pause he adds 'well, what are you waiting for? Aren't you going to ask me the way to Grandma's house?' He looks at me quizzically for a moment and then slaps his forehead.    'My humblest apologies!' he says. 'You are not the young woman I seek. She of the crimson shawl must needs be elsewhere!' He darts back into the forest.   At a fork in the path I'm at a fork in the path. Not the kind of fork you eat with, the sort where the path forks; you know, like a fork. To the north the path widens into a clearing. To the west it disappears into the forest.
Reward 3: 0
Action: wear wrist Q_t: 17.00 Q_o1: 24.75 Q_o2: 12.40
Obs: I put on the wristwatch.
Reward 4: 0
Action: east Q_t: 16.39 Q_o1: 25.39 Q_o2: 9.43
Obs: On a path in the forest I'm on a path winding through the Enchanted Forest. The path winds its way from the north towards the west. In every other direction there are dark, sinister looking trees that wave their branches at me in an anthropomorphic kind of way. Ooh, I'm soooo scared!
Reward 5: 0
Action: remove these Q_t: 15.71 Q_o1: 24.27 Q_o2: 13.06
Obs: (the wristwatch) I take off the wristwatch.
Reward 6: 0
Action: west Q_t: 16.99 Q_o1: 23.31 Q_o2: 12.48
Obs: At a fork in the path I'm at a fork in the path. Not the kind of fork you eat with, the sort where the path forks; you know, like a fork. To the north the path widens into a clearing. To the west it disappears into the forest.
Reward 7: 0
Action: sit in north Q_t: 17.02 Q_o1: 26.87 Q_o2: 11.59
Obs: In a clearing in the forest I'm in a clearing. There is a teeny tiny cottage here complete with a thatched roof and leaded windows. It's just like something out of a fairy tale, funnily enough, except for the large hand painted sign designed to deter trespassers. Just to the north is the front porch. A little path lined by rows of white painted stones leads around the side of the house to the west.
Reward 8: 0
Action: south Q_t: 17.09 Q_o1: 26.97 Q_o2: 12.43
Obs: At a fork in the path I'm at a fork in the path. Not the kind of fork you eat with, the sort where the path forks; you know, like a fork. To the north the path widens into a clearing. To the west it disappears into the forest.
Reward 9: 0
Action: north Q_t: 17.88 Q_o1: 24.75 Q_o2: 12.35
Obs: In a clearing in the forest I'm in a clearing. There is a teeny tiny cottage here complete with a thatched roof and leaded windows. It's just like something out of a fairy tale, funnily enough, except for the large hand painted sign designed to deter trespassers. Just to the north is the front porch. A little path lined by rows of white painted stones leads around the side of the house to the west.
Reward 10: 0
Action: west Q_t: 18.90 Q_o1: 26.40 Q_o2: 10.87
Obs: In the garden I'm in a little garden alongside the cottage. There's a vegetable plot in a sunny spot by the wall and on the other side of the garden a somewhat neglected ornamental pond. There's obviously been some attempt at landscaping but the effect is less Capability Brown and more incapable bruin. A ramshackle potting shed completes the image of horticultural incompetence. To the north, an overgrown pergola leads into a meadow.
Reward 11: 0
Action: undo wristwatc with overripe Q_t: 17.60 Q_o1: 27.46 Q_o2: 8.95
Obs: (first taking the overripe pumpkin) That doesn't seem to be something I can unlock.
Reward 12: 0
Action: put on leaves Q_t: 20.13 Q_o1: 28.64 Q_o2: 11.16
Obs: (first taking some leaves) I take hold of the leaves and give them a good yank. Something pops out of the ground in a great shower of earth! There seems to be some kinda root vegetable actually attached to the leaves. Weird; I always thought these things came from tins!  [My score has just gone up by three points.]
Reward 13: 3
Action: north Q_t: 17.56 Q_o1: 26.21 Q_o2: 6.55
Obs: In a small meadow I'm in a small meadow at the back of the cottage, separated from the forest by a rickety fence. In the middle of the meadow is what appears to be a homemade wind turbine! To the south, through the tangled pergola, is the garden. Daisy the cow is standing here, chewing disinterestedly on a mouthful of hay.
Reward 14: 0
Action: put down wrist Q_t: 18.71 Q_o1: 24.84 Q_o2: 11.60
Obs: Dropped.
Reward 15: 0
Action: put down turnip Q_t: 18.16 Q_o1: 25.52 Q_o2: 10.35
Obs: I drop the turnip on the ground. It bounces a couple of times and rolls under Daisy's nose. On seeing this delicious morsel, Daisy gobbles down the turnip excitedly before sidling over to me and nuzzling my chest. I get impression she thinks I've got more turnips hidden somewhere about my person. I can't imagine where she gets that idea.  [My score has just gone up by three points.]
Reward 16: 3
Action: take all Q_t: 15.30 Q_o1: 22.74 Q_o2: 10.29
Obs: wristwatch: Taken. oil can: I can't reach it from here! It's all the way up on top of that rickety structure! Daisy the cow: I don't suppose Daisy the cow would care for that.  Daisy's back half attempts to dance the tango with her front half.
Reward 17: 0
Action: put other down Q_t: 15.72 Q_o1: 23.03 Q_o2: 9.99
Obs: (the wristwatch down) Dropped.  Daisy curtsies towards some imaginary audience.
Reward 18: 0
Action: get in south Q_t: 16.10 Q_o1: 22.53 Q_o2: 9.38
Obs: In the garden I'm in a little garden alongside the cottage. There's a vegetable plot in a sunny spot by the wall and on the other side of the garden a somewhat neglected ornamental pond. There's obviously been some attempt at landscaping but the effect is less Capability Brown and more incapable bruin. A ramshackle potting shed completes the image of horticultural incompetence. To the north, an overgrown pergola leads into a meadow. Daisy the cow walks towards me from the north.
Reward 19: 0
Action: enter Q_t: 15.54 Q_o1: 23.17 Q_o2: 9.74
Obs: In the potting shed I'm in the potting shed, a ramshackle structure that was obviously cobbled together in five minutes some idle Sunday afternoon.  I didn't know bears had green fingers. In fact I didn't know they had fingers at all. Plant pots of all sizes are scattered about the shed untidily and the whole place is festooned with cobwebs. In the corner is an old metal trunk marked 'FISHING TACKLE.'  I can also see a pair of secateurs and a ratty old paintbrush here. Daisy the cow walks towards me from outside.
Reward 20: 0
Action: wear pair Q_t: 15.62 Q_o1: 22.69 Q_o2: 7.84
Obs: (first taking the pair of secateurs) I can't wear those!  Daisy begins to gambol and frolic, then seems to realise how silly she looks and stops abruptly.
Reward 21: 0
Action: sit on top of cow Q_t: 15.80 Q_o1: 22.55 Q_o2: 9.96
Obs: The moment I climb onto Daisy's back she begins to buck like a bull in a rodeo! I'm thrown high into the air and land in an ungraceful heap. Ouch. Let's not try that again!  Daisy's back half attempts to dance the tango with her front half.
Reward 22: 0
Action: take secateurs Q_t: 15.78 Q_o1: 23.39 Q_o2: 10.14
Obs: Taken.  Daisy begins to gambol and frolic, then seems to realise how silly she looks and stops abruptly.
Reward 23: 0
Action: out Q_t: 15.55 Q_o1: 22.51 Q_o2: 6.83
Obs: In the garden I'm in a little garden alongside the cottage. There's a vegetable plot in a sunny spot by the wall and on the other side of the garden a somewhat neglected ornamental pond. There's obviously been some attempt at landscaping but the effect is less Capability Brown and more incapable bruin. A ramshackle potting shed completes the image of horticultural incompetence. To the north, an overgrown pergola leads into a meadow. Daisy the cow walks towards me from inside.
Reward 24: 0
Action: enter Q_t: 15.35 Q_o1: 23.12 Q_o2: 9.70
Obs: In the potting shed I'm in the potting shed, a ramshackle structure that was obviously cobbled together in five minutes some idle Sunday afternoon.  I didn't know bears had green fingers. In fact I didn't know they had fingers at all. Plant pots of all sizes are scattered about the shed untidily and the whole place is festooned with cobwebs. In the corner is an old metal trunk marked 'FISHING TACKLE.'  I can also see an overripe pumpkin and a ratty old paintbrush here. Daisy the cow walks towards me from outside.
Reward 25: 0
Action: take paintbrus Q_t: 15.51 Q_o1: 23.05 Q_o2: 9.20
Obs: Taken.  Daisy curtsies towards some imaginary audience.
Reward 26: 0
Action: out Q_t: 15.22 Q_o1: 22.87 Q_o2: 10.67
Obs: In the garden I'm in a little garden alongside the cottage. There's a vegetable plot in a sunny spot by the wall and on the other side of the garden a somewhat neglected ornamental pond. There's obviously been some attempt at landscaping but the effect is less Capability Brown and more incapable bruin. A ramshackle potting shed completes the image of horticultural incompetence. To the north, an overgrown pergola leads into a meadow. Daisy the cow walks towards me from inside.
Reward 27: 0
Action: east Q_t: 15.37 Q_o1: 22.58 Q_o2: 9.79
Obs: In a clearing in the forest I'm in a clearing. There is a teeny tiny cottage here complete with a thatched roof and leaded windows. It's just like something out of a fairy tale, funnily enough, except for the large hand painted sign designed to deter trespassers. Just to the north is the front porch. A little path lined by rows of white painted stones leads around the side of the house to the west. Daisy the cow walks towards me from the west.
Reward 28: 0
Action: put other down Q_t: 15.52 Q_o1: 22.78 Q_o2: 8.81
Obs: (the ratty old paintbrush down) Dropped.  Daisy's back half attempts to dance the tango with her front half.
Reward 29: 0
Action: enter west Q_t: 15.38 Q_o1: 22.79 Q_o2: 10.02
Obs: In the garden I'm in a little garden alongside the cottage. There's a vegetable plot in a sunny spot by the wall and on the other side of the garden a somewhat neglected ornamental pond. There's obviously been some attempt at landscaping but the effect is less Capability Brown and more incapable bruin. A ramshackle potting shed completes the image of horticultural incompetence. To the north, an overgrown pergola leads into a meadow. Daisy the cow walks towards me from the east.
Reward 30: 0
Action: sit on top of e Q_t: 15.44 Q_o1: 22.38 Q_o2: 8.91
Obs: In a clearing in the forest I'm in a clearing. There is a teeny tiny cottage here complete with a thatched roof and leaded windows. It's just like something out of a fairy tale, funnily enough, except for the large hand painted sign designed to deter trespassers. Just to the north is the front porch. A little path lined by rows of white painted stones leads around the side of the house to the west.  I can see a ratty old paintbrush here. Daisy the cow walks towards me from the west.
Reward 31: 0
Action: put on paintbrus Q_t: 15.55 Q_o1: 23.10 Q_o2: 7.94
Obs: (first taking the ratty old paintbrush) I can't wear that!  Daisy curtsies towards some imaginary audience.
Reward 32: 0
Action: west Q_t: 16.01 Q_o1: 23.40 Q_o2: 10.24
Obs: In the garden I'm in a little garden alongside the cottage. There's a vegetable plot in a sunny spot by the wall and on the other side of the garden a somewhat neglected ornamental pond. There's obviously been some attempt at landscaping but the effect is less Capability Brown and more incapable bruin. A ramshackle potting shed completes the image of horticultural incompetence. To the north, an overgrown pergola leads into a meadow. Daisy the cow walks towards me from the east.
Reward 33: 0
Action: enter Q_t: 15.03 Q_o1: 21.44 Q_o2: 9.39
Obs: In the potting shed I'm in the potting shed, a ramshackle structure that was obviously cobbled together in five minutes some idle Sunday afternoon.  I didn't know bears had green fingers. In fact I didn't know they had fingers at all. Plant pots of all sizes are scattered about the shed untidily and the whole place is festooned with cobwebs. In the corner is an old metal trunk marked 'FISHING TACKLE.'  I can also see an overripe pumpkin here. Daisy the cow walks towards me from outside.
Reward 34: 0
Action: out Q_t: 15.54 Q_o1: 23.38 Q_o2: 9.29
Obs: In the garden I'm in a little garden alongside the cottage. There's a vegetable plot in a sunny spot by the wall and on the other side of the garden a somewhat neglected ornamental pond. There's obviously been some attempt at landscaping but the effect is less Capability Brown and more incapable bruin. A ramshackle potting shed completes the image of horticultural incompetence. To the north, an overgrown pergola leads into a meadow. Daisy the cow walks towards me from inside.
Reward 35: 0
Action: sit in potting Q_t: 15.41 Q_o1: 22.81 Q_o2: 8.60
Obs: In the potting shed I'm in the potting shed, a ramshackle structure that was obviously cobbled together in five minutes some idle Sunday afternoon.  I didn't know bears had green fingers. In fact I didn't know they had fingers at all. Plant pots of all sizes are scattered about the shed untidily and the whole place is festooned with cobwebs. In the corner is an old metal trunk marked 'FISHING TACKLE.'  I can also see an overripe pumpkin here. Daisy the cow walks towards me from outside.
Reward 36: 0
Action: undo east with pumpkin Q_t: 15.25 Q_o1: 22.77 Q_o2: 9.97
Obs: (first taking the overripe pumpkin) That doesn't seem to be something I can unlock.  Daisy curtsies towards some imaginary audience.
Reward 37: 0
Action: stand up Q_t: 15.35 Q_o1: 23.56 Q_o2: 9.63
Obs: In the garden I'm in a little garden alongside the cottage. There's a vegetable plot in a sunny spot by the wall and on the other side of the garden a somewhat neglected ornamental pond. There's obviously been some attempt at landscaping but the effect is less Capability Brown and more incapable bruin. A ramshackle potting shed completes the image of horticultural incompetence. To the north, an overgrown pergola leads into a meadow. Daisy the cow walks towards me from inside.
Reward 38: 0
Action: sit on top of east Q_t: 15.45 Q_o1: 22.70 Q_o2: 10.18
Obs: In a clearing in the forest I'm in a clearing. There is a teeny tiny cottage here complete with a thatched roof and leaded windows. It's just like something out of a fairy tale, funnily enough, except for the large hand painted sign designed to deter trespassers. Just to the north is the front porch. A little path lined by rows of white painted stones leads around the side of the house to the west. Daisy the cow walks towards me from the west.
Reward 39: 0
Action: south Q_t: 15.34 Q_o1: 24.18 Q_o2: 9.43
Obs: At a fork in the path I'm at a fork in the path. Not the kind of fork you eat with, the sort where the path forks; you know, like a fork. To the north the path widens into a clearing. To the west it disappears into the forest. Daisy the cow walks towards me from the north.
Reward 40: 0
Action: north Q_t: 15.34 Q_o1: 22.56 Q_o2: 6.68
Obs: In a clearing in the forest I'm in a clearing. There is a teeny tiny cottage here complete with a thatched roof and leaded windows. It's just like something out of a fairy tale, funnily enough, except for the large hand painted sign designed to deter trespassers. Just to the north is the front porch. A little path lined by rows of white painted stones leads around the side of the house to the west. Daisy the cow walks towards me from the south.
Reward 41: 0
Action: south Q_t: 15.03 Q_o1: 22.85 Q_o2: 9.76
Obs: At a fork in the path I'm at a fork in the path. Not the kind of fork you eat with, the sort where the path forks; you know, like a fork. To the north the path widens into a clearing. To the west it disappears into the forest. Daisy the cow walks towards me from the north.
Reward 42: 0
Action: east Q_t: 15.76 Q_o1: 23.18 Q_o2: 7.07
Obs: On a path in the forest I'm on a path winding through the Enchanted Forest. The path winds its way from the north towards the west. In every other direction there are dark, sinister looking trees that wave their branches at me in an anthropomorphic kind of way. Ooh, I'm soooo scared! Daisy the cow walks towards me from the west.
Reward 43: 0
Action: stand on north Q_t: 16.25 Q_o1: 24.81 Q_o2: 10.82
Obs: In the enchanted forest A large wooden sign informs me that I'm in the Enchanted Forest. What it doesn't tell me is how the hell I get out! Everywhere I look there are trees, trees, trees. This place definately has a 'tree' theme going. A clearly marked path leads south and west, whilst in every other direction are, well, trees. An old pedlar is sitting on a log here, tending to his bunions. Daisy the cow walks towards me from the south.   'Ah, now there's a lovely specimen!' says the Pedlar, getting up from his log and feeling Daisy's knees. He seems satisfied. 'all present and correct! Here you go love   you'll be cured in no time!' He opens his suitcase and hands me a small packet, then leads Daisy off into the forest.  [My score has just gone up by three points.]
Reward 44: 3
Action: east Q_t: 14.17 Q_o1: 22.11 Q_o2: 9.41
Obs: Lost in the forest I'm walking in a trackless wilderness of tall, straight trees whose mighty trunks remind me of the time I got lost among the Axminsters at CarpetWorld aged six. I can go in just about any direction, but they all look the same to me.
Reward 45: 0
Action: sit in east Q_t: 14.59 Q_o1: 21.87 Q_o2: 12.75
Obs: Lost in the forest I'm wandering in the forest lost and completely disorientated! Everywhere I look there are strange, twisted trees, and what's more they all look exactly the same! I'm beginning to think it was a really bad idea to leave the path...
Reward 46: 0
Action: east Q_t: 15.09 Q_o1: 22.62 Q_o2: 11.72
Obs: On a path in the forest I'm on a path winding through the Enchanted Forest. The path winds its way from the north towards the west. In every other direction there are dark, sinister looking trees that wave their branches at me in an anthropomorphic kind of way. Ooh, I'm soooo scared!
Reward 47: 0
Action: north Q_t: 14.76 Q_o1: 22.36 Q_o2: 12.53
Obs: In the enchanted forest A large wooden sign informs me that I'm in the Enchanted Forest. What it doesn't tell me is how the hell I get out! Everywhere I look there are trees, trees, trees. This place definately has a 'tree' theme going. A clearly marked path leads south and west, whilst in every other direction are, well, trees.
Reward 48: 0
Action: north Q_t: 14.61 Q_o1: 21.63 Q_o2: 7.98
Obs: Lost in the forest I'm wandering in the forest lost and completely disorientated! Everywhere I look there are strange, twisted trees, and what's more they all look exactly the same! I'm beginning to think it was a really bad idea to leave the path...  From somewhere comes the blood curdling howl of a wolf...
Reward 49: 0
Action: go in east Q_t: 14.55 Q_o1: 23.47 Q_o2: 11.69
Obs: On a path in the forest I'm on a path winding through the Enchanted Forest. The path winds its way from the north towards the west. In every other direction there are dark, sinister looking trees that wave their branches at me in an anthropomorphic kind of way. Ooh, I'm soooo scared!
Reward 50: 0
Episode Score 9
\end{lstlisting}
\end{appendices}

\end{document}